\documentclass{article}

\usepackage[preprint]{neurips_2026}

\usepackage[utf8]{inputenc} 
\usepackage[T1]{fontenc}    
\usepackage{hyperref}       
\usepackage{url}            
\usepackage{booktabs}       
\usepackage{amsfonts}       
\usepackage{nicefrac}       
\usepackage{microtype}      
\usepackage{xcolor}         
\usepackage{graphicx}       

\usepackage{amsmath}
\usepackage{amssymb}
\usepackage{mathtools}
\usepackage{bm}
\usepackage{tabularx}
\usepackage{array}
\usepackage{float}
\usepackage{subcaption}
\usepackage{wrapfig}
\newcommand{\Var}{\mathrm{Var}}
\usepackage{subcaption}
\title{Complete-muE: Optimal Hyperparameter Transfer and Scaling for MoE Models}

\author{%
  \textbf{Hongwu Peng, Ohiremen Dibua, Yuanjun Xiong, Yifan Gong, Jin Huang, Lantao Yu,} \\ \textbf{Mamshad nayeem Rizve, Kevin Duarte, Siddharth Iyer, Jianming Zhang, Yan Kang} \\
      Adobe Research \\
  \texttt{\{hongwup, dibua, yxiong, yifang, jinhuang, lantaoy\}@adobe.com} \\
  \texttt{\{mrizve, kduarte, siddharthi, jianmzha, yankang\}@adobe.com}
  }

\begin{document}

\maketitle

\begin{abstract}
We propose Complete-muE, a framework which targets hyperparameter transfer across dense FFN and any Mixture-of-Experts (MoE) setups in transformer blocks. 
Existing tools such as $\mu$P (requires fixed architectue) or SDE (requires fixed per-step token count) cannot directly solve the hyperparameter transfer problem in MoE setups because Dense to MoE transfer or MoE total experts scaling changes both architecture and tokens per expert. Complete-muE solves this challenge with a two-bridge system: Bridge~I maps between dense FFN and Dense MoE by active-width $\mu$P with a normalized router scale. Bridge~II maps between Dense MoE and sparse MoE by activated-expert scaling, where the first-order SDE LR/WD correction cancels while a bounded residual $\sigma_0$ shift remains. The resulting transfer rule, which we term as Complete muE, covers changes in activated experts, total capacity, granularity, and shared/group-balanced hybrids for MoE models as well as network width/depth, batch size, and duration changes for general Transformer models. Extensive language model and diffusion model pretraining experiments confirm that complete-muE yields relatively stable hyperparameter optima across model architectures and parameter counts---with only minor drift consistent with the non-strict SDE behavior of Bridge~II. In practice this drift is small enough that hyperparameters tuned on a single dense reference transfer near-optimally to all MoE configurations---\emph{tune dense once, transfer to all} is the practical recipe at the core of Complete-muE. This enables MoE models to achieve accelerated convergence speedup over dense models when scaling model capacity without costly hyperparameter search.
\end{abstract}

\section{Introduction}

The goal of this paper is to transfer tuned initialization and AdamW hyperparameters across dense FFN and Mixture-of-Experts (MoE) transformer blocks. MoE exposes scaling axes absent from dense FFNs---per-expert width $h$, activated experts $a$, total experts $N$, shared experts, and group-balanced routing \cite{shazeer2017moe,fedus2022switch,zoph2022stmoe,dai2024deepseekmoe,deepseekv3_2024,tang2025pangu,huawei2025pangu,meituan2025longcat}. Changing these axes also changes expert data exposure: with balanced routing, each expert processes roughly $Ba/N$ tokens ($B$ is per-iteration batch size) per step and $TBa/N$ ($T$ is total training iterations) tokens over training. Dense-to-sparse transfer and total-expert scaling therefore couple architecture and workload transfer.

Existing tools cover only one side. $\mu$P, originally developed for dense models, connects dense FFNs to Dense MoE through active width but cannot deal with changes in per-expert token batch size \cite{yang2022tensor,malasnicki2025mupmoe}; SDE rules transfer across token batch size for a fixed model architecture but cannot handle architecture changes \cite{malladi2022sdes,cerebras2025completep,apple2025completedmup}. Complete-muE solves parameter transfer of MoE, which needs to simultaneously deal with both batch token count and architecture changes, with two bridges: (i) dense FFN $\leftrightarrow$ Dense MoE via active-width $\mu$P and route scale $r_a=a$; and (ii) Dense MoE $\leftrightarrow$ sparse MoE via activated-expert scaling, where expert-side batch and duration ratios match and the first-order SDE LR/WD correction cancels, while a residual $\sigma_0$ shift remains and produces bounded, minor hyperparameter drift in practice. Capacity, granularity, shared experts, group-balanced routing, and standard width/depth/batch/duration changes then follow by composition.

We extensively evaluated Complete muE on language modeling (LM)/diffusion (DF) tasks, and find Complete muE enables relatively stable hyperparameter transfer across all MoE architecture/scale/training duration combinations, with only minor drift consistent with Bridge~II's non-strict SDE behavior. In practice this drift is small enough that hyperparameters tuned on a single dense reference transfer near-optimally to all MoE configurations---\emph{tune dense once, transfer to all MoE settings} is the practical recipe at the core of Complete-muE. Both controlled small-scale axis sweeps and large-scale multimodal/LM runs directly verify this recipe: a single dense calibration delivers consistent MoE gains across MoE axes and across modalities ($256$P/$512$P images, $240$P key frames, $240$P $5$s videos, LM). We also benchmark MoE granularity vs capacity to show the real scaling trade-offs, and observe that capacity scaling under moderate granularity scaling is more beneficial. Our large scale MoE runs with Complete-muE enabled reach roughly $4.5\times$ speedup for $240$P $5$s video diffusion model and $5.3\times$--$5.5\times$ LLM convergence speedups with 100k training iterations. Our contributions are as follows: (1) we identify the key bottlenecks of dense-sparse MoE and MoE capacity scaling transfer cases; (2) to bridge the gap, we derive the two-bridge activated experts transfer rule through dense-dense MoE, and dense MoE-sparse MoE transfers; (3) we further compose those transfer rules to Complete muE, a framework which generalize to transfer across any MoE setups; and (4) we validate Complete muE in controlled small scale sweep and larger-scale LLM/multimodal experiments, and find Complete muE yields relatively stable hyperparameter optima across all MoE setups, with mild drift consistent with the non-strict SDE behavior of Bridge~II. With Complete muE enabled, large scale MoE training can achieve significant convergence speedup with minimum tuning overhead.

\section{Related Work and Theoretical Motivation}

Complete-muE is motivated by a mismatch between two lines of scaling work. MoE layers add active-width and routing-structure choices, while sparse routing changes the token batch size seen by each expert. Existing parameterizations handle the former for dense model-size changes, and SDE rules handle the latter for fixed parameterized models; neither alone covers changes that alter both the routed layer and expert-side data exposure.

\paragraph{Sparse MoE architectures and stability.}
Sparse MoE has progressed from sparsely gated layers to switch/top-$k$ transformers with improved routing, load balancing, specialization, and system support \cite{shazeer2017moe,fedus2022switch,zoph2022stmoe,lepikhin2020gshard}. Recent large systems add fine-grained experts, shared experts, group-balanced routing, and optimized sparse training \cite{dai2024deepseekmoe,deepseekv3_2024,tang2025pangu,huawei2025pangu,meituan2025longcat}; related diffusion/image MoE systems show similar benefits for generative models \cite{sun2024ec,akiti2026nucleus}. These works motivate our target design space but do not provide deterministic transfer rules for initialization, learning rate, weight decay, and parameter groups across $h$, $a$, $N$, and hybrid layouts.

\paragraph{Hyperparameter transfer and $\mu$-parameterization.}
Tensor Programs and $\mu$P provide width-wise zero-shot hyperparameter transfer by preserving feature-learning dynamics \cite{yang2022tensor}. Later work extends dense transfer to depth, practical transformer training, batch/duration effects, and diffusion transformers \cite{yang2023feature,bordelon2024infinite,cerebras2025completep,apple2025completedmup,zheng2025mupdiffusion}. We use this layer-level matching as the first ingredient: dense FFN and Dense MoE can be related through active width. However, plain $\mu$P has no explicit variable for the routed-token workload of a sparsely used expert, so it does not explain optimizer stability when $a$ or $N$ changes expert data exposure.

\paragraph{MoE-specific transfer.}
Recent MoE parameterization studies preserve learning-rate ranges under expert-count sweeps, study sparse MoE transformers across width/depth/expert axes, or analyze granularity and RMS-preserving routing \cite{malasnicki2025mupmoe,jiang2026moemup,ren2026rethinking}. Complete-muE is complementary: it stays in the standard AdamW regime and seeks one compositional rule for activated experts, total capacity, fixed-density granularity, shared experts, and group-balanced routing. The empirical stability observed in expert-count sweeps suggests that capacity scaling is not a separate primitive hyperparameter-transfer rule in $N$, but can be explained by composing active-width transfer with a sparsification step.

\paragraph{Expert workload and route scale.}
SDE-based optimizer transfer emphasizes the joint dependence on effective batch size and training horizon \cite{malladi2022sdes,apple2025completedmup}. For sparse MoE, the expert-side quantities scale as $B_{\mathrm{exp}}\propto Ba/N$ and $D_{\mathrm{exp}}\propto TBa/N$, so both must be tracked. At fixed $(N,h,B,T)$, changing $a$ scales these ratios together; after active-width matching, the first-order raw LR/WD SDE multiplier cancels. Separately, normalized routing requires an explicit route scale proportional to the activated-expert count to preserve the routed-branch update, consistent with routed/shared-branch scaling used in recent systems \cite{deepseekv3_2024,huawei2025pangu,meituan2025longcat}. These two observations define the bridges developed next.

\section{Method}
\label{sec:method}

Complete-muE turns FFN/MoE tuning into a dense-proxy calibration problem: scan a reference dense FFN once, then transfer the output multiplier, down-projection initialization, normalized route scale, and optimizer multipliers to the target FFN/MoE layout. The method first writes dense, routed, sparse, and hybrid blocks in one notation, then uses two bridges---active-width matching and expert-workload bookkeeping---to cover the full MoE design space.

\begin{figure}[t]
\centering
\includegraphics[width=\linewidth]{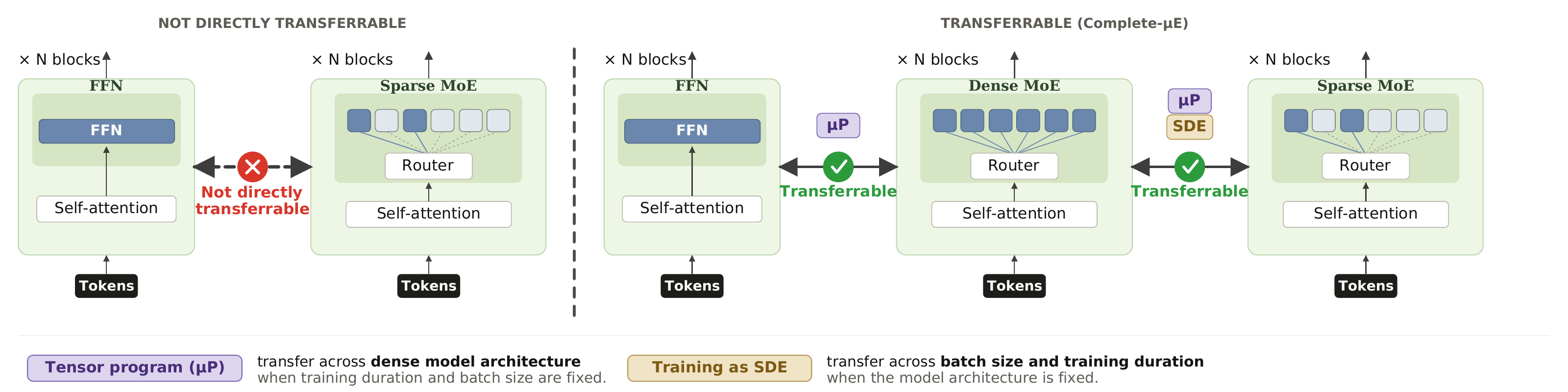}
\caption{Complete-muE transfers across dense FFN, Dense MoE, and sparse MoE}
\label{fig:complete-mue-teaser}
\end{figure}

\subsection{Problem statement and unified FFN/MoE formulation}
\label{sec:mue-setup}
\label{sec:mue-densemoe}

Let $d$ be the residual width. A dense FFN has hidden width $H$; an MoE layer has $N$ experts, per-expert width $h$, and $a$ activated experts, with routed active width $H_a=ah$. For hybrid blocks with dense/shared branches $\mathcal{D}$ and routed groups $\mathcal{G}$, $H_{\mathrm{tot}}=\sum_{m\in\mathcal{D}}H_m+a\sum_{g\in\mathcal{G}} h_g$. We write $\rho_d=d/d_\star$, $\rho_L=L/L_\star$, $\rho_B=B/B_\star$, and $\rho_D=D/D_\star$ with $D=TB$; Appendix~\ref{app:complete-mue} gives the full notation.

All FFN/MoE cases are represented as
\begin{equation}
\label{eq:unified-moe}
y(x)=A(H_{\mathrm{act}})\sum_{i=1}^{M} g_i(x)\,o_i(x),
\qquad
o_i(x)=W_{\mathrm{down}}^{(i)}u_i(x),
\end{equation}
where $H_{\mathrm{act}}$ is the hidden width active on token $x$. In MoE blocks, routing weights are normalized over the active experts, $\pi_i(x)\ge0$ and $\sum_{i\in\mathcal{A}(x)}\pi_i(x)=1$. Table~\ref{tab:unified-moe-cases} instantiates Eq.~\eqref{eq:unified-moe} for dense FFN, Dense MoE, and sparse MoE.

\begin{table}[t]
\centering
\small
\setlength{\tabcolsep}{4.5pt}
\renewcommand{\arraystretch}{1.08}
\caption{How Eq.~\eqref{eq:unified-moe} specializes to dense FFN, Dense MoE, and sparse MoE.
}
\label{tab:unified-moe-cases}
\begin{tabularx}{\linewidth}{
>{\raggedright\arraybackslash}p{0.18\linewidth}
>{\raggedright\arraybackslash}X
>{\centering\arraybackslash}p{0.18\linewidth}
>{\centering\arraybackslash}p{0.14\linewidth}}
\toprule
Case & Choice of $g_i(x)$ in Eq.~\eqref{eq:unified-moe} & Active width $H_{\mathrm{act}}$ & Route scale \\
\midrule
Dense FFN
& One always-active dense block ($M=1$), \;$g_1(x)=1$
& $H$
& $1$ \\

Dense MoE
& All experts are active ($a=N$), \;$g_i(x)=r_N\pi_i(x)$,\; $\sum_{i=1}^{N}\pi_i(x)=1$
& $Nh$
& $r_N=N$ \\

Sparse MoE
& Top-$a$ routed experts, \;$g_i(x)=r_a\pi_i(x)\mathbf{1}\{i\in\mathcal{A}(x)\}$,\; $|\mathcal{A}(x)|=a$
& $ah$
& $r_a=a$ \\
\bottomrule
\end{tabularx}
\renewcommand{\arraystretch}{1.0}
\end{table}

Throughout this work we employ token-choice (top-$k$) routing, in which each token independently selects its $a$ highest-scoring experts. Compared with expert-choice routing---which assigns a fixed quota of tokens per expert but requires access to the full token sequence and thereby leaks future-token information in autoregressive language models, causing a train--inference mismatch~\cite{zhou2022mixture}---token-choice routing is causally consistent at inference and applies uniformly across both language and visual modalities. Importantly, token-choice routing fixes the active width $H_a=ah$ deterministically for every token, which makes the $\mu$P update-size matching in the bridges below exact: the per-token functional update scales are controlled by the fixed active set size $a$, rather than by a stochastic token-assignment process.

The transfer problem is to reuse a dense-FFN scan rather than retune every MoE target. Tensors whose fan-in is controlled by the residual width $d$---the FFN up/gate projections and router readout---follow ordinary backbone-width $\mu$P. Complete-muE adds rules only for the FFN/MoE output branch and normalized routed sums, using the AdamW gradient-magnitude-normalized regime in which positive scalar rescalings of raw gradients do not change the leading preconditioned update direction.

\subsection{Existing gaps in hyperparameter transfer for MoE}
\label{sec:mue-gap}

MoE design varies along $h$, $a$, and $N$, and modern layers may mix routed groups with shared experts. A useful dense-to-MoE rule must therefore work compositionally: the same dense-FFN scan should transfer when one axis changes or when several are combined.

Neither existing tool provides this guarantee alone. $\mu$P matches parameter updates for model-size changes and gives the correct dense FFN $\leftrightarrow$ Dense MoE transfer, but sparse routing changes expert-side token batch size: each expert processes roughly $Ba/N$ tokens per step and $TBa/N$ tokens over training. Standard SDE transfer can adjust optimizer hyperparameters for batch and horizon changes, but it assumes a fixed parameterized model and therefore does not fit into transfer problem of dense FFN, Dense MoE, and sparse MoE architectures directly \cite{malladi2022sdes,cerebras2025completep,apple2025completedmup}. Total-expert scaling has the same issue: changing $N$ at fixed $(a,h)$ changes both routed capacity and per-expert workload. Prior $\mu$P-matched expert-count sweeps nevertheless show only mild optimizer drift \cite{malasnicki2025mupmoe,jiang2026moemup}, suggesting that MoE capacity transfer might be feasible.

\subsection{Two bridges for compositional FFN/MoE transfer}
\label{sec:mue-bridges}

The missing step is to avoid treating dense-to-sparse MoE transfer, or any other MoE transfer problem, as one direct jump. We instead connect a dense-FFN reference to any target FFN/MoE layout through two better-grounded bridge cases. Bridge~I matches a dense FFN to a Dense MoE at the same active width, introducing the output multiplier, output initialization rule, and normalized-route correction. Bridge~II then moves from Dense MoE to sparse MoE by changing the number of activated experts while keeps the overall parameterized architecture unchanged; after the layer update size are matched by $\mu P$ rule, the remaining change is an expert-side token batch size changes that can be analyzed with SDE bookkeeping. These two bridges are the only primitive MoE-specific ingredients; all capacity, granularity, shared-expert, and group-balanced-routing rules in the next subsection are compositions of them.

\subsubsection{Bridge I: dense FFN $\leftrightarrow$ Dense MoE via active-width $\mu$P}
\label{sec:mue-expansion}

Bridge~I matches a dense FFN with active width $H$ to a Dense MoE with the same active width. At fixed backbone width $d$, matching forward variance and the AdamW one-step functional update to the unit-expansion dense companion gives
\begin{equation}
\label{eq:equivalent-mup}
A(H)=\frac{d}{H},
\qquad
\sigma_{\mathrm{down}}(d,H)=\left(\frac{H}{d}\right)^{1/2}\sigma_{\mathrm{down}}^{(1)}(d),
\qquad
\eta_{\mathrm{down}}(d,H)=\eta_{\mathrm{down}}^{(1)}(d).
\end{equation}
Backbone-width changes compose with ordinary $\mu$P factors; Appendix~\ref{app:dense-derivation} gives the derivation.

\paragraph{Dense-MoE factorization and normalized route scale.}
\label{sec:mue-dense-router}
When the same active width is factored as $H=ah$ with $a=N$ active experts, Eq.~\eqref{eq:equivalent-mup} still applies. Normalized routing, however, averages expert outputs and would shrink the update by $1/a$, so we use
\begin{equation}
\label{eq:route-scale-main}
r_a=a,
\end{equation}
with $r_N=N$ for Dense MoE. The optional forward-variance correction $F_{a,N}$ is derived to Appendix~\ref{app:router-derivation}, and normally $F_{a,N}$ can be set to 1 for sigmoid routing. Thus dense FFN $\leftrightarrow$ Dense MoE transfer reduces to active-width $\mu$P plus route-scale correction; sparse MoE still needs Bridge~II because it changes each expert's token batch size.

\subsubsection{Bridge II: Dense MoE $\leftrightarrow$ sparse MoE via active width and expert-side SDE}
\label{sec:mue-sparsity}

The second bridge handles transfer across the number of activated experts. A Dense MoE is the special case $a=N$; a sparse MoE uses $a<N$. Complete-muE resolves this sparse transfer in two steps. First, it matches the routed-FFN layer scales through the active width $H_a=ah$. Second, it analyzes how changing $a$ perturbs the expert-side stochastic training trajectory.

\paragraph{Layer-level transfer across activated experts.}
\label{sec:mue-sparsity-width}
With total experts $N$ and per-expert width $h$ fixed, Complete-muE applies the dense active-width rule to $H_a$ and retains the normalized-router correction:
\begin{equation}
\label{eq:sparsity-transfer}
A_a=\frac{d}{ah},
\qquad
r_a=a,
\qquad
\sigma_{\mathrm{down}}(d,a)=\left(\frac{ah}{d}\right)^{1/2}\sigma_{\mathrm{down}}^{(1)}(d),
\qquad
\eta_{\mathrm{down}}(d,a)=\eta_{\mathrm{down}}^{(1)}(d).
\end{equation}
Thus a sparse MoE is matched to the dense companion through its \emph{active width}, not through its total expert count. If $d$ also changes, the same ordinary backbone-width factors are multiplied on top. Appendix~\ref{app:router-derivation} shows that Eq.~\eqref{eq:sparsity-transfer} matches the one-step MoE layer update size and the forward scale up to the bounded factor $F_{a,N}$, where $F_{a,N}$ can be set to 1 for most cases.

\begin{figure}[t]
\centering
\begin{subfigure}[t]{0.499\textwidth}
\centering
\includegraphics[width=0.497\linewidth]{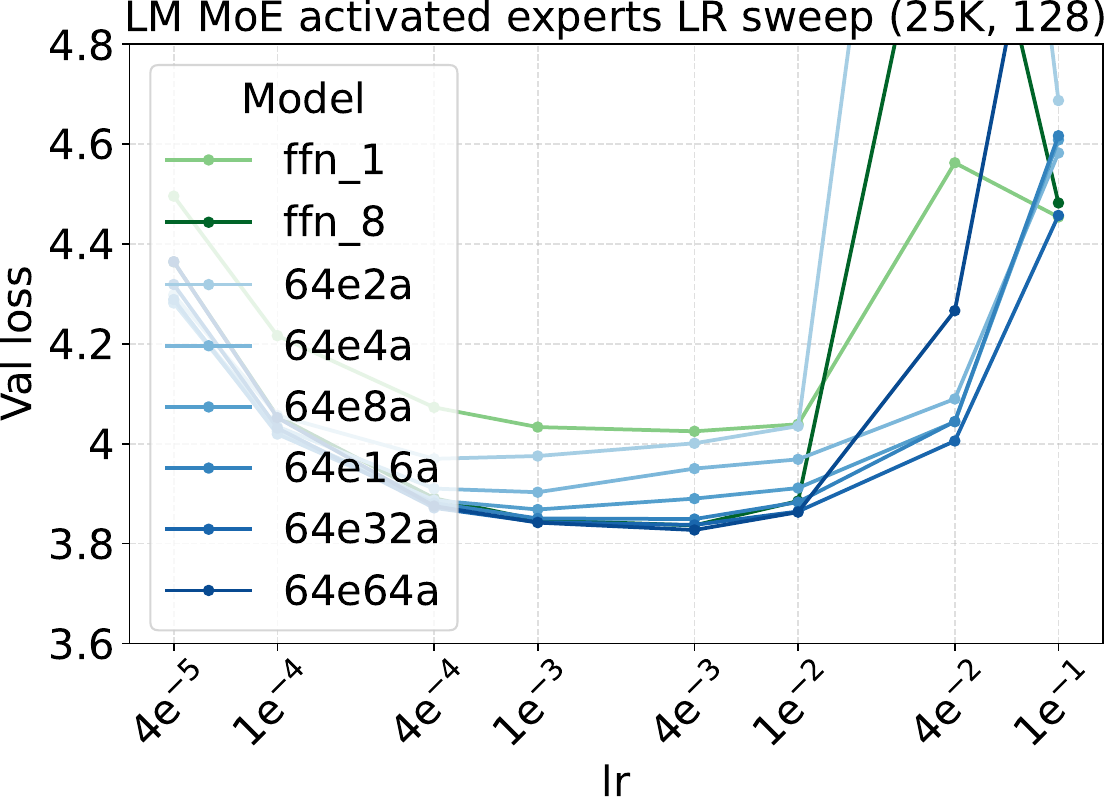}%
\hspace{0.006\linewidth}%
\includegraphics[width=0.497\linewidth]{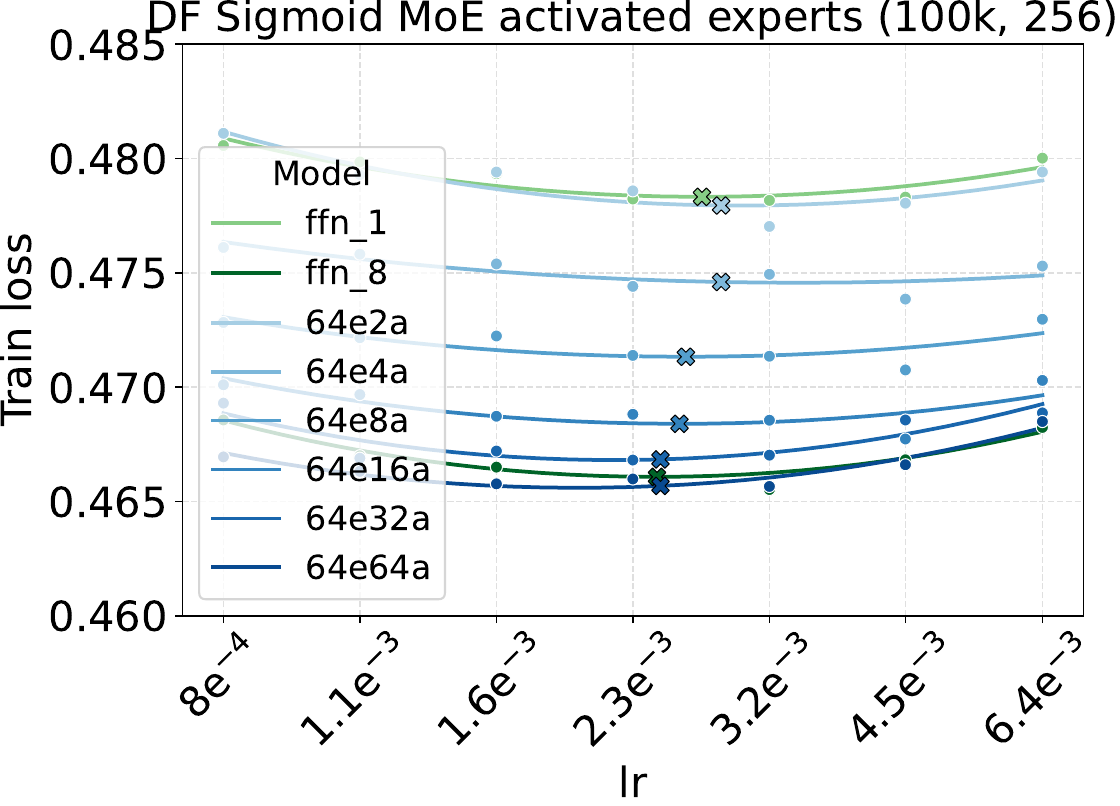}
\caption{Learning rate sweep}
\label{fig:activated-experts-sweep}
\end{subfigure}%
\hspace{0.002\textwidth}%
\begin{subfigure}[t]{0.499\textwidth}
\centering
\includegraphics[width=0.497\linewidth]{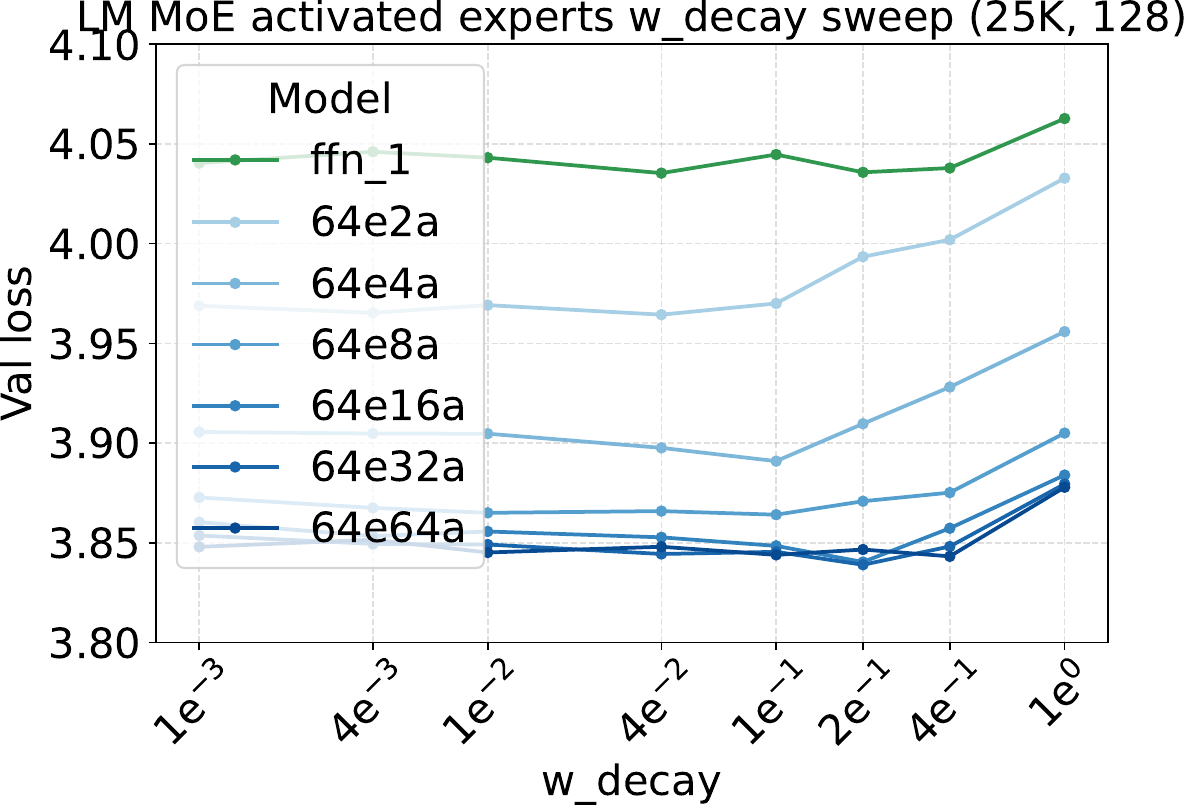}%
\hspace{0.006\linewidth}%
\includegraphics[width=0.497\linewidth]{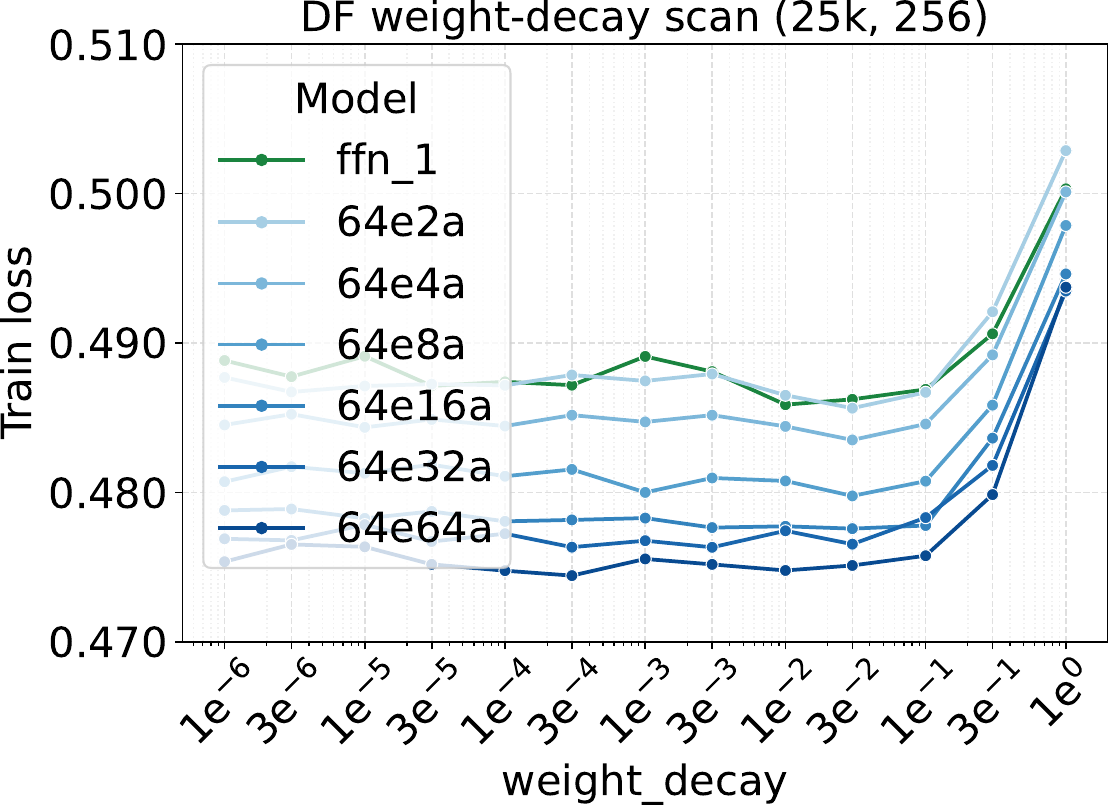}
\caption{Weight decay sweep}
\label{fig:activated-experts-weight-decay}
\end{subfigure}
\par\medskip
\begin{subfigure}[t]{0.499\textwidth}
\centering
\includegraphics[width=0.497\linewidth]{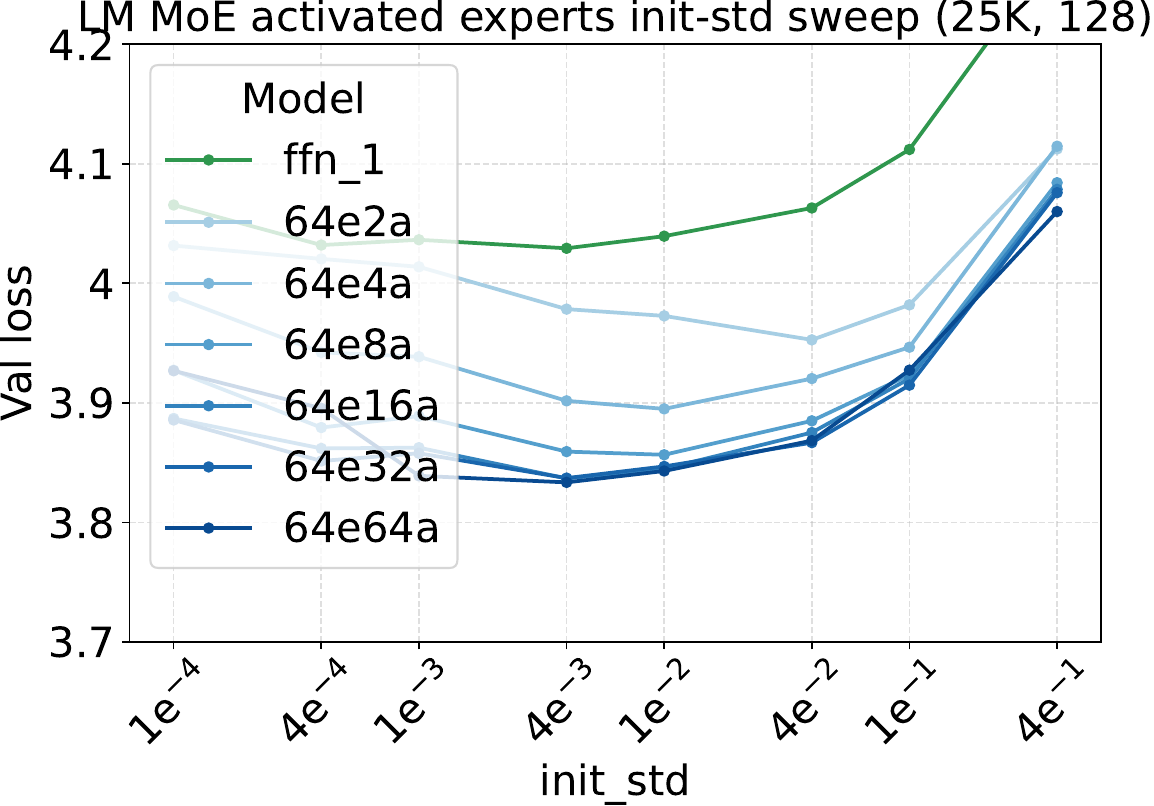}%
\hspace{0.006\linewidth}%
\includegraphics[width=0.497\linewidth]{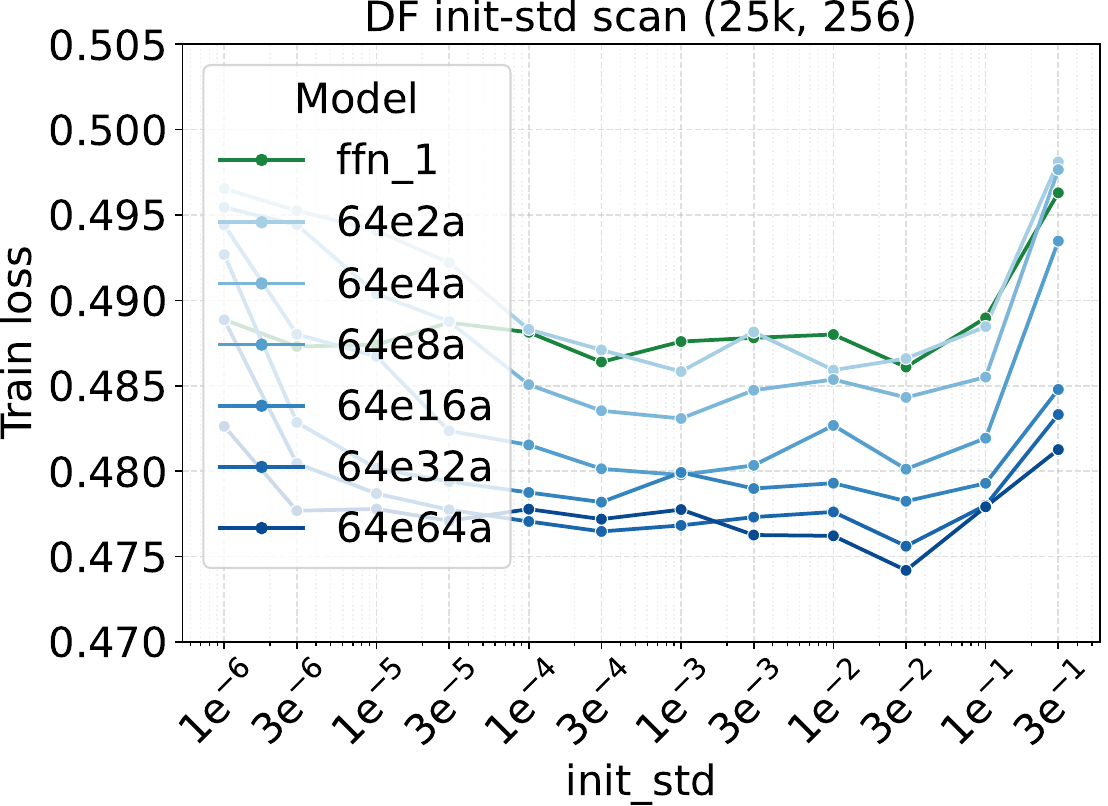}
\caption{Init std sweep}
\label{fig:activated-experts-init-std}
\end{subfigure}%
\hspace{0.002\textwidth}%
\begin{subfigure}[t]{0.499\textwidth}
\centering
\includegraphics[width=0.497\linewidth]{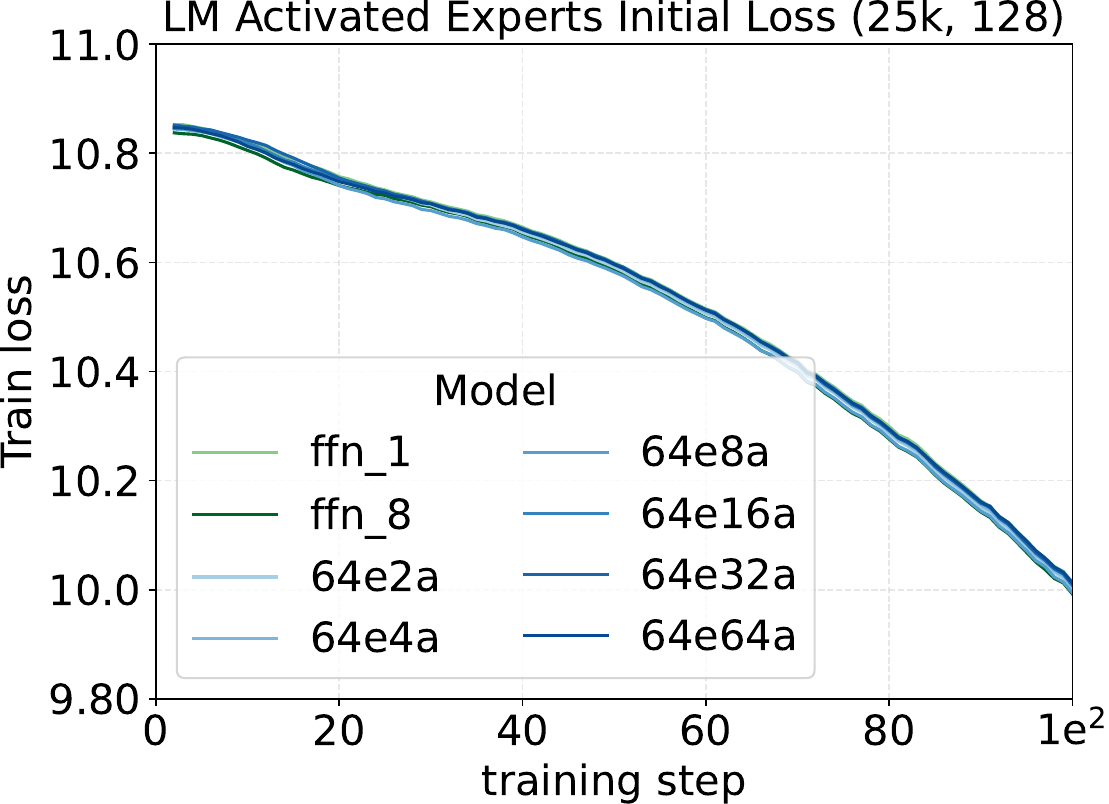}%
\hspace{0.006\linewidth}%
\includegraphics[width=0.497\linewidth]{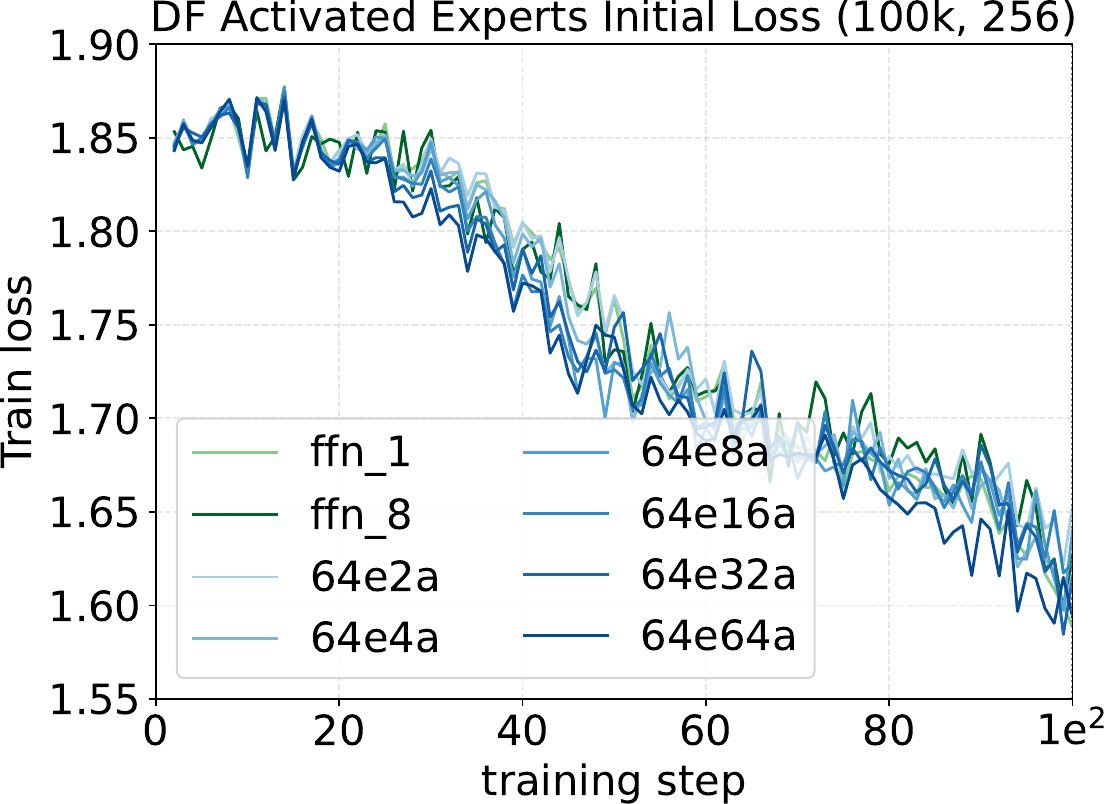}
\caption{Initial training loss}
\label{fig:activated-experts-initial-loss}
\end{subfigure}
\caption{Complete muE transfer across activated experts for both language-model (LM) and diffusion-model (DF)}
\label{fig:activated-experts-scaling}
\end{figure}

\paragraph{Expert-side SDE and cancellation across activated experts.}
\label{sec:mue-sparsity-sde}
The remaining effect is stochastic training process. Under approximate load balancing, one expert receives
\begin{equation}
\label{eq:expert-activated-factors}
B_{\mathrm{exp}}(a)\approx B\frac{a}{N},
\qquad
D_{\mathrm{exp}}(a)\approx TB\frac{a}{N},
\end{equation}
and therefore $\sigma_{\mathrm{exp}}(a)\propto B_{\mathrm{exp}}(a)^{-1/2}$. For a change $a\to a'$ at fixed global batch $B$ and fixed optimizer steps $T$, the expert-side batch and duration ratios are identical, $\rho_B^{\mathrm{exp}}=\rho_D^{\mathrm{exp}}=a'/a$---unlike a pure global batch increase, where only $\rho_B^{\mathrm{exp}}$ would change and the exact square-root rule $\eta'\!=\!\sqrt{\rho_B^{\mathrm{exp}}}\,\eta$ applies. Following the approximation philosophy of Complete$(d)$P~\cite{apple2025completedmup}, we hypothesize that after the layer-level matching of Eq.~\eqref{eq:sparsity-transfer}, the combined expert-side correction $\eta'\approx\eta\sqrt{\rho_B^{\mathrm{exp}}/\rho_D^{\mathrm{exp}}}$ captures the dominant LR/WD effect. Since $\rho_B^{\mathrm{exp}}=\rho_D^{\mathrm{exp}}$, the dense-style SDE correction cancels:
\begin{equation}
\label{eq:sde-cancel}
\eta'\approx \eta\sqrt{\frac{\rho_B^{\mathrm{exp}}}{\rho_D^{\mathrm{exp}}}}=\eta,
\qquad
\lambda'\approx \lambda\sqrt{\frac{\rho_B^{\mathrm{exp}}}{\rho_D^{\mathrm{exp}}}}=\lambda.
\end{equation}
What does shift is the expert-side signal-to-noise parameter: $\sigma_0(a')=\sigma_0(a)/\sqrt{\rho_B^{\mathrm{exp}}}$. Reducing sparsity (larger $a'$) lowers $\sigma_0$ and thereby improves expert-side SNR at fixed optimization horizon $H_{\mathrm{SDE}}=T\eta^2$---this is why denser MoEs can reach lower attainable loss even when raw hyperparameters are unchanged (see Appendix~\ref{app:sde-derivation}). This is the bridge result underlying Complete-muE. Under the approximate load-balancing assumptions above, after the layer-level reparameterization in Eq.~\eqref{eq:sparsity-transfer}, changing the number of activated experts does not require an additional first-order raw LR/WD correction. Activated-expert transfer is therefore not a strict SDE invariance: the residual $\sigma_0(a')=\sigma_0(a)/\sqrt{\rho_B^{\mathrm{exp}}}$ shift is not absorbed by the $\eta,\lambda$ correction, so some mild hyperparameter drift across $a$ is expected. Figure~\ref{fig:activated-experts-scaling} confirms this picture empirically: the loss curves versus learning rate, weight decay, and initialization standard deviation remain relatively stable across activated-expert counts---with only minor drift, consistent with the non-strict SDE transfer rather than with exact invariance---while larger $a$ consistently reaches lower attainable loss, as predicted by the improved expert-side SNR. This gives the second bridge, which we compose with the Dense FFN $\leftrightarrow$ Dense MoE bridge to handle the remaining MoE variants. Appendix~\ref{app:sde-derivation} gives the full SDE derivation, including the imbalanced-routing extension.

\subsection{Applications of Complete-muE: scaling across MoE settings}
\label{sec:mue-granularity}

The two bridges connect the dense reference to the remaining MoE settings by composition. The organizing insight is that total-expert and granularity changes are \emph{not} new primitive rules---they are derivable by composing the dense-width and activated-expert transfer rules established above, with the active width $H_a=ah$ as the single governing quantity in every case. In each case, the main-text rule is to match the target active width, apply the route scale on normalized routed groups, and rely on Appendix~\ref{app:composition-derivation} for the detailed cancellation arguments.

\paragraph{Total expert count (capacity scaling).}
For $(N,a,h)\to(N',a,h)$, introduce Dense-MoE companions of widths $Nh$ and $N'h$, then sparsify the target companion back to $a$ activated experts. This is an explicit two-step composition: (1)~\emph{Dense-width step}---scale the Dense-MoE companion from total width $Nh$ to $N'h$ via Bridge~I; (2)~\emph{Reverse-sparsity step}---apply Bridge~II to restore $a$ activated experts. The dense-width and sparsification factors cancel, leaving the same sparse-layer rule
\[
A_a=\frac{d}{ah},\qquad r_a=a,\qquad
\sigma_{\mathrm{down}}(d,a)=\left(\frac{ah}{d}\right)^{1/2}\sigma_{\mathrm{down}}^{(1)}(d),
\]
up to the bounded routing factor $F_{a,N}$. Because the reverse-sparsity step inherits Bridge~II's non-strict SDE behavior---$\sigma_0$ shifts with the per-expert workload and is not absorbed by the first-order $\eta,\lambda$ correction---some mild hyperparameter drift across $N$ is expected. Figure~\ref{fig:total-experts-transfer} confirms this empirically: the optimal loss region remains relatively stable across capacity settings, with only minor drift consistent with the non-strict SDE transfer. Increasing $N$ still changes the attainable loss through a capacity--noise trade-off: total specialization capacity grows, while each expert receives $B_{\mathrm{exp}},D_{\mathrm{exp}}\propto a/N$ tokens, worsening expert-side SNR ($\sigma_{\mathrm{exp}}\propto\sqrt{N/(Ba)}$) even as the raw transferred AdamW hyperparameters remain unchanged.

\begin{table*}[htb!]
\centering
\footnotesize
\setlength{\tabcolsep}{5pt}
\renewcommand{\arraystretch}{1.08}
\caption{Complete-muE layer-level rules. }
\label{tab:mue-rules}
\begin{tabularx}{\textwidth}{
>{\raggedright\arraybackslash}p{0.16\textwidth}
>{\raggedright\arraybackslash}X
>{\centering\arraybackslash}p{0.11\textwidth}
>{\centering\arraybackslash}p{0.08\textwidth}
>{\centering\arraybackslash}p{0.10\textwidth}
>{\centering\arraybackslash}p{0.06\textwidth}}
\toprule
Case & What is transferred & Multiplier $A$ & Route $R$ & Init std & LR \\
\midrule
Backbone width
& Router readout $W_{\mathrm{gate}}$, $\rho_d=d/d_\star$
& --
& --
& $\rho_d^{-1/2}$
& $\rho_d^{-1}$ \\

Backbone width
& FFN/MoE up \& gate projections $W_{\mathrm{up}},W_v,W_g$ (including per-expert copies in dense and sparse MoE), $\rho_d=d/d_\star$
& --
& --
& $\rho_d^{-1/2}$
& $\rho_d^{-1}$ \\

Dense FFN / gated MLP
& Output projection $W_{\mathrm{down}}$, $\rho_H=H/d$
& $\rho_H^{-1}$
& $1$
& $\rho_d^{-1/2}\rho_H^{1/2}$
& $\rho_d^{-1}$ \\

Dense MoE
& All experts active, $a=N$, $\rho_H=Nh/d$
& $\rho_H^{-1}$
& $N$
& $\rho_d^{-1/2}\rho_H^{1/2}$
& $\rho_d^{-1}$ \\

Activated experts
& Fixed $(N,h)$, active width $\rho_{H_a}=ah/d$
& $\rho_{H_a}^{-1}$
& $a$
& $\rho_d^{-1/2}\rho_{H_a}^{1/2}$
& $\rho_d^{-1}$ \\

Capacity / total experts
& Fixed $(a,h)$; compose Dense-MoE companion transfer and sparsification back to $a$
& $\rho_{H_a}^{-1}$
& $a$
& $\rho_d^{-1/2}\rho_{H_a}^{1/2}$
& $\rho_d^{-1}$ \\

Granularity (fixed $s=a/N$)
& Same ratio as Dense-MoE companion transfer, but final sparse active width $\rho_{H_a}=ah/d$
& $\rho_{H_a}^{-1}$
& $a$
& $\rho_d^{-1/2}\rho_{H_a}^{1/2}$
& $\rho_d^{-1}$ \\

Shared / group-balanced / hybrid
& Dense/shared branch, $\rho_{H_{\mathrm{tot}}}=H_{\mathrm{tot}}/d$
& $\rho_{H_{\mathrm{tot}}}^{-1}$
& $1$
& $\rho_d^{-1/2}\rho_{H_{\mathrm{tot}}}^{1/2}$
& $\rho_d^{-1}$ \\

Shared / group-balanced / hybrid
& Routed experts (group-balanced selection), $\rho_{H_{\mathrm{tot}}}=H_{\mathrm{tot}}/d$
& $\rho_{H_{\mathrm{tot}}}^{-1}$
& $a$
& $\rho_d^{-1/2}\rho_{H_{\mathrm{tot}}}^{1/2}$
& $\rho_d^{-1}$ \\
\bottomrule
\end{tabularx}
\renewcommand{\arraystretch}{1.0}
\end{table*}

\paragraph{Granularity and fixed-density scaling.}
If the routing density $s=a/N$ is fixed, the sparse active width $H_a=ah=sNh$ has the same ratio as the Dense-MoE companion width. Thus fixed-density granularity reduces to Dense-MoE width transfer followed by the same sparse output rule, with route scale $r_a=a$ on the source and $r_{a'}=a'$ on the target. Since each expert sees $Bs$ tokens per step and $TBs$ tokens over training---independent of the specific partition $(N,h)$, so $\rho_B^{\mathrm{exp}}=\rho_D^{\mathrm{exp}}=1$ across source and target---no extra first-order SDE correction is introduced.

\subsubsection{Shared, group-balanced-routing, and hybrid blocks}
\label{sec:mue-extensions}

The rule follows three steps: (1)~always-active dense/shared branches contribute active width with no route scale; (2)~routed MoE groups are selected using group-balanced routing, sharing one global route scale $r=a$ and global normalization $\sum_{g,e}\pi_{g,e}(x)=1$; (3)~one common FFN-output ABC parametrization is applied jointly to the total active width $H_{\mathrm{tot}}=\sum_{m\in\mathcal{D}}H_m+a\sum_{g\in\mathcal{G}} h_g$. For hybrid blocks, Complete-muE uses one active-width multiplier for the total active width and applies route scale only to routed groups:
\begin{equation}
\label{eq:hybrid-general}
y(x)=\frac{d}{H_{\mathrm{tot}}}\left[
\sum_{m\in\mathcal{D}} o_m(x)
+
a\sum_{g\in\mathcal{G}}\sum_{e=1}^{N_g} \pi_{g,e}(x)\,o_g^{(e)}(x)
\right].
\end{equation}
For one shared branch and one routed group,
\begin{equation}
\label{eq:shared-specialization}
y(x)=\frac{d}{H_{\mathrm{tot}}}\left[o_{\mathrm{sh}}(x)+a\sum_{e=1}^{N} \pi_e(x)\,o^{(e)}(x)\right],
\qquad
H_{\mathrm{tot}}=H_{\mathrm{sh}}+ah.
\end{equation}
Equivalently, shared, group-balanced-routing, and general hybrid blocks are treated as one expanded FFN of width $H_{\mathrm{tot}}$, with one global route scale $r=a$ on all selected routed experts; the special cases of shared experts (Eq.~\eqref{eq:shared-specialization}), grouped MoE without shared branches ($\mathcal{D}=\varnothing$, multiple routed groups), and shared-plus-grouped hybrids all follow from this general rule. The derivation is in Appendix~\ref{app:hybrid-derivation}.

\subsection{Operational recipe and global optimizer composition}
\label{sec:mue-summary}

Complete-muE's workflow begins from a single dense-FFN reference scan---tune once at the chosen backbone width, batch size, and training duration---and then maps to any target FFN/MoE layout through two composable tables applied in sequence.

\paragraph{Table~\ref{tab:mue-rules}: layer-level rule.}
For any target FFN/MoE block, (1)~identify the active width: $H$ for dense FFN, $ah$ for sparse MoE, or $H_{\mathrm{tot}}=\sum_{m\in\mathcal{D}}H_m+a\sum_{g\in\mathcal{G}}h_g$ for hybrid; (2)~read the output multiplier $A$, route scale $R$, and down-projection initialization $\sigma_{\mathrm{down}}$ from the corresponding row; (3)~apply route scale only to normalized routed sums---tensors controlled by backbone width $d$ (FFN up/gate, router readout) follow ordinary $\mu$P and need no additional rule. All MoE variants---activated-expert, total-expert, fixed-density granularity, shared-expert, and group-balanced-routing---map to the same active-width row because their layer-level and expert-side SDE effects cancel compositionally, as established in the preceding subsections.

\paragraph{Table~\ref{tab:optimizer-rules}: global optimizer factors.}
Two cases govern how the global AdamW multipliers interact with the layer-level rule. \emph{Case A---global schedule change at fixed architecture}: when global batch or duration changes, every routed expert sees the same proportional change ($\rho_B^{\mathrm{exp}}=\rho_B$, $\rho_D^{\mathrm{exp}}=\rho_D$), so the $\sqrt{\rho_B/\rho_D}$ factor applies uniformly to all parameters. Within Case A: when total tokens are fixed ($\rho_D=1$, batch up / steps down), all three global SDE objects $\sigma_0$, $\widetilde\lambda$, $H_{\mathrm{SDE}}$ are simultaneously preserved and $\eta'=\sqrt{\rho_B}\,\eta$ is exact; when training steps are fixed instead ($\rho_D=\kappa_B=\rho_B$, batch up / token budget grows), the horizon is preserved with $\eta'\approx\eta$ but $\sigma_0$ shifts by $1/\sqrt{\kappa_B}$, making the transfer approximate---empirically (Figure~\ref{fig:batch-size-transfer-fixed-iterations}) the optimal LR region remains relatively stable across batch sizes, with only minor drift consistent with the $\sigma_0$ shift. \emph{Case B---MoE architectural change at fixed schedule}: when expert count, routing density, or granularity changes at fixed global batch and duration, the expert-side ratios cancel ($\rho_B^{\mathrm{exp}}=\rho_D^{\mathrm{exp}}$), so Table~\ref{tab:mue-rules} already absorbs all such changes---no additional LR/WD multiplier is needed. Appendix~\ref{app:global-optimizer} gives the full SDE derivation of both cases. Across all three $\sigma_0$-shift situations (Bridge~II activated experts, capacity composition, and Case~A fixed-iteration batch transfer), the empirical drift is small enough that dense-tuned hyperparameters give near-optimal performance throughout, validating the \emph{tune dense once, transfer to all MoE settings} recipe that this section operationalizes.

\begin{table*}[htb!]
\centering
\footnotesize
\setlength{\tabcolsep}{4pt}
\renewcommand{\arraystretch}{1.05}
\caption{Remaining global AdamW transfer used together with Table~\ref{tab:mue-rules}.}
\label{tab:optimizer-rules}
\begin{tabularx}{\textwidth}{
>{\raggedright\arraybackslash}p{0.12\textwidth}
>{\centering\arraybackslash}p{0.08\textwidth}
>{\raggedright\arraybackslash}p{0.15\textwidth}
>{\centering\arraybackslash}p{0.10\textwidth}
>{\centering\arraybackslash}p{0.10\textwidth}
>{\centering\arraybackslash}p{0.10\textwidth}
>{\raggedright\arraybackslash}X}
\toprule
Axis & Ratio & Layer multiplier & $\eta,\lambda$ & AdamW $\epsilon$ & $1-\beta_{1,2}$ & Setting / comment \\
\midrule
Depth
& $\rho_L$
& residual branch $\rho_L^{-1}$
& $1$
& $1$
& $1$
& use standard CompleteP depth-sensitive parameter groups~\cite{cerebras2025completep} \\

Batch size
& $\rho_B$
& --
& $\rho_B^{1/2}$
& $\rho_B^{-1/2}$
& $\rho_B$
& fixed $D$; $T\mapsto T/\rho_B$ \\

Token budget / duration
& $\rho_D$
& --
& $\rho_D^{-1/2}$
& $\rho_D^{1/2}$
& $\rho_D^{-1}$
& fixed $B$; $T\mapsto \rho_D T$ \\
\midrule
\multicolumn{7}{l}{\footnotesize Combined batch+duration rule: $\eta,\lambda \propto \sqrt{\rho_B/\rho_D}$, \; AdamW $\epsilon \propto \sqrt{\rho_D/\rho_B}$, \; $1-\beta_{1,2} \propto \rho_B/\rho_D$.} \\
\bottomrule
\end{tabularx}
\renewcommand{\arraystretch}{1.0}
\end{table*}

\section{Experiments}

We evaluate Complete-muE along three axes: controlled LM and diffusion-transformer proxy sweeps, a single-H100 latency benchmark for MoE capacity/granularity scaling, and larger multimodal diffusion and language-model runs. Because many minima are broad, we focus on alignment of the low-loss region rather than the exact best learning rate in each sweep.

\subsection{Experimental setup and stable proxy-training recipe}
\label{sec:exp-setup}

\paragraph{Model families.}
The LM proxy uses the GPT-NeoX-20B tokenizer \cite{black2022gpt} with sequence length $2048$; the default configuration has width $d=128$, $32$ layers, batch size $128$, and $25$k optimizer steps, with $12.5$k/$6.25$k runs for duration and batch size studies. The diffusion proxy is a latent-diffusion transformer at $512\times512$ resolution with $16\times$ VAE compression, Qwen3-VL text conditioning \cite{bai2025qwen3}, width $d=128$, batch size $256$, and $100$k steps, with 25k/50k/200k steps runs for duration and batch size studies. The diffusion model family uses flow-matching linear interpolant \cite{lipman2022flow} and is trained for velocity prediction. Both families use gated attention \cite{qiu2025gated}, SwiGLU MLPs \cite{shazeer2020glu}, and AdamW with $\beta_1=\beta_2$ \cite{orvieto2025search,fernandez2026adam}; diffusion also uses AdaLN timestep conditioning \cite{peebles2023scalable}.

\paragraph{Schedules and reporting.}
We use a warmup-stable-decay schedule \cite{hu2024minicpm,hagele2024scaling,tissuescaling} with $1$k warmup for most runs. Decay occupies $20\%$ of $25$k/$50$k runs, $15\%$ of $100$k runs, and $10\%$ of $200$k runs; the short LM runs use the adjusted warmup/decay points described by their sweep. We report the last-window average loss: $1$k steps for $25$k runs, $1.5$k for $50$k, and $2$k for $100$k/$200$k. Large-scale LM evaluations use $13$ downstream benchmarks, detailed in Appendix~\ref{app:nlp-evaluation-details}.

Since all training runs process each sample at most once (single-epoch training), the training loss is an unbiased estimator of held-out loss and is interchangeable with validation loss in principle. In practice, the per-step diffusion training loss exhibits higher variance; we therefore report the last-window average training loss as described above to reduce this variance. For LLMs, the per-step validation loss variance is sufficiently low that we report it directly.

\paragraph{MoE notation and sweep protocol.}
We write \texttt{XeYa} for $X$ total experts and $Y$ activated experts, with \texttt{Zs} for shared experts and \texttt{Gg} for groups; e.g., \texttt{128e8a4g1s}. Unless stated otherwise, granularity, capacity, shared-expert, and group-balanced-routing studies match the dense unit-expansion active width. In activated-expert sweeps, \texttt{64e2a}, \texttt{64e4a}, \texttt{64e8a}, and \texttt{64e16a} correspond to active-width ratios $0.25$, $0.5$, $1$, and $2$. Each sweep applies Complete-muE first, then varies the base hyperparameter.

\subsection{Transfer across all MoE scaling axes}
\label{sec:exp-transfer}

Figure~\ref{fig:activated-experts-scaling} verifies Bridge~II at fixed total experts and per-expert width. Complete-muE keeps LR, WD, and init optima broad and aligned: representative ranges are LR $4\times10^{-4}$--$4\times10^{-3}$; WD $0.01$--$0.2$ for LM and $10^{-4}$--$3\times10^{-2}$ for diffusion; and init std $4\times10^{-4}$--$4\times10^{-2}$ for LM and $10^{-4}$--$3\times10^{-2}$ for diffusion. The first $100$ iterations nearly overlap, while larger $a$ lowers attainable loss.

\begin{figure}[t]
\centering
\begin{subfigure}[t]{0.499\textwidth}
\centering
\includegraphics[width=0.497\linewidth]{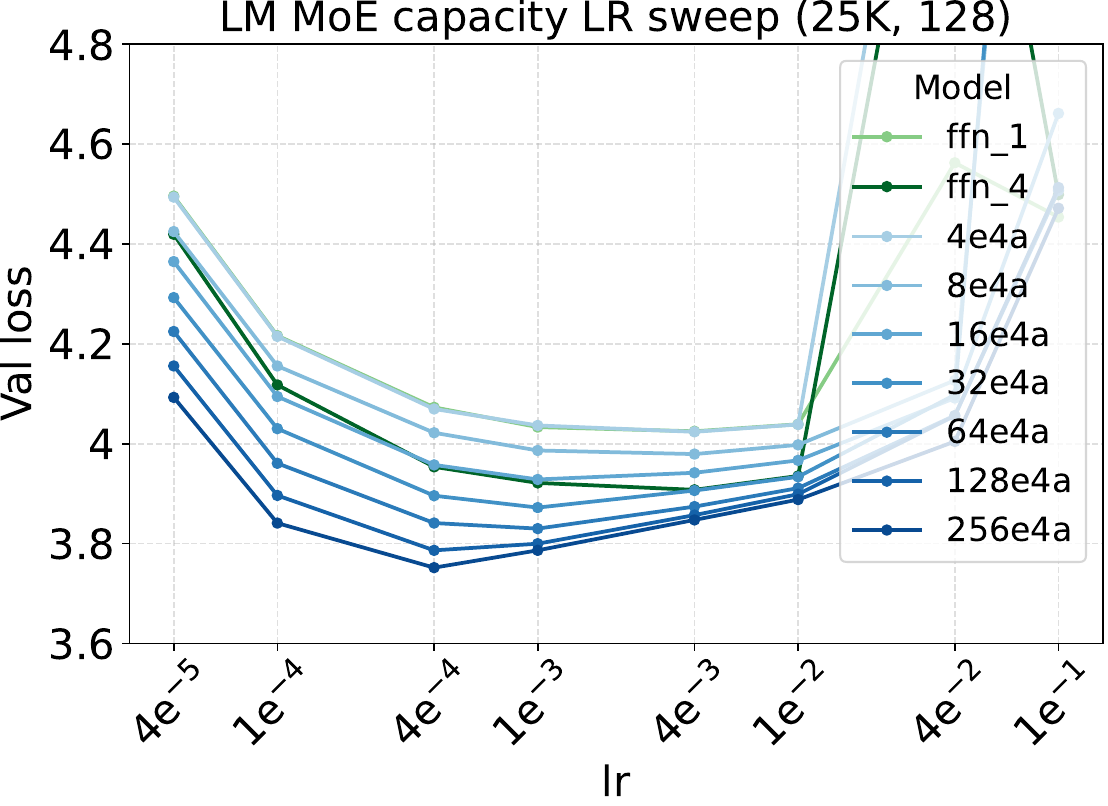}%
\hspace{0.006\linewidth}%
\includegraphics[width=0.497\linewidth]{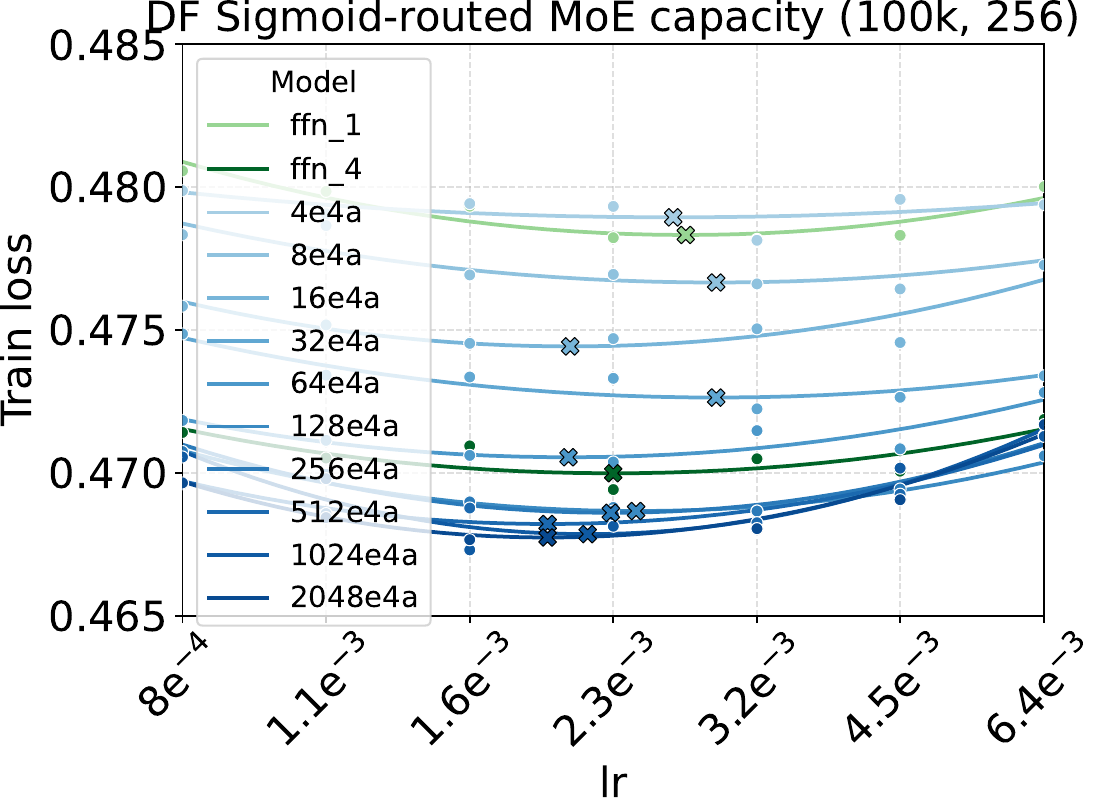}
\caption{MoE capacity learning-rate sweep}
\label{fig:total-experts-transfer}
\end{subfigure}%
\hspace{0.002\textwidth}%
\begin{subfigure}[t]{0.499\textwidth}
\centering
\includegraphics[width=0.497\linewidth]{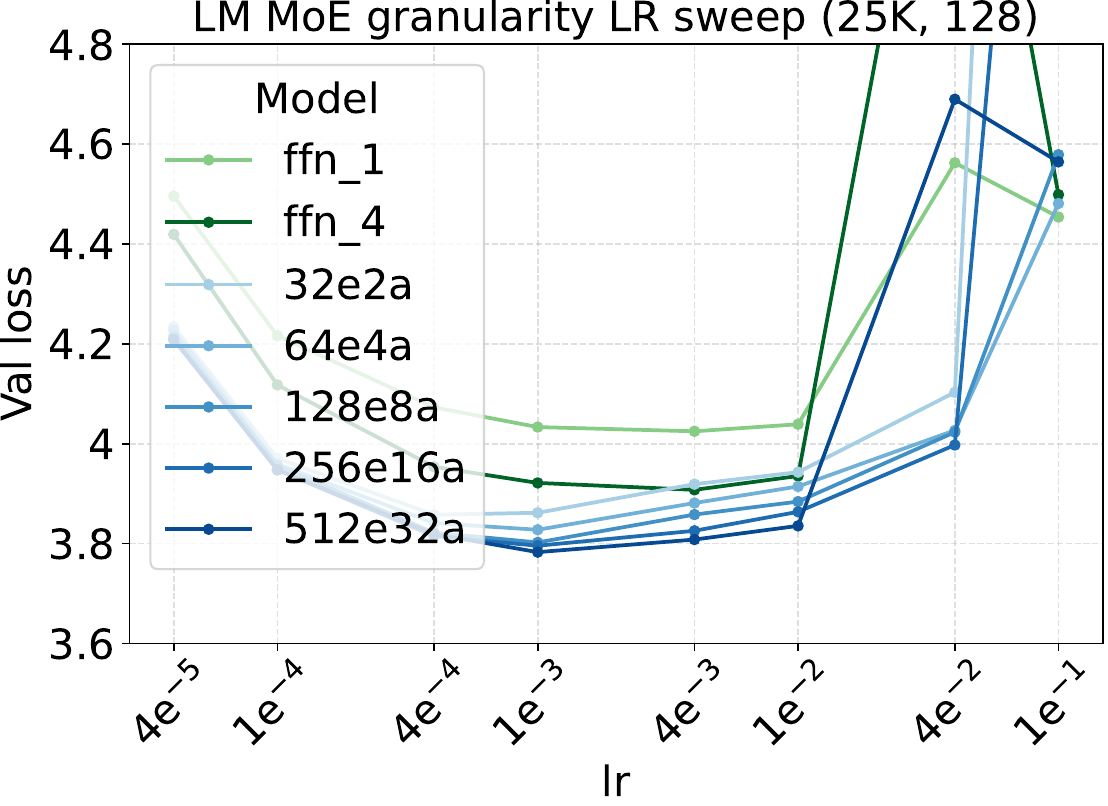}%
\hspace{0.006\linewidth}%
\includegraphics[width=0.497\linewidth]{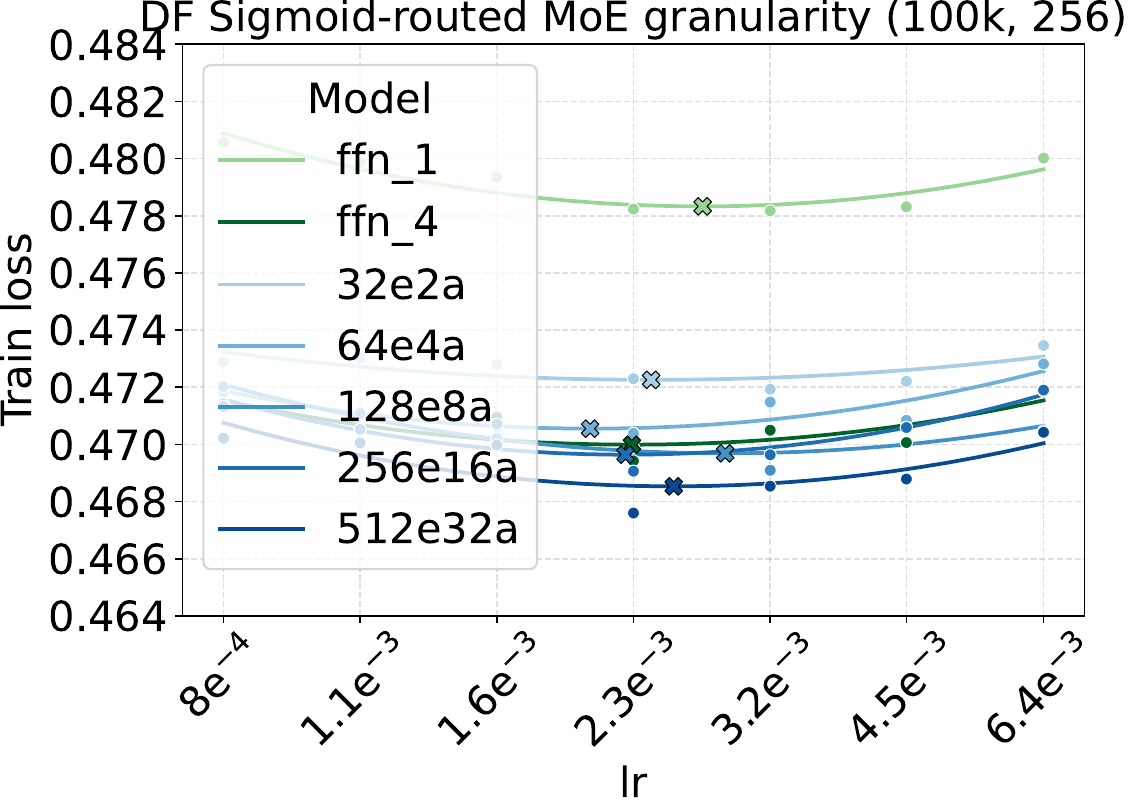}
\caption{MoE granularity learning-rate sweep}
\label{fig:moe-granularity-transfer}
\end{subfigure}
\par\medskip
\begin{subfigure}[t]{0.499\textwidth}
\centering
\includegraphics[width=0.497\linewidth]{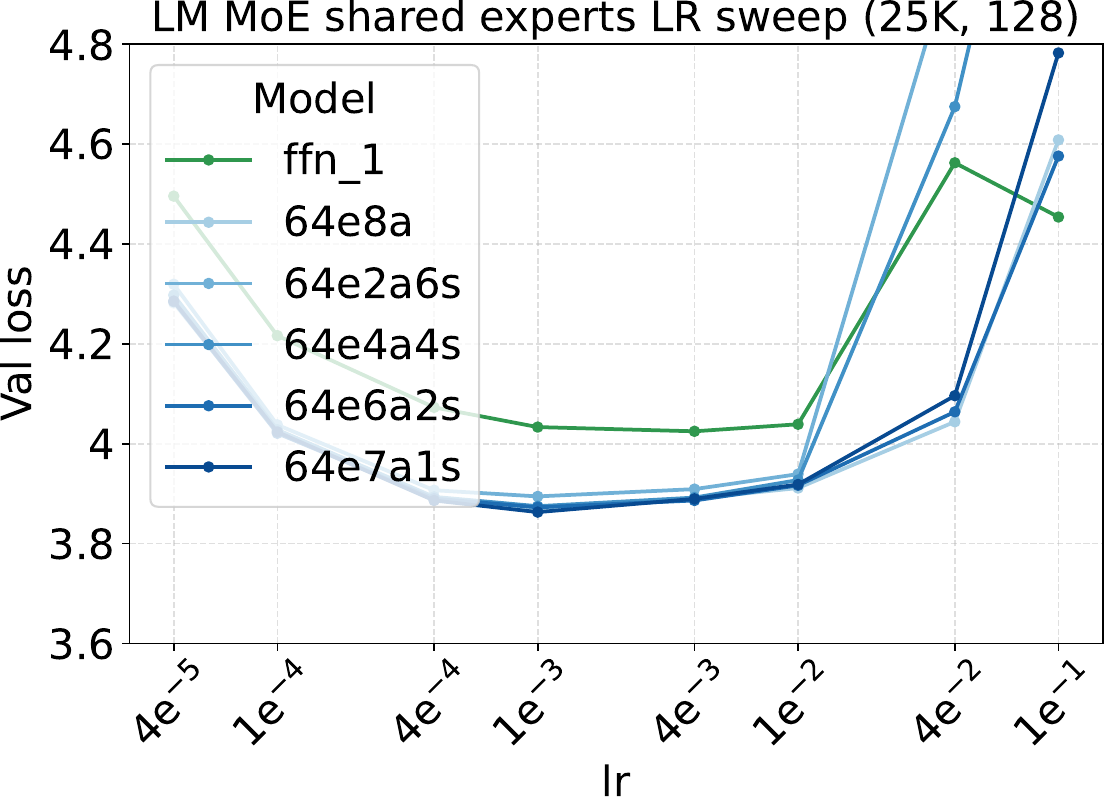}%
\hspace{0.006\linewidth}%
\includegraphics[width=0.497\linewidth]{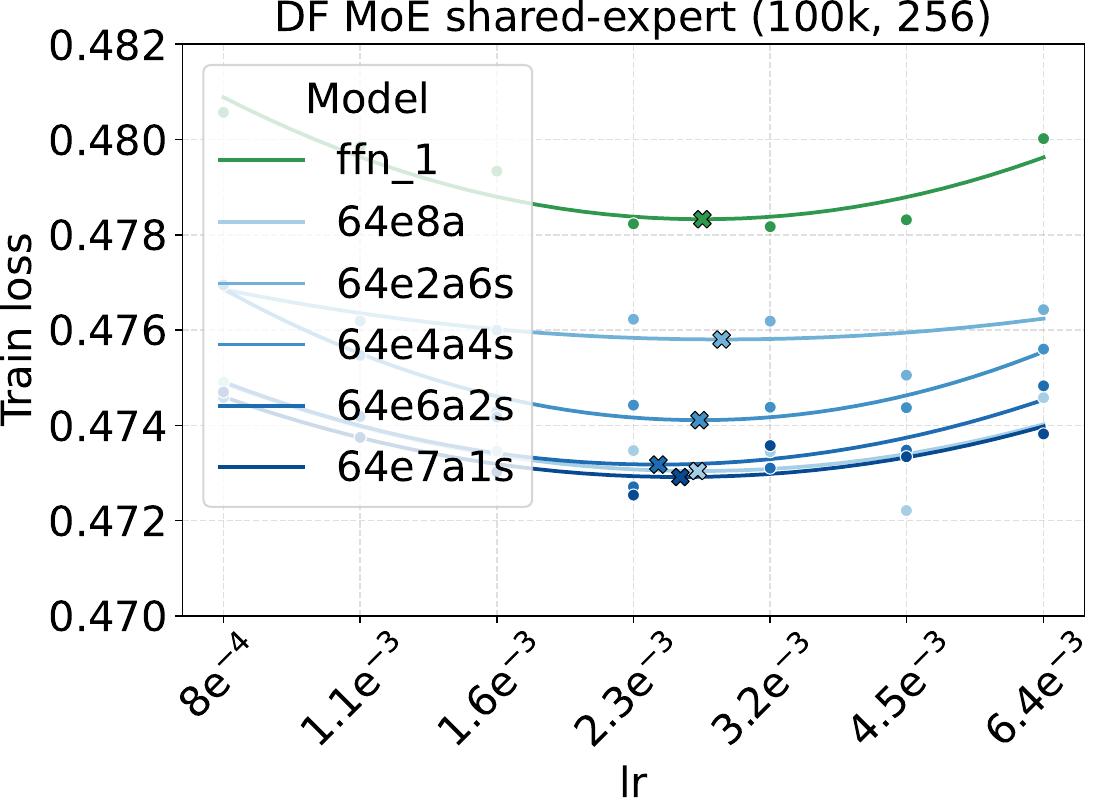}
\caption{Shared-expert MoE learning-rate sweep}
\label{fig:moe-shared-experts-transfer}
\end{subfigure}%
\hspace{0.002\textwidth}%
\begin{subfigure}[t]{0.499\textwidth}
\centering
\includegraphics[width=0.497\linewidth]{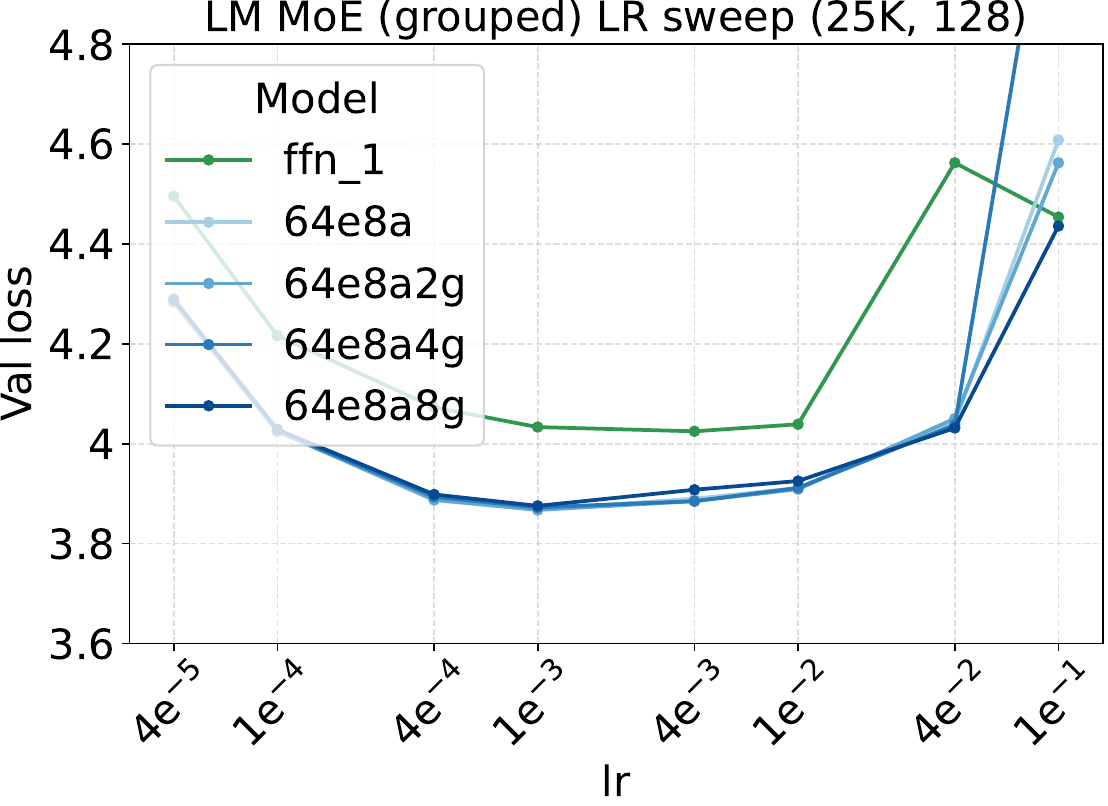}%
\hspace{0.006\linewidth}%
\includegraphics[width=0.497\linewidth]{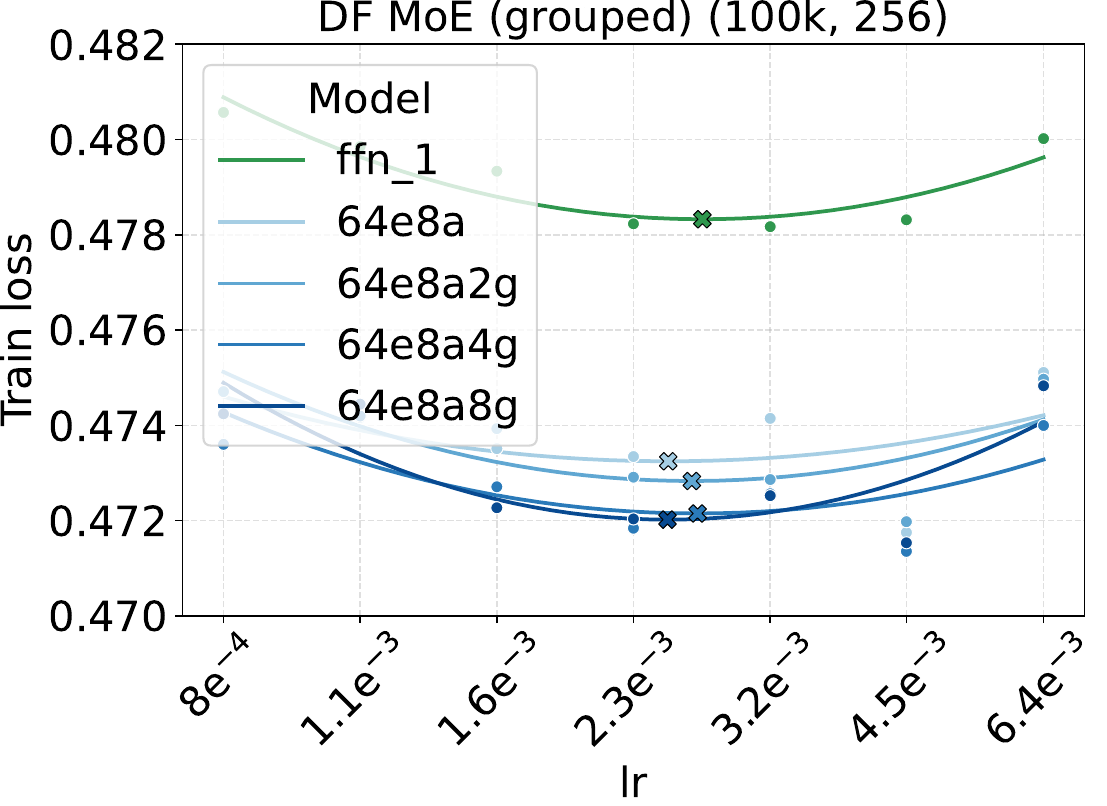}
\caption{Group-balanced-routing MoE learning-rate sweep}
\label{fig:moe-group-balanced-routing-transfer}
\end{subfigure}
\par\medskip
\begin{subfigure}[t]{0.499\textwidth}
\centering
\includegraphics[width=0.497\linewidth]{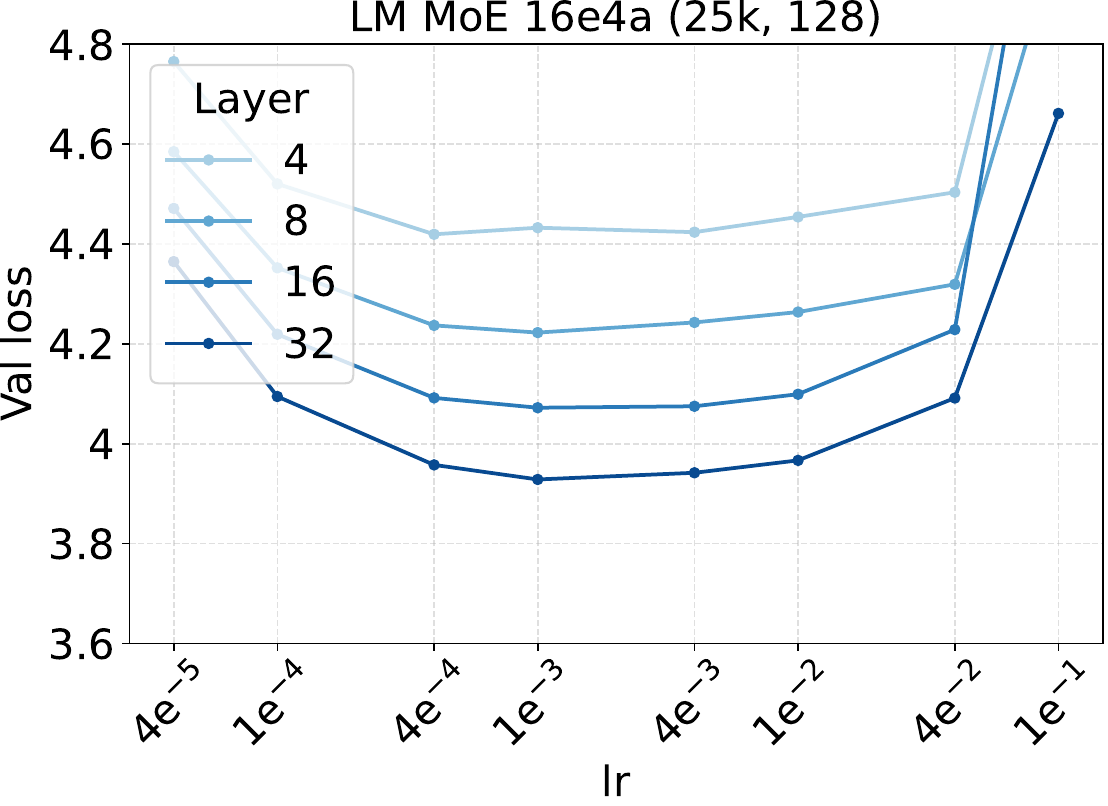}%
\hspace{0.006\linewidth}%
\includegraphics[width=0.497\linewidth]{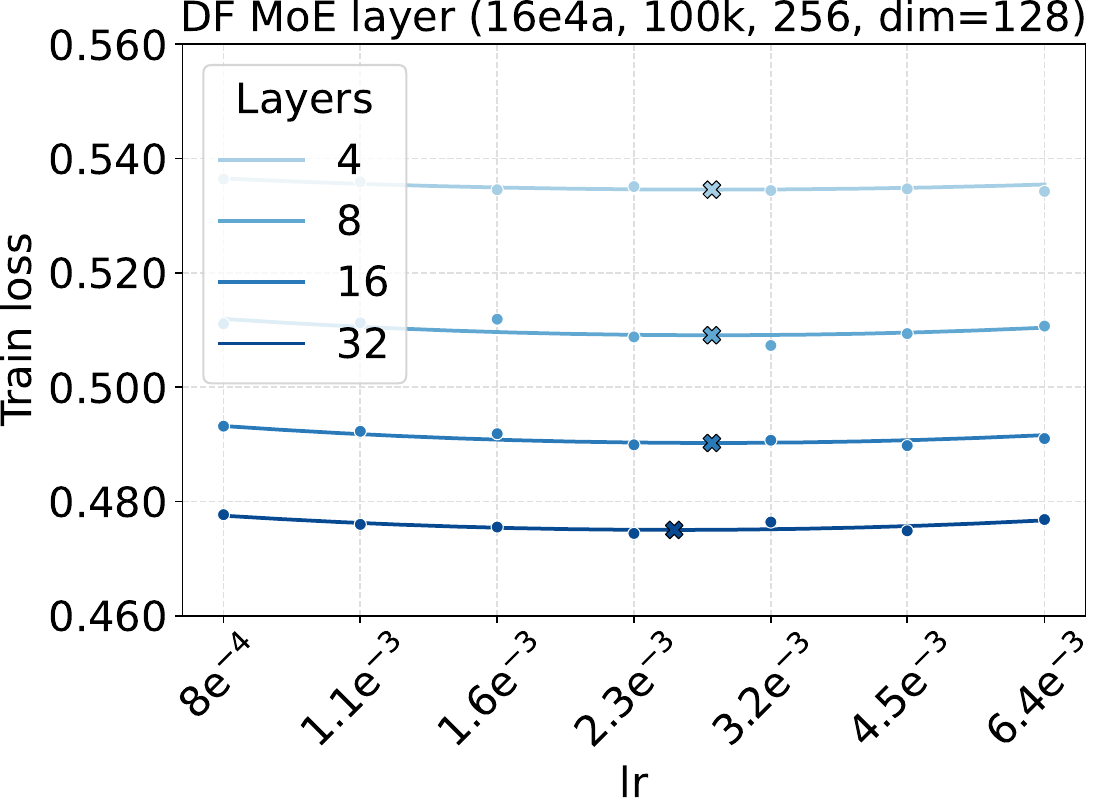}
\caption{MoE depth learning-rate sweep}
\label{fig:moe-depth-transfer}
\end{subfigure}%
\hspace{0.002\textwidth}%
\begin{subfigure}[t]{0.499\textwidth}
\centering
\includegraphics[width=0.497\linewidth]{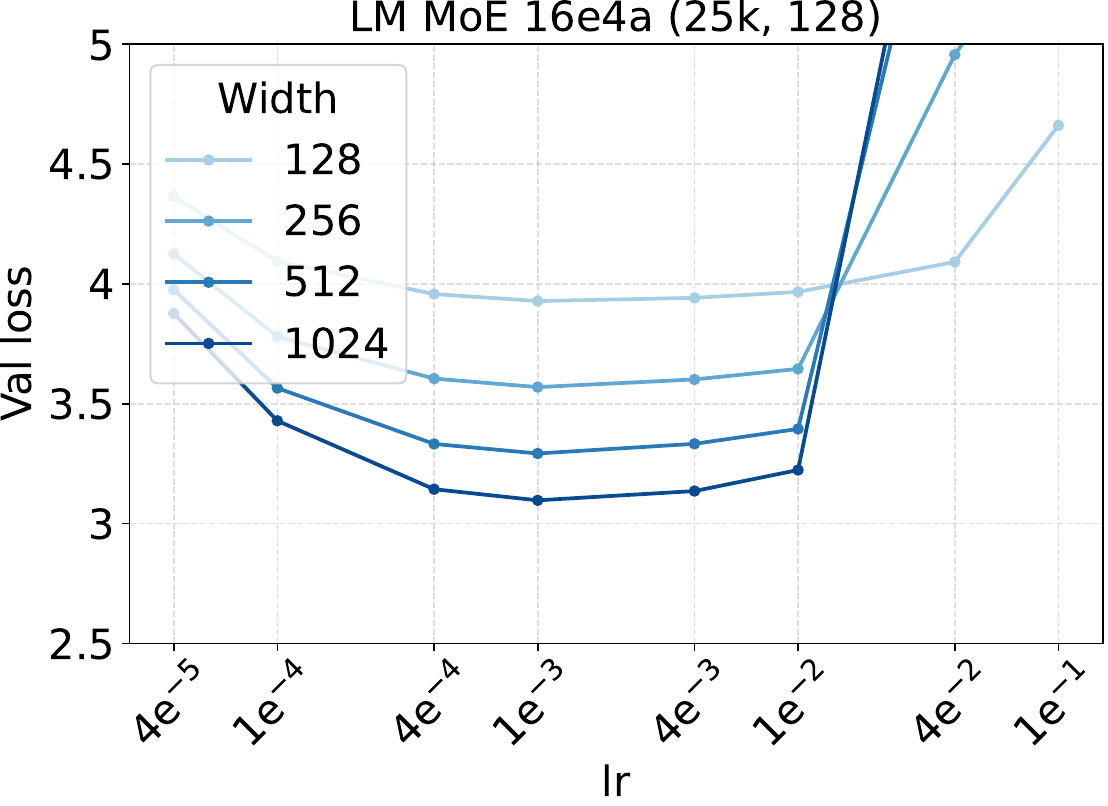}%
\hspace{0.006\linewidth}%
\includegraphics[width=0.497\linewidth]{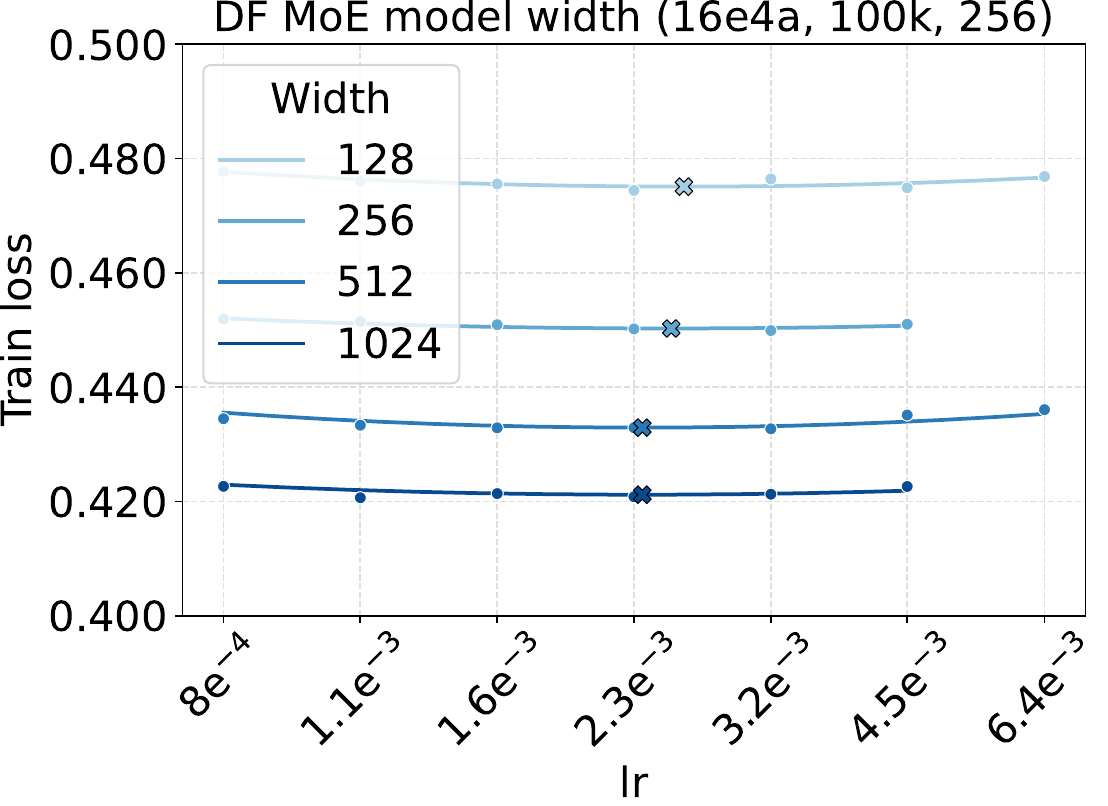}
\caption{MoE width learning-rate sweep}
\label{fig:moe-width-transfer}
\end{subfigure}
\caption{Transfer across MoE architectural variants. }
\label{fig:moe-architectural-transfer}
\end{figure}

\begin{figure}[t]
\centering
\begin{subfigure}[t]{0.499\textwidth}
\centering
\includegraphics[width=0.497\linewidth]{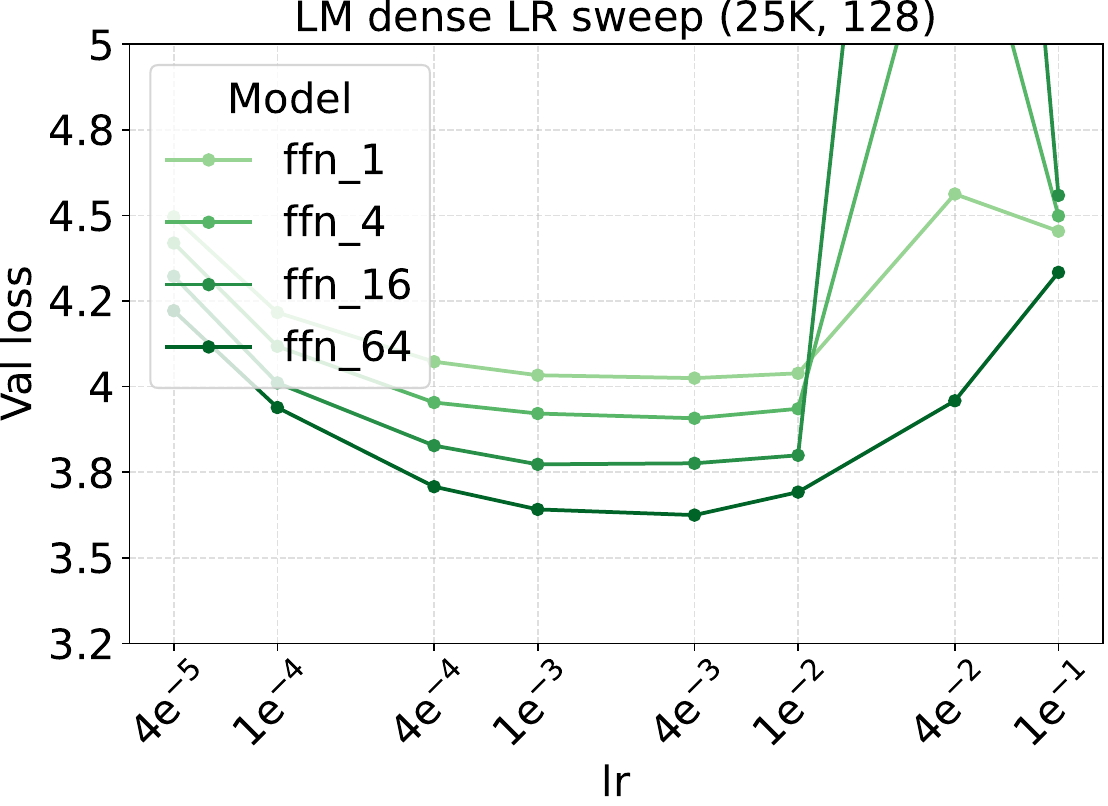}%
\hspace{0.006\linewidth}%
\includegraphics[width=0.497\linewidth]{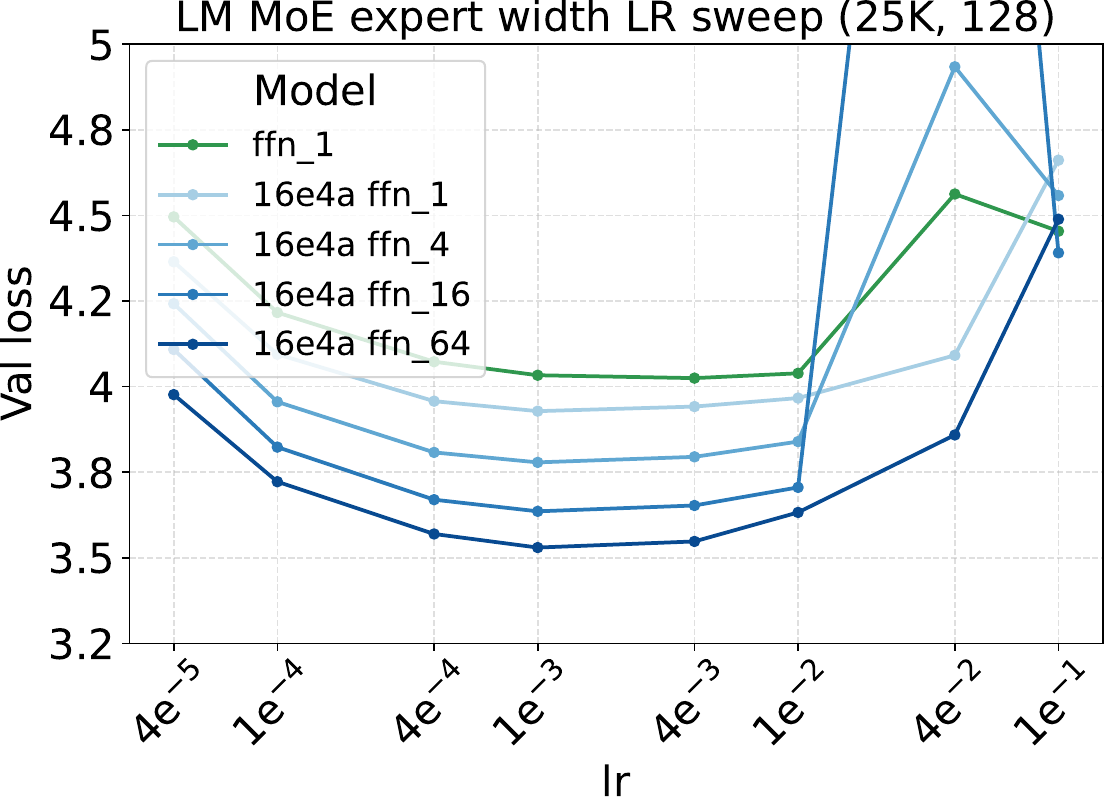}
\caption{FFN/MoE FFN width learning-rate sweep}
\label{fig:ffn-moe-ffn-width-expansion-lr}
\end{subfigure}%
\hspace{0.002\textwidth}%
\begin{subfigure}[t]{0.499\textwidth}
\centering
\includegraphics[width=0.497\linewidth]{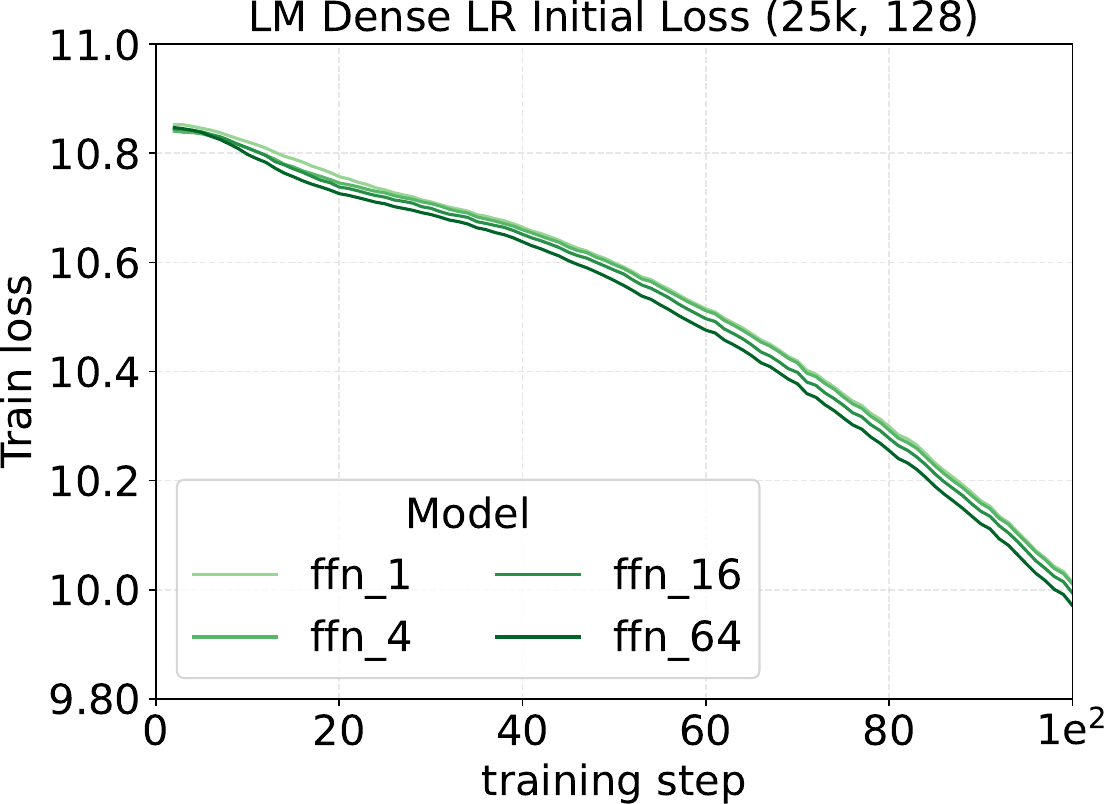}%
\hspace{0.006\linewidth}%
\includegraphics[width=0.497\linewidth]{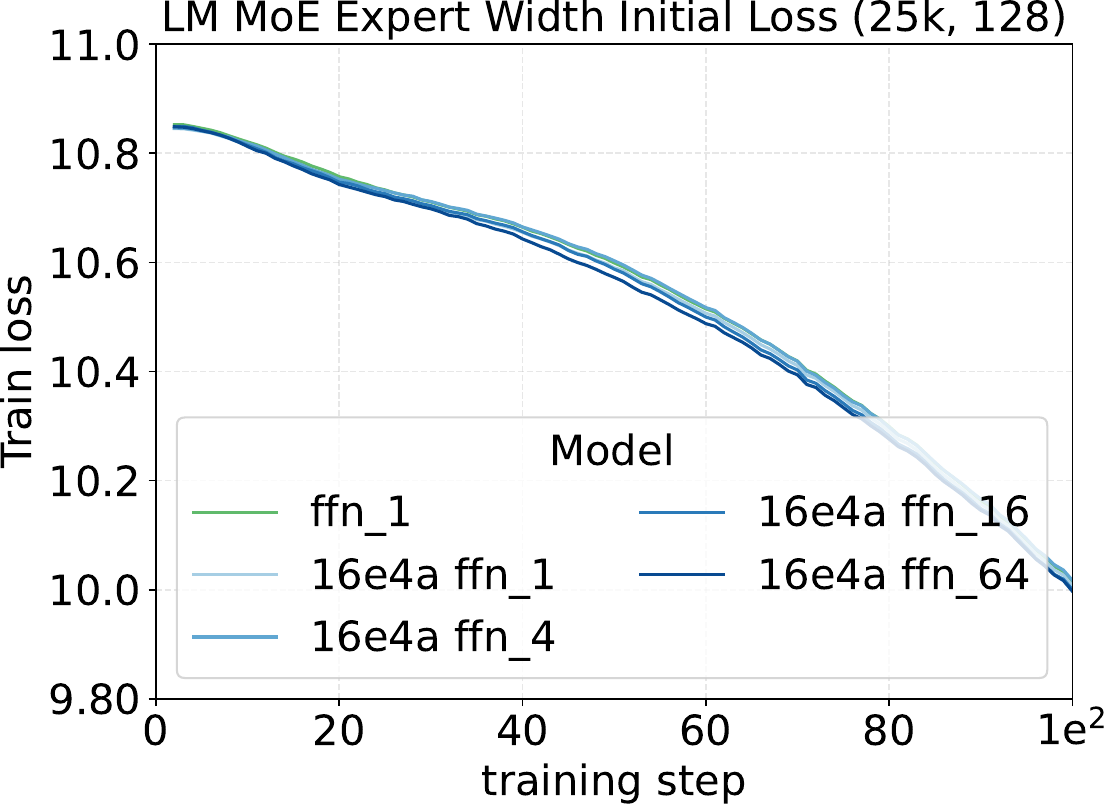}
\caption{FFN/MoE FFN width sweep initial training loss}
\label{fig:ffn-moe-ffn-width-expansion-initial-loss}
\end{subfigure}
\caption{Transfer across FFN/MoE FFN width expansion.}
\label{fig:ffn-moe-ffn-width-expansion-combined}
\end{figure}

Figure~\ref{fig:moe-architectural-transfer} and Figure~\ref{fig:ffn-moe-ffn-width-expansion-combined} covers the remaining axes with normalized sigmoid routing. Capacity, fixed-density granularity, shared experts, group-balanced routing, depth, width, and active FFN width expansion ratio both preserve similar LR optima. Capacity/granularity scaling both leads to lower loss. Increasing the number of shared experts consistently leads to higher loss, suggesting that shared experts are not necessary for performance unless they are used to overlap computation and communication in expert-parallel training/inference.

\begin{figure}[t]
\centering
\begin{subfigure}[t]{0.499\textwidth}
\centering
\includegraphics[width=0.497\linewidth]{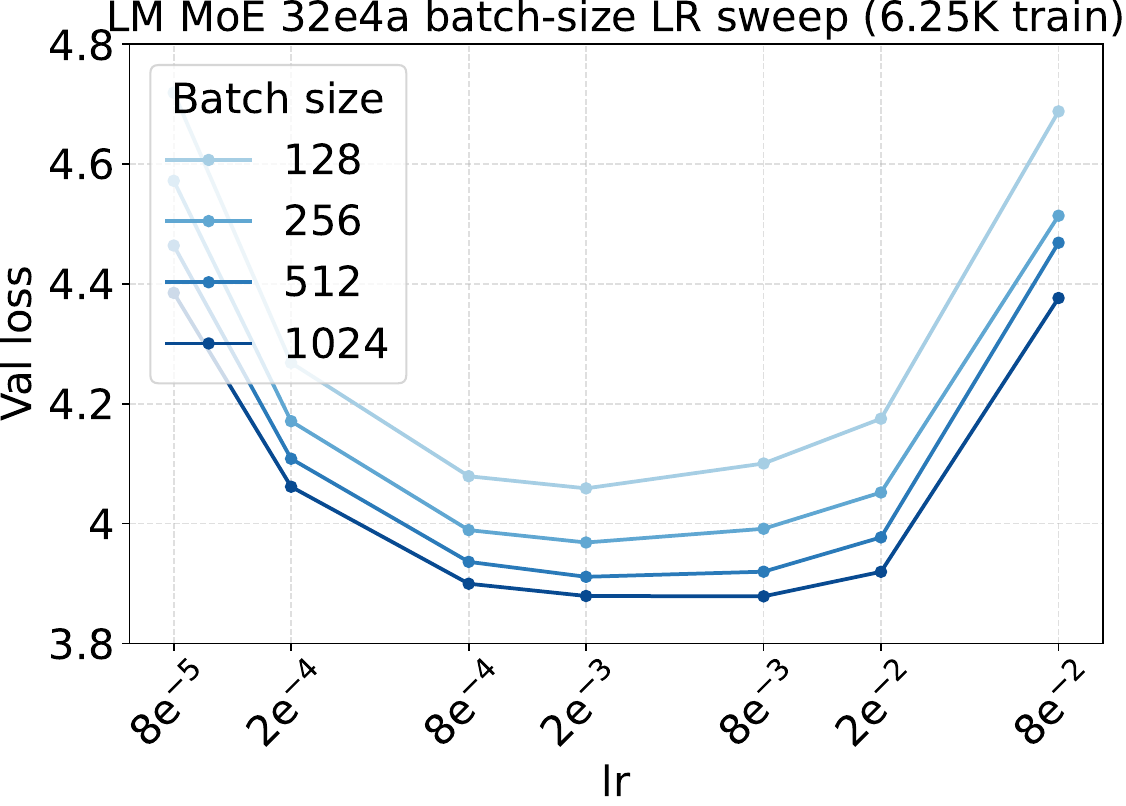}%
\hspace{0.006\linewidth}%
\includegraphics[width=0.497\linewidth]{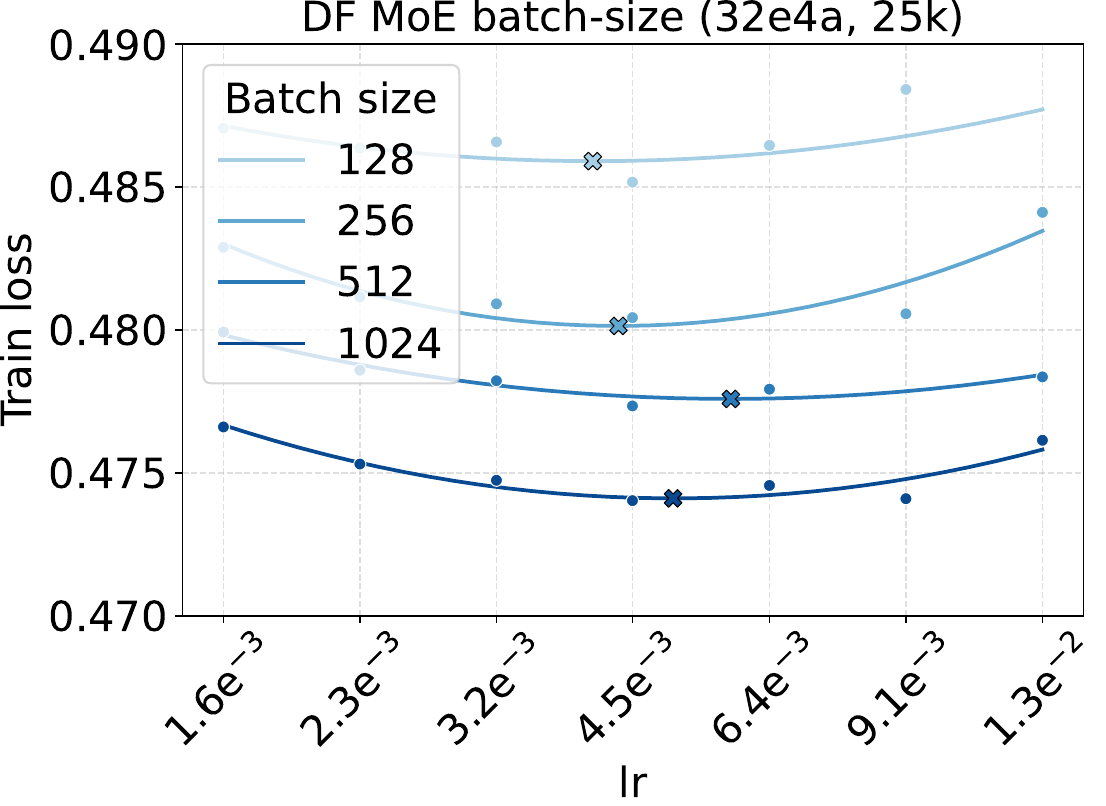}
\caption{Batch-size study at fixed training iterations}
\label{fig:batch-size-transfer-fixed-iterations}
\end{subfigure}%
\hspace{0.002\textwidth}%
\begin{subfigure}[t]{0.499\textwidth}
\centering
\includegraphics[width=0.497\linewidth]{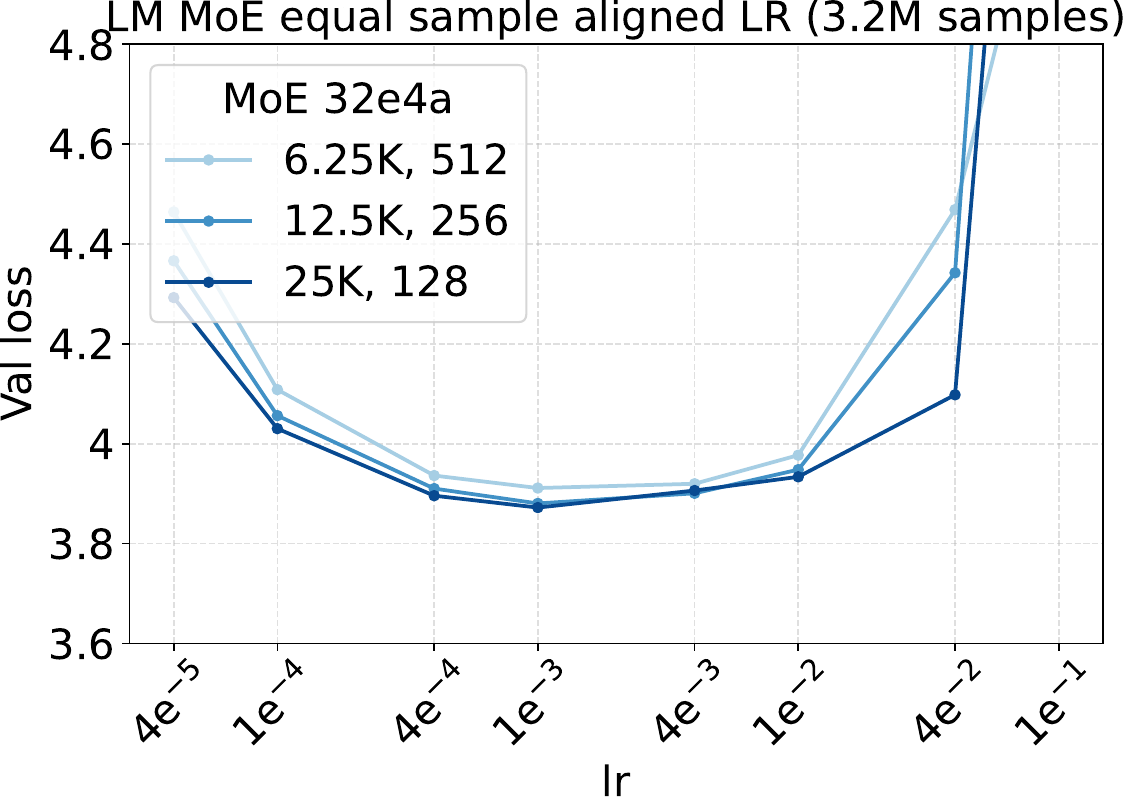}%
\hspace{0.006\linewidth}%
\includegraphics[width=0.497\linewidth]{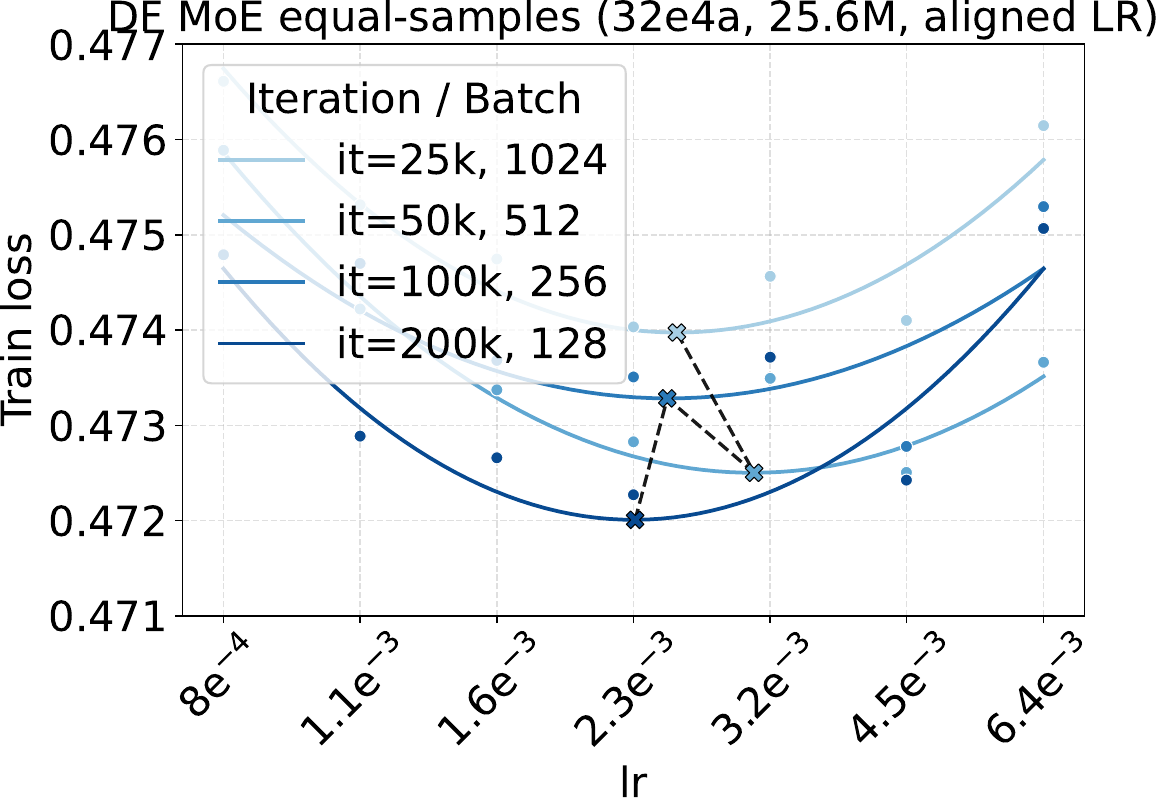}
\caption{Batch-size study at fixed total trained tokens}
\label{fig:batch-size-transfer-fixed-tokens}
\end{subfigure}
\caption{Transfer across batch size. }
\label{fig:batch-size-transfer-combined}
\end{figure}
Figure~\ref{fig:batch-size-transfer-combined} shows that batch/duration transfer composes with the MoE rules. For batch sizes $128$--$1024$, optima stay close at fixed steps ($6.25$k LM, $25$k diffusion) and fixed samples ($3.2$M LM, $25.6$M diffusion); at fixed samples, the learning rate in the figure is aligned and scaled according to $\rho_D=1$ and $\eta\propto\sqrt{\rho_B}$ (an exact SDE transfer with all three SDE objects preserved), while at fixed steps the transfer is approximate---the optimal LR region remains relatively stable across batch sizes with only minor drift, consistent with the $\sigma_0$ shift at fixed horizon.

\begin{figure}[t]
\centering
\begin{subfigure}[t]{0.245\textwidth}
\centering
\includegraphics[width=\linewidth]{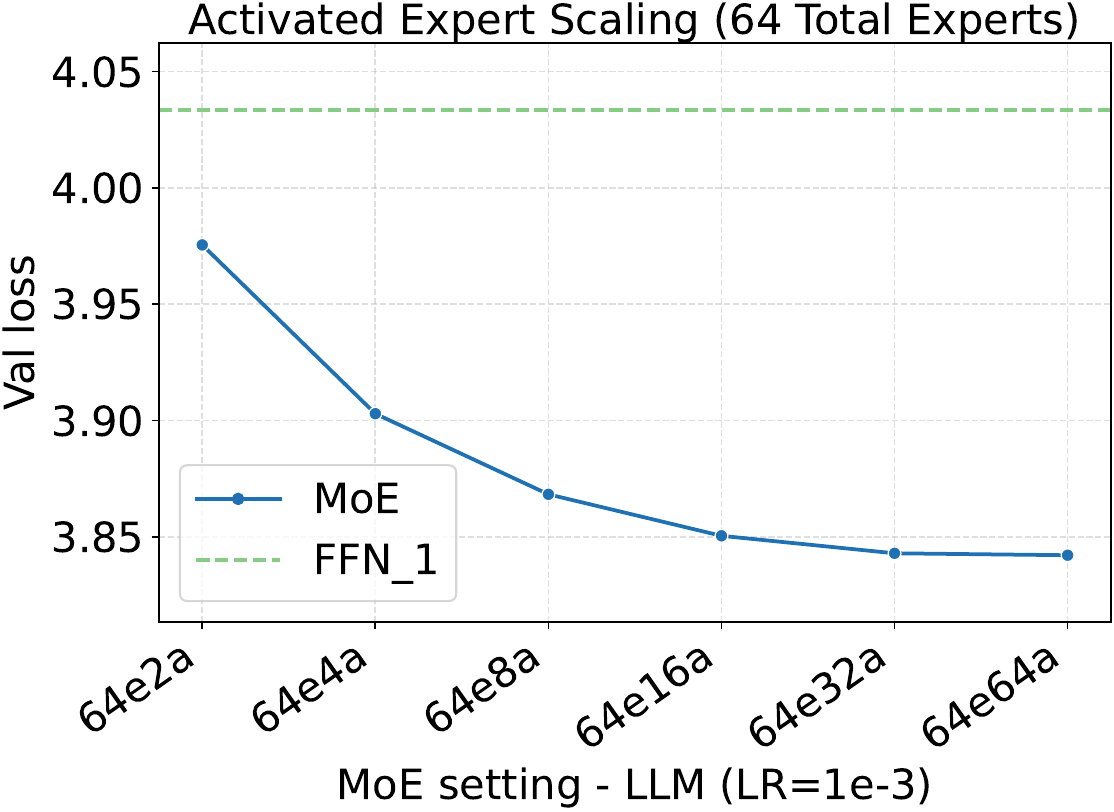}
\caption{LM, activated experts}
\label{fig:fixed-lr-llm-activated}
\end{subfigure}%
\hfill
\begin{subfigure}[t]{0.245\textwidth}
\centering
\includegraphics[width=\linewidth]{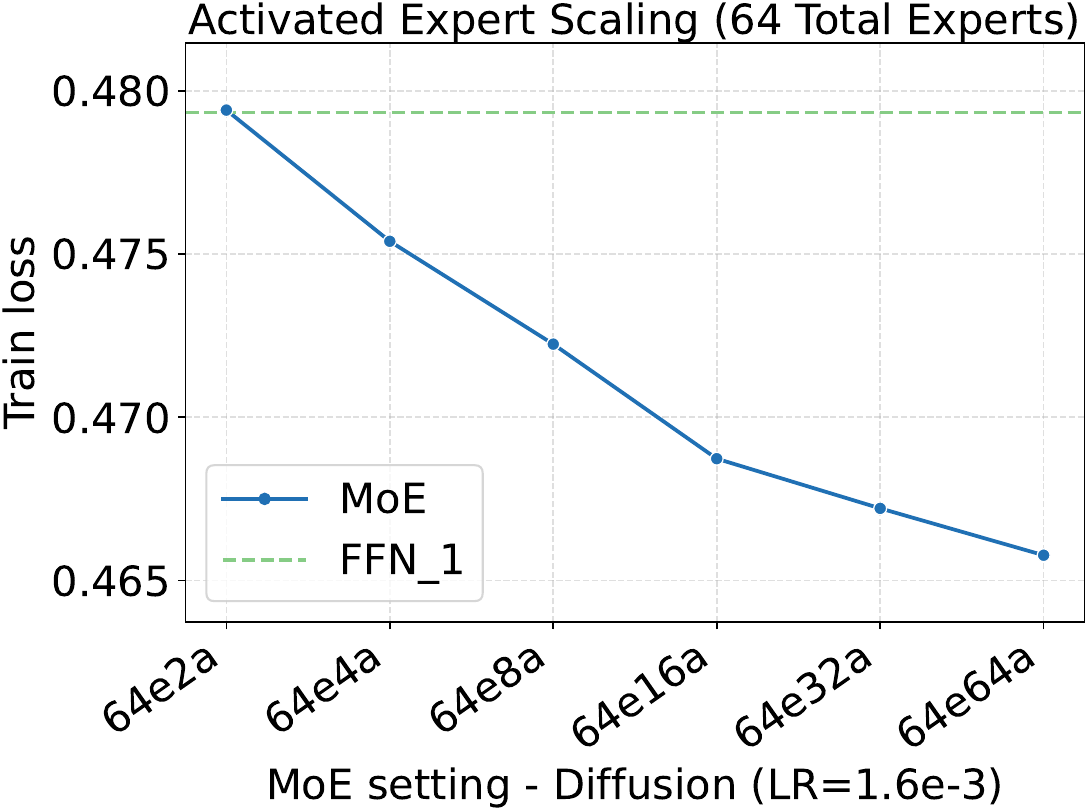}
\caption{DF, activated experts}
\label{fig:fixed-lr-diffusion-activated}
\end{subfigure}%
\hfill
\begin{subfigure}[t]{0.245\textwidth}
\centering
\includegraphics[width=\linewidth]{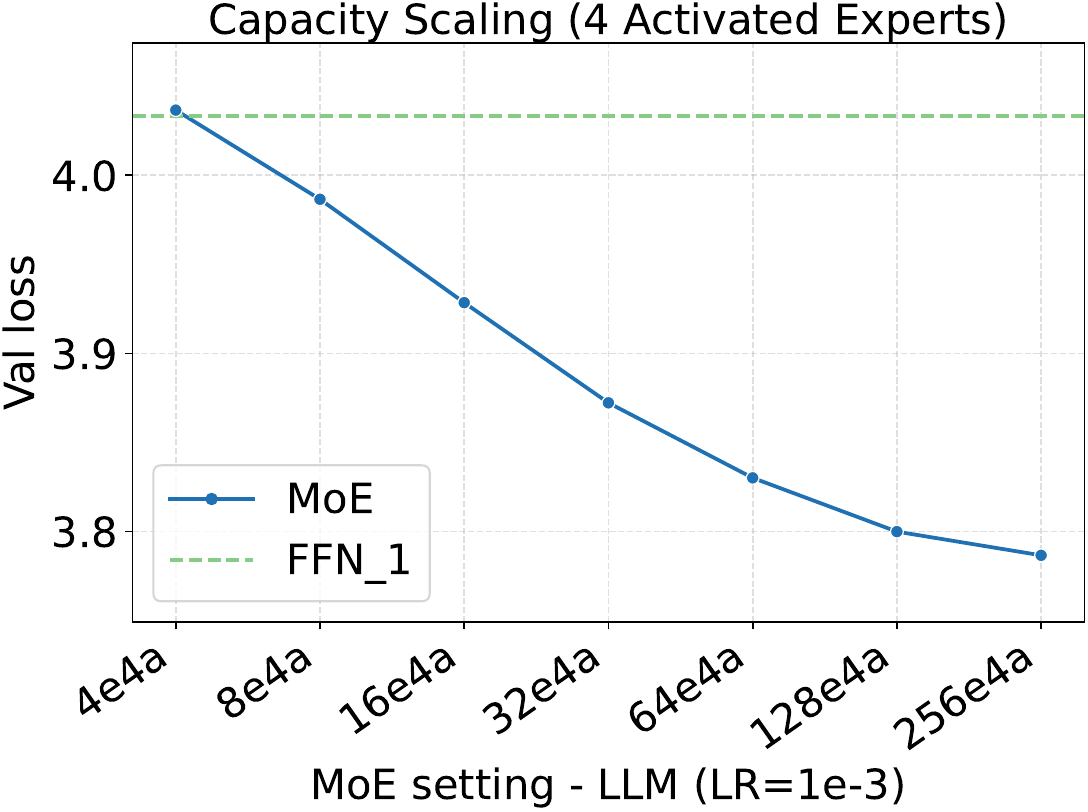}
\caption{LM, capacity}
\label{fig:fixed-lr-llm-capacity}
\end{subfigure}%
\hfill
\begin{subfigure}[t]{0.245\textwidth}
\centering
\includegraphics[width=\linewidth]{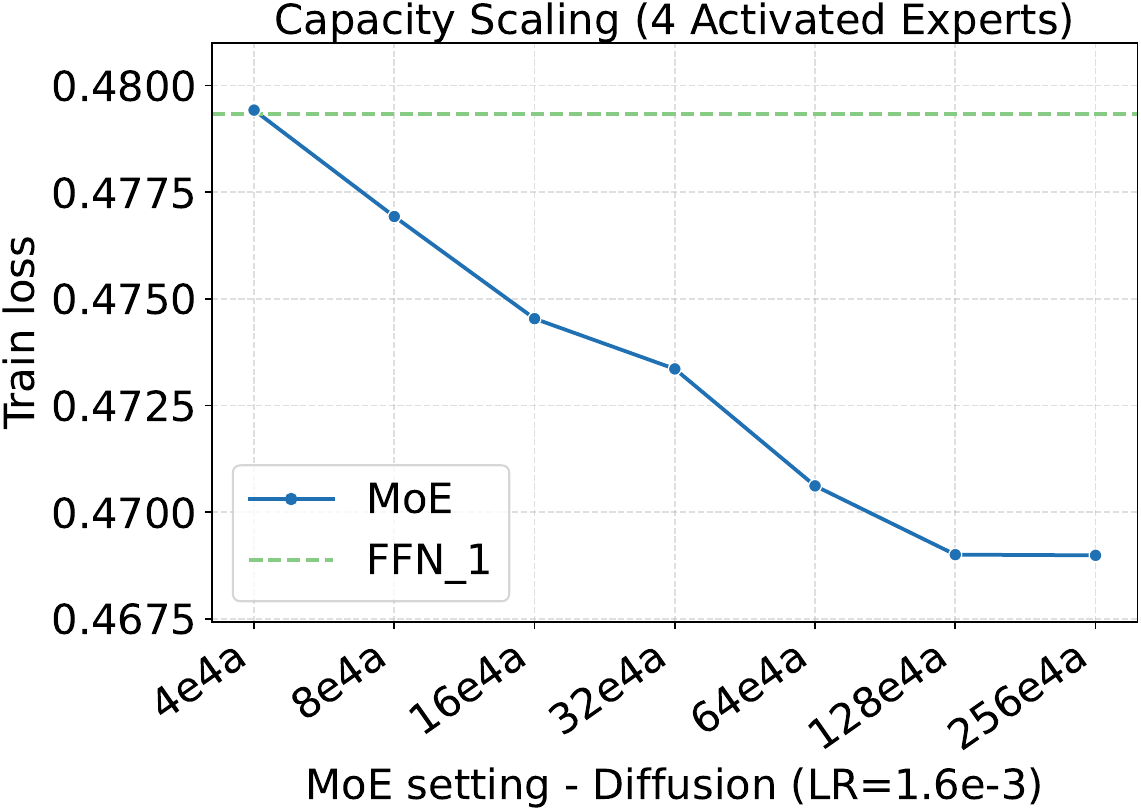}
\caption{DF, capacity}
\label{fig:fixed-lr-diffusion-capacity}
\end{subfigure}
\par\medskip
\begin{subfigure}[t]{0.245\textwidth}
\centering
\includegraphics[width=\linewidth]{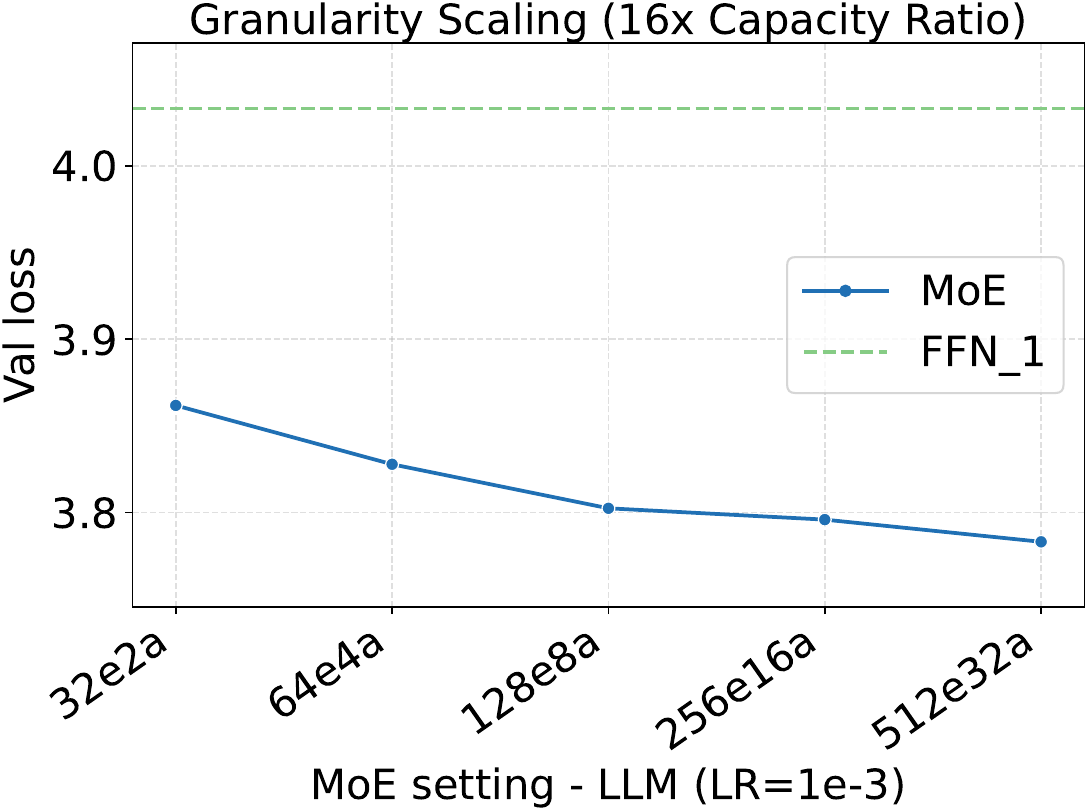}
\caption{LM, granularity}
\label{fig:fixed-lr-llm-granularity}
\end{subfigure}%
\hfill
\begin{subfigure}[t]{0.245\textwidth}
\centering
\includegraphics[width=\linewidth]{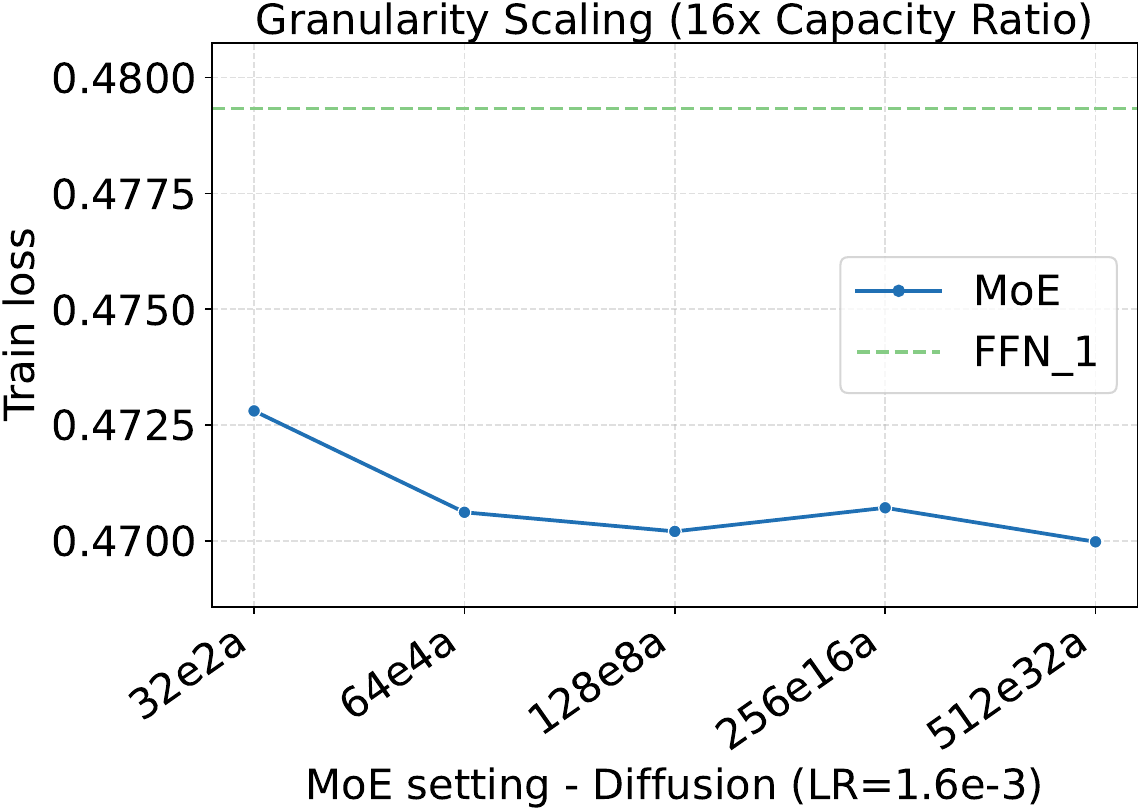}
\caption{DF, granularity}
\label{fig:fixed-lr-diffusion-granularity}
\end{subfigure}%
\hfill
\begin{subfigure}[t]{0.245\textwidth}
\centering
\includegraphics[width=\linewidth]{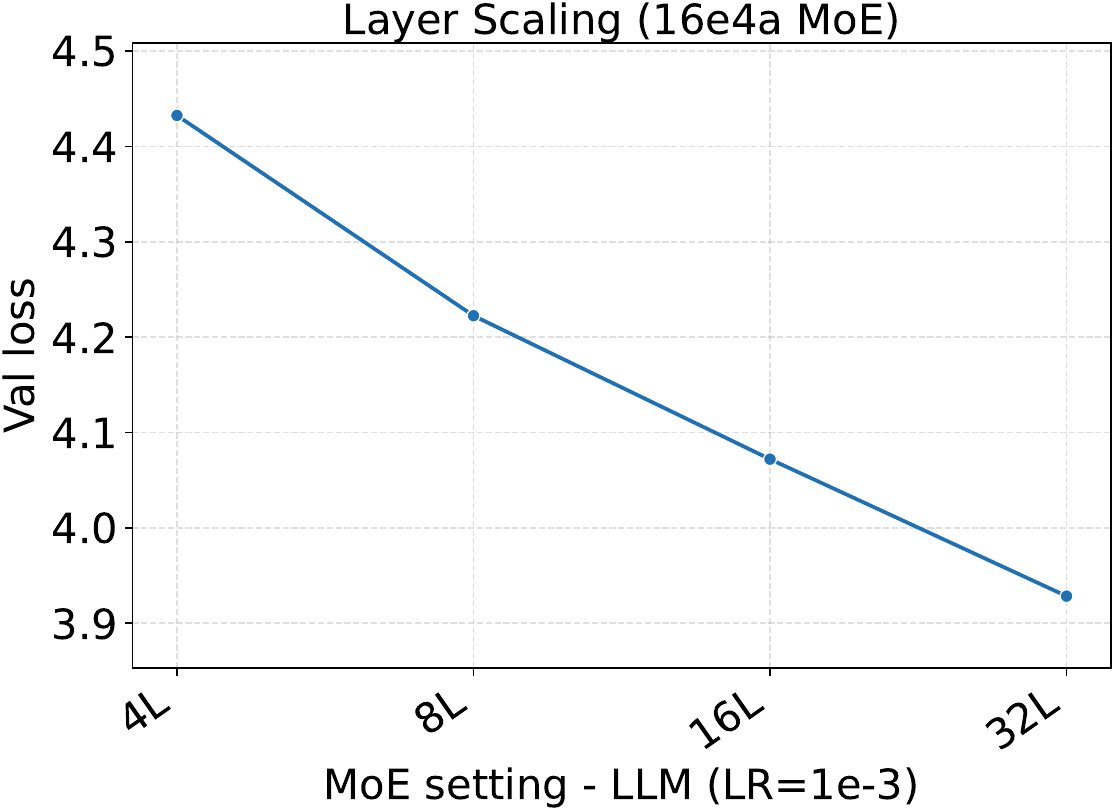}
\caption{LM, layers}
\label{fig:fixed-lr-llm-layers}
\end{subfigure}%
\hfill
\begin{subfigure}[t]{0.245\textwidth}
\centering
\includegraphics[width=\linewidth]{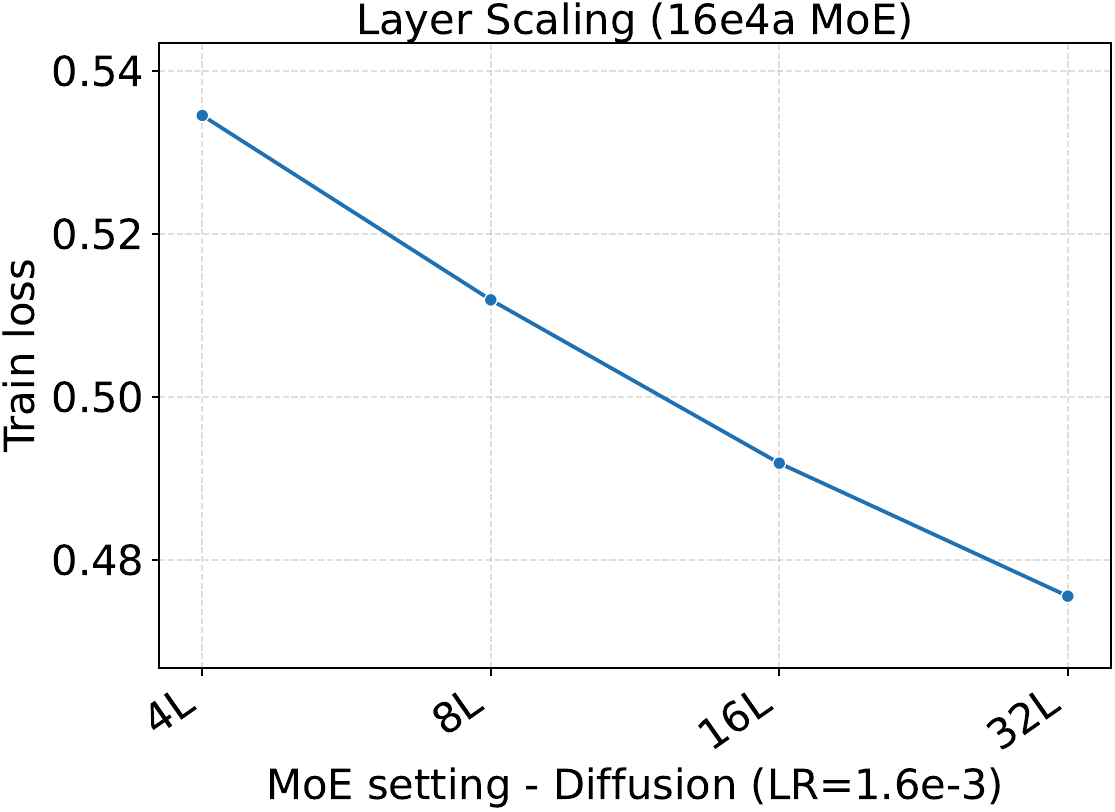}
\caption{DF, layers}
\label{fig:fixed-lr-diffusion-layers}
\end{subfigure}
\caption{Fixed-hyperparameter loss scaling across MoE axes. We fix AdamW hyperparameters at the dense-tuned values (LM: LR=$10^{-3}$, init std=$10^{-2}$, WD=$0.1$; DF: LR=$1.6\times10^{-3}$, init std=$2\times10^{-2}$, WD=$10^{-2}$) and vary the MoE architecture along four axes for both LM and DF. With Complete-muE, each axis produces consistent loss improvement without per-setting hyperparameter retuning---direct evidence for the tune-dense-once-and-transfer recipe.}
\label{fig:fixed-hp-loss-scaling}
\end{figure}

Figure~\ref{fig:fixed-hp-loss-scaling} provides the direct empirical evidence for the tune-dense-once-and-transfer recipe. Fixing AdamW hyperparameters at the dense-tuned values (LM: LR=$10^{-3}$, init std=$10^{-2}$, WD=$0.1$; DF: LR=$1.6\times10^{-3}$, init std=$2\times10^{-2}$, WD=$10^{-2}$) and varying only the MoE architecture along four axes---activated experts, total capacity, granularity, and depth---yields consistent loss reduction for both LM and DF without per-setting retuning. Combined with the LR-sweep evidence above, this shows that Complete-muE's bounded residual drift is small enough in practice for a single dense calibration to deliver consistent scaling gains across the entire MoE design space.

\subsection{Systems benchmark for MoE scaling axes}
\label{sec:exp-system-benchmark}

\begin{wrapfigure}[12]{r}{0.55\textwidth}
\centering
\begin{subfigure}[t]{0.485\linewidth}
\centering
\includegraphics[width=\linewidth]{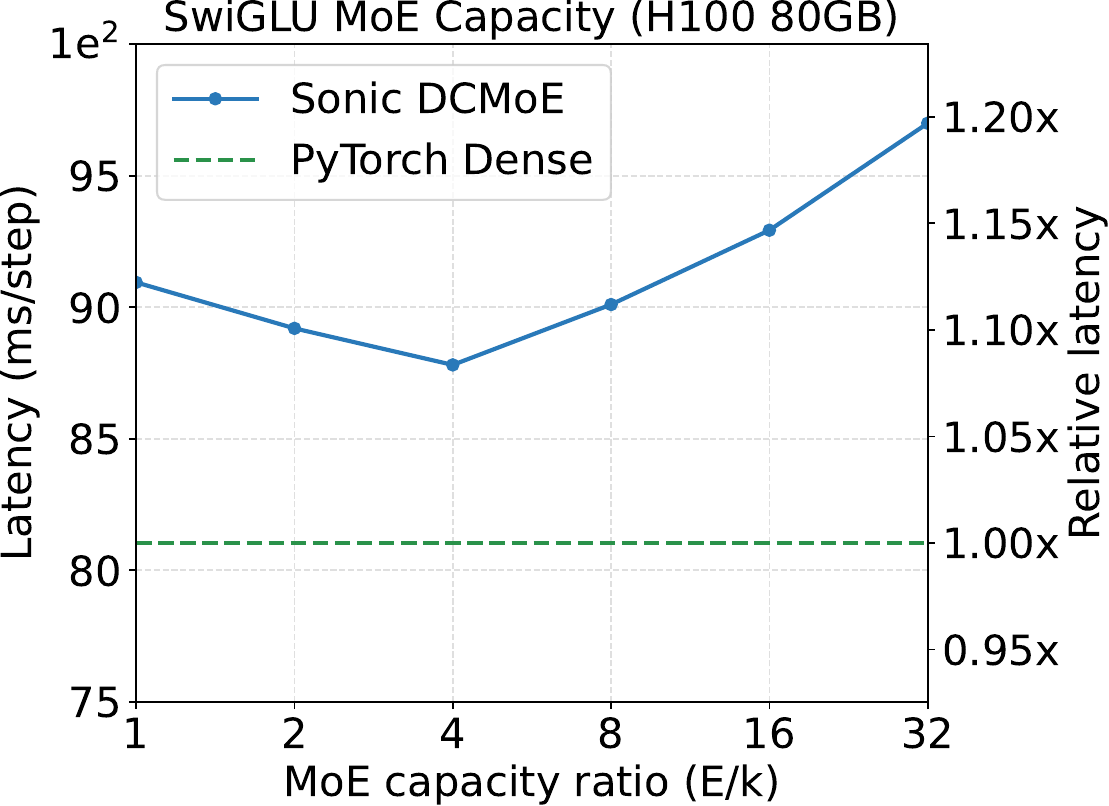}
\caption{Capacity scaling}
\label{fig:benchmark-moe-capacity-scaling}
\end{subfigure}%
\hfill
\begin{subfigure}[t]{0.485\linewidth}
\centering
\includegraphics[width=\linewidth]{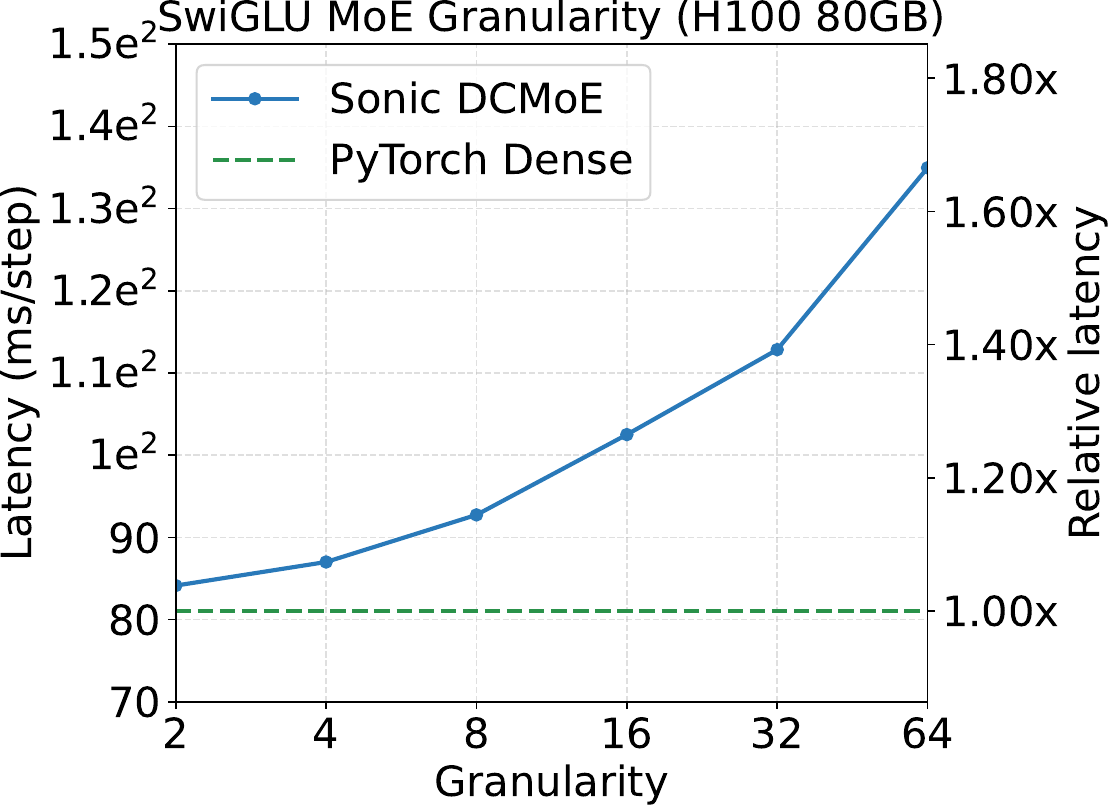}
\caption{Granularity scaling}
\label{fig:benchmark-moe-granularity-scaling}
\end{subfigure}
\caption{SwiGLU MoE latency benchmark on a single H100 80GB GPU.}
\label{fig:benchmark-moe-speed}
\end{wrapfigure}

The preceding sweeps show two quality-improving MoE axes---more total capacity at fixed activated experts, and finer granularity at fixed active width. Figure~\ref{fig:benchmark-moe-speed} compares their training latency on a single H100 80GB GPU using bfloat16, sequence length $40$k, input width $4096$, and Sonic DCMoE, our internal MoE implementation following SonicMoE-style design \cite{guo2025sonicmoe}.

Capacity scaling fixes expert hidden size $2048$ and $k=8$ activated experts while varying $E=8$--$256$; latency stays at $87.8$--$97.0$ ms/step, only $1.08\times$--$1.20\times$ slower than the $81.0$ ms dense SwiGLU baseline. Granularity scaling fixes active width $h_{\mathrm{dim}}k=16{,}384$ and $E/k=8$ while increasing $k=2$--$64$; latency rises from $84.1$ to $135.0$ ms/step. Dense SwiGLU width scaling is far more expensive, from $81.0$ to $667.6$ ms/step before larger settings run out of memory. Thus capacity scaling is the cheaper axis when token batch size per device is large enough.

\subsection{Larger-scale Complete-muE experiments}
\label{sec:exp-large-scale}
We apply the transferred recipe to larger runs. The multimodal diffusion setting trains for $100$k iterations on $256$P images (batch $3072$), $512$P images (batch $768$), $240$P key frames (batch $96$), and $240$P $5$s videos (batch $48$), using Knapformer workload balancing \cite{zhang2025knapformer}, per-layer AdaLN, and 3D RoPE. The large-scale LM uses 1D RoPE, and 512 batch size (2K seq length), trained for $100$k iterations. Unless noted, models have $32$ layers, width $1024$, and MoE configuration \texttt{128e8a4g1s} with per-expert width $512$ and total active FFN width $4608$. Dense and MoE models have about $0.62$B active non-embedding parameters; MoE variants have $6.29$B total non-embedding parameters.

Both settings use a WSD schedule with $100$k total iterations, $1{,}000$ warmup steps, and a $15\%$ decay phase. For diffusion training, the 3D RoPE uses channel splits $[48,\,48,\,24]$; AdamW is configured with learning rate $2.26\times10^{-3}$, $\beta_1{=}\beta_2{=}0.95$, and weight decay $0.01$. For LLM training, AdamW uses learning rate $5\times10^{-4}$, $\beta_1{=}\beta_2{=}0.95$, and weight decay $0.05$; LLM validation is performed on C4~\citep{dodge2021documenting,raffel2020exploring}. In both cases the target backbone width $d{=}1024$ is $8{\times}$ the proxy width $d_{\star}{=}128$, giving the Complete-muE width-expansion ratio $\rho_d{=}8$.

\begin{figure}[t]
\centering
\begin{subfigure}[t]{0.322\textwidth}
\centering
\includegraphics[width=\linewidth]{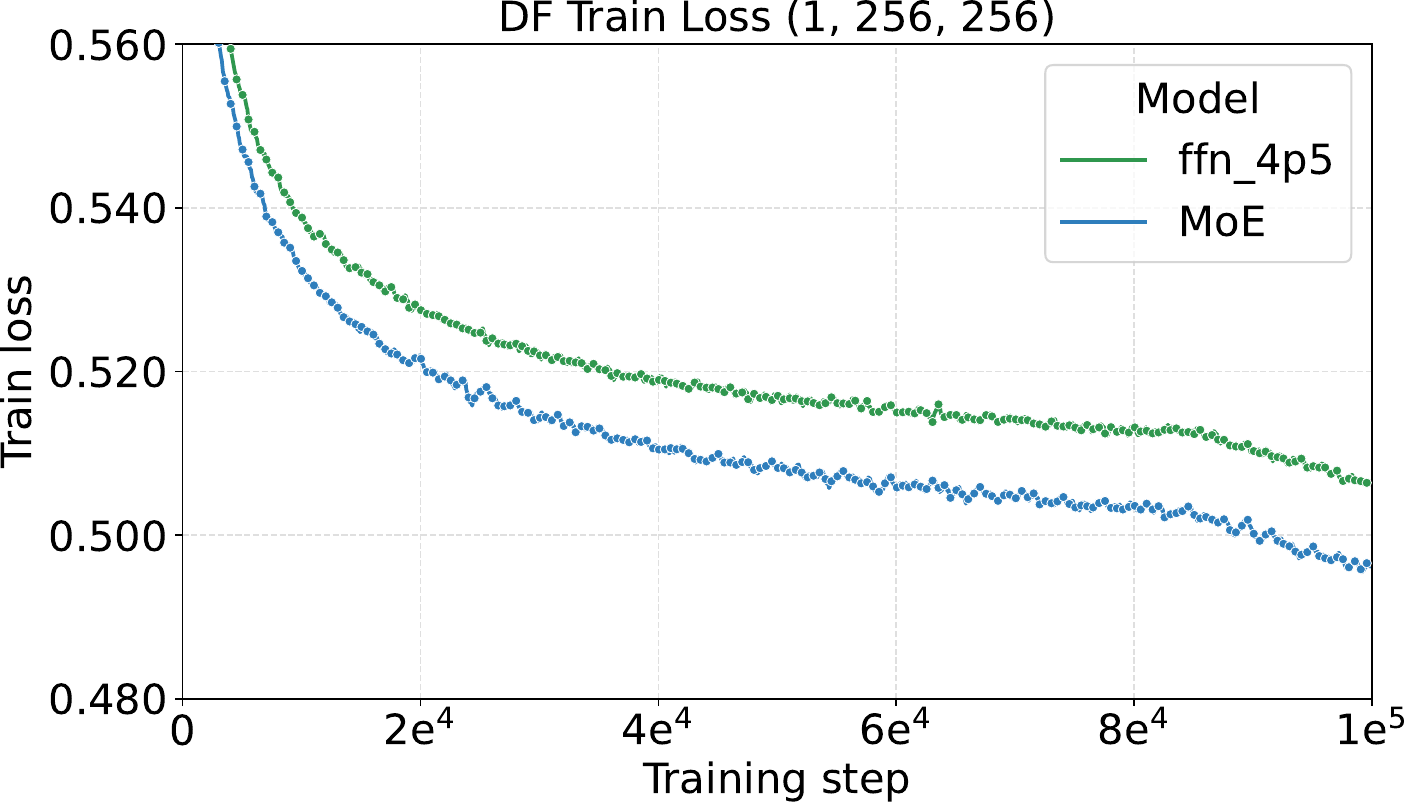}
\caption{256P image training loss}
\label{fig:large-run-256p-image-loss}
\end{subfigure}%
\hspace{0.012\textwidth}%
\begin{subfigure}[t]{0.322\textwidth}
\centering
\includegraphics[width=\linewidth]{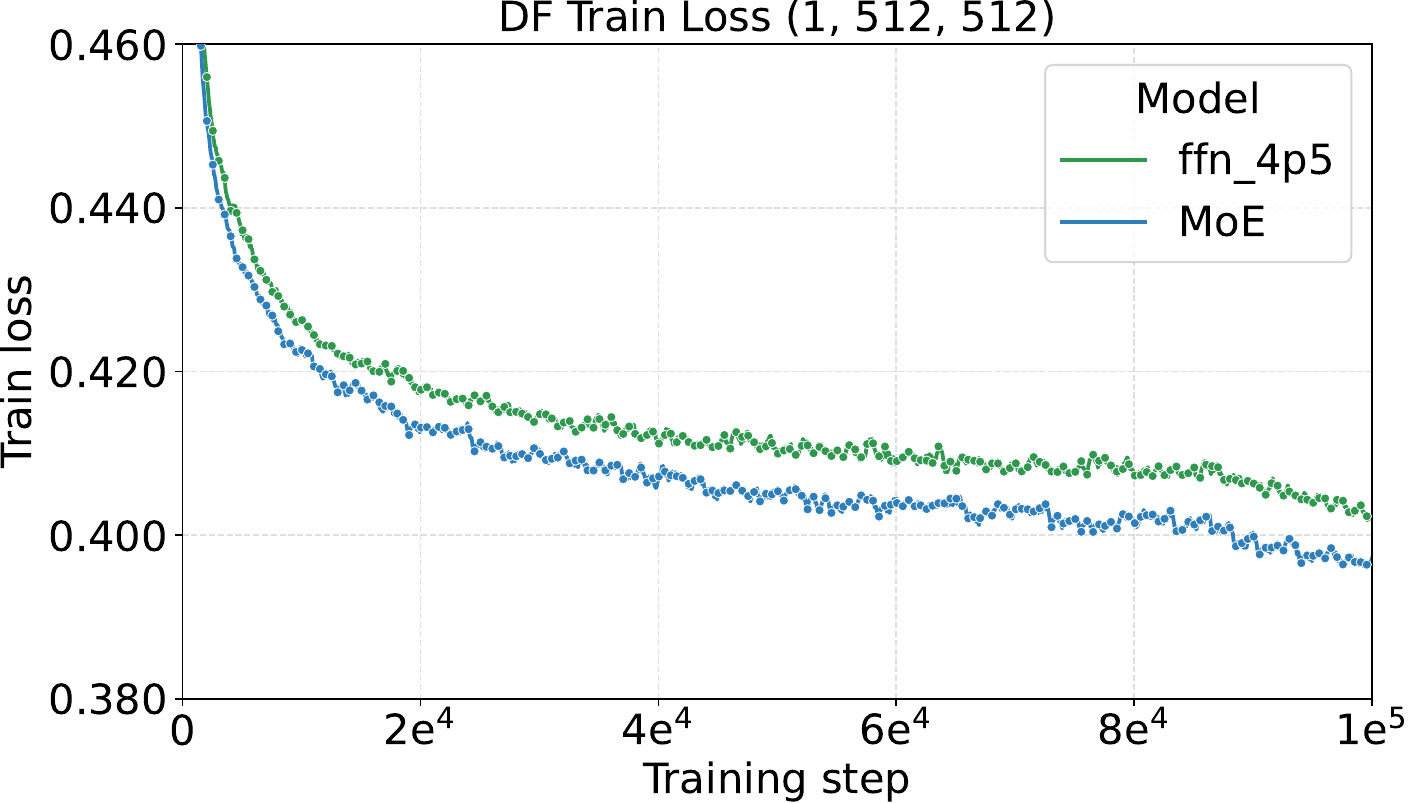}
\caption{512P image training loss}
\label{fig:large-run-512p-image-loss}
\end{subfigure}%
\hspace{0.012\textwidth}%
\begin{subfigure}[t]{0.322\textwidth}
\centering
\includegraphics[width=\linewidth]{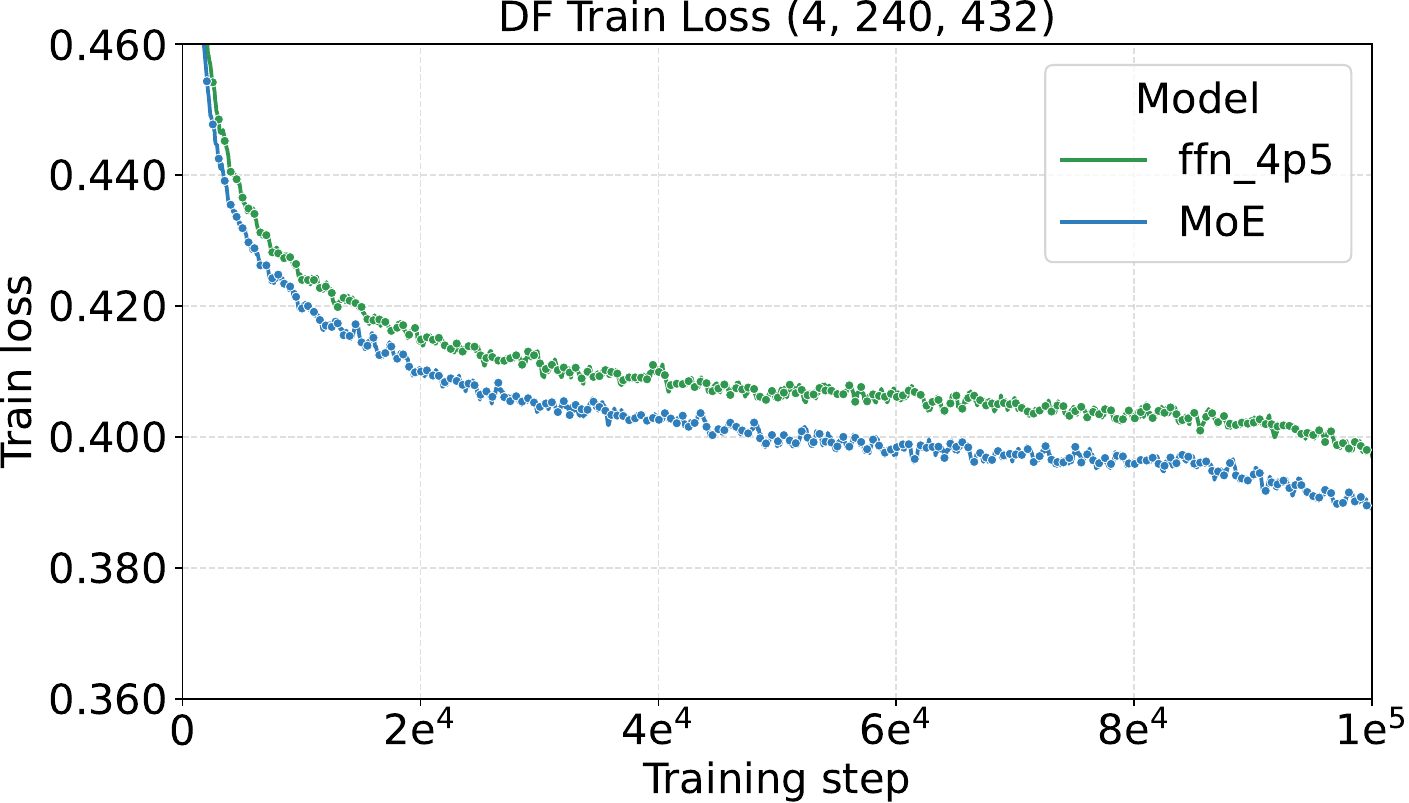}
\caption{240P key-frame training loss}
\label{fig:large-run-240p-keyframe-loss}
\end{subfigure}
\par\medskip
\begin{subfigure}[t]{0.322\textwidth}
\centering
\includegraphics[width=\linewidth]{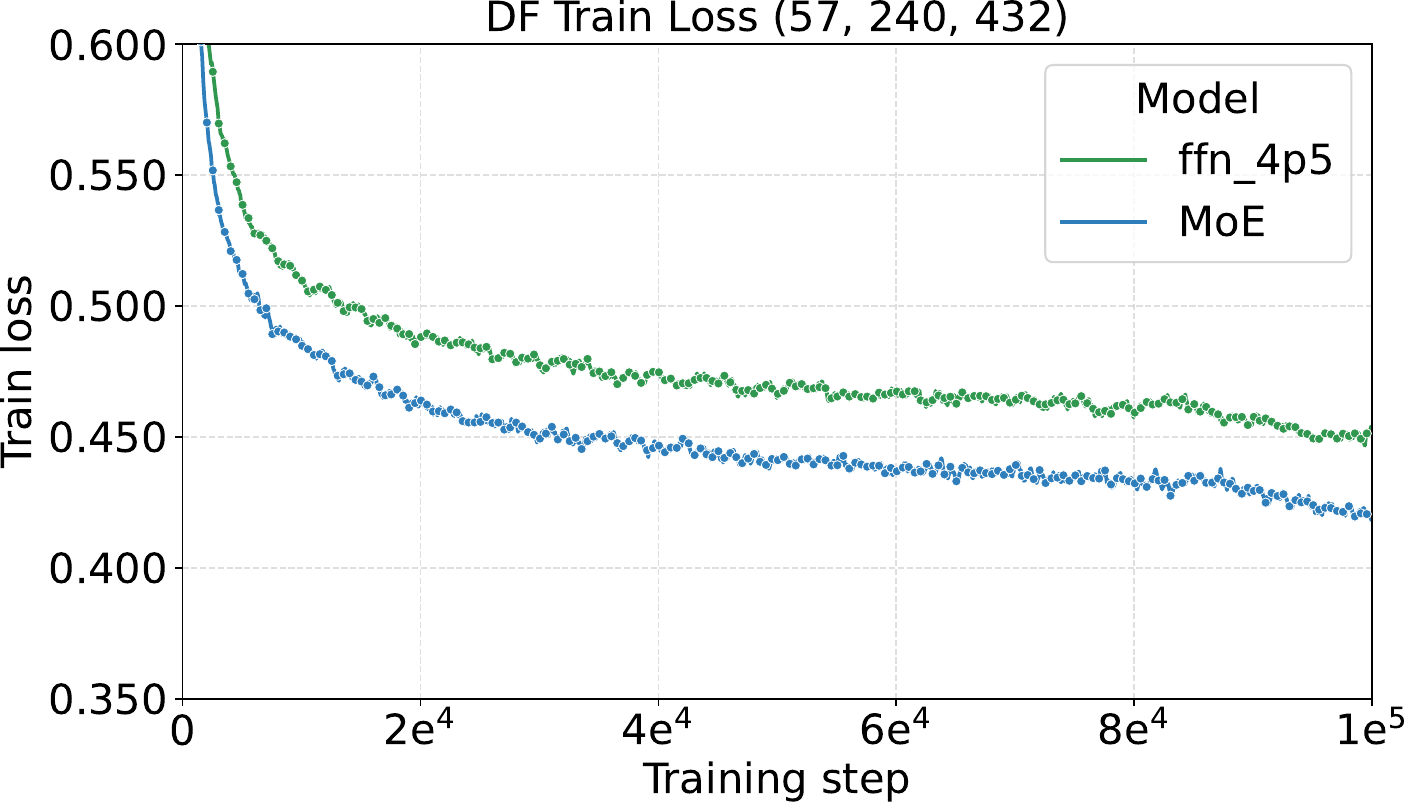}
\caption{240P 5s video training loss}
\label{fig:large-run-240p-video-loss}
\end{subfigure}%
\hspace{0.012\textwidth}%
\begin{subfigure}[t]{0.322\textwidth}
\centering
\includegraphics[width=\linewidth]{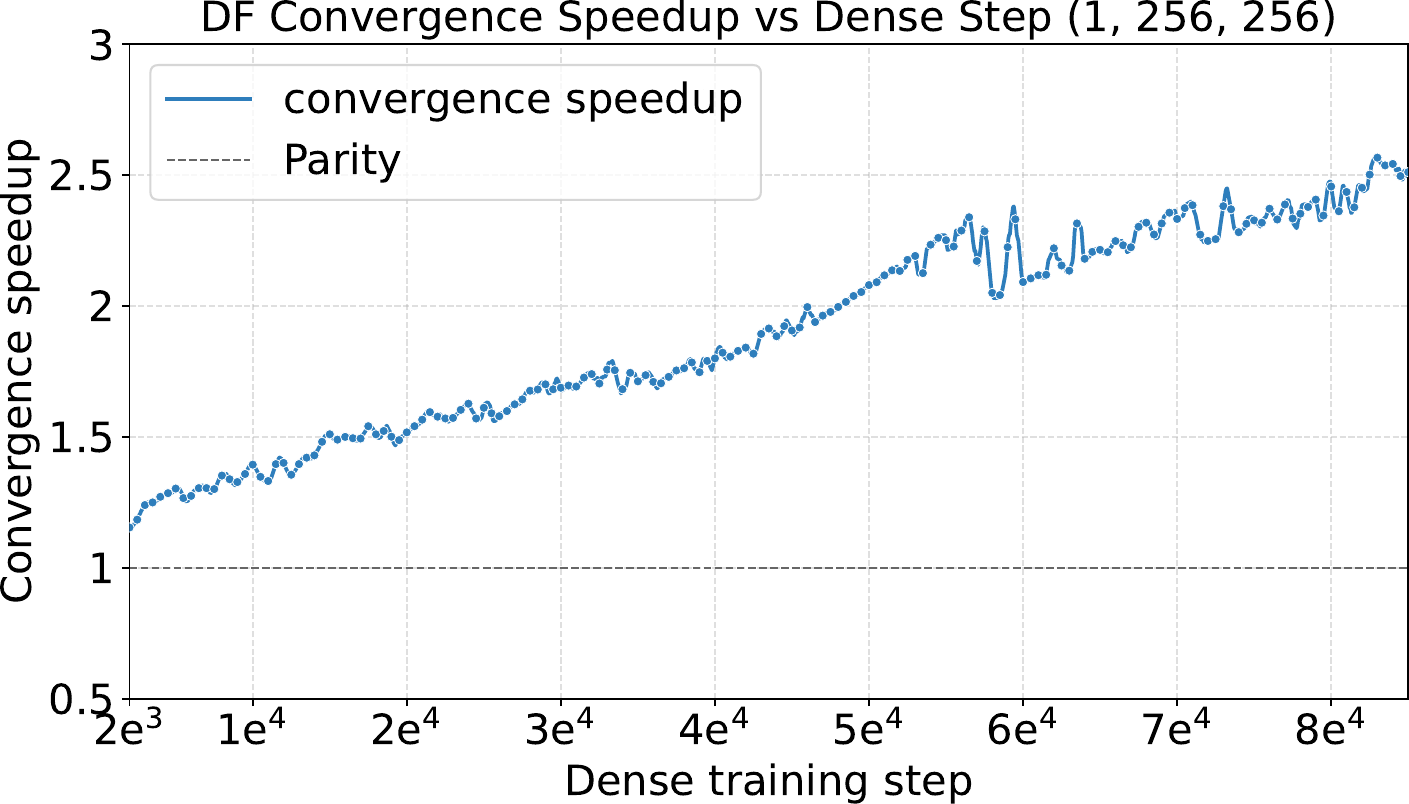}
\caption{256P image convergence speedup}
\label{fig:large-run-256p-image-speedup}
\end{subfigure}%
\hspace{0.012\textwidth}%
\begin{subfigure}[t]{0.322\textwidth}
\centering
\includegraphics[width=\linewidth]{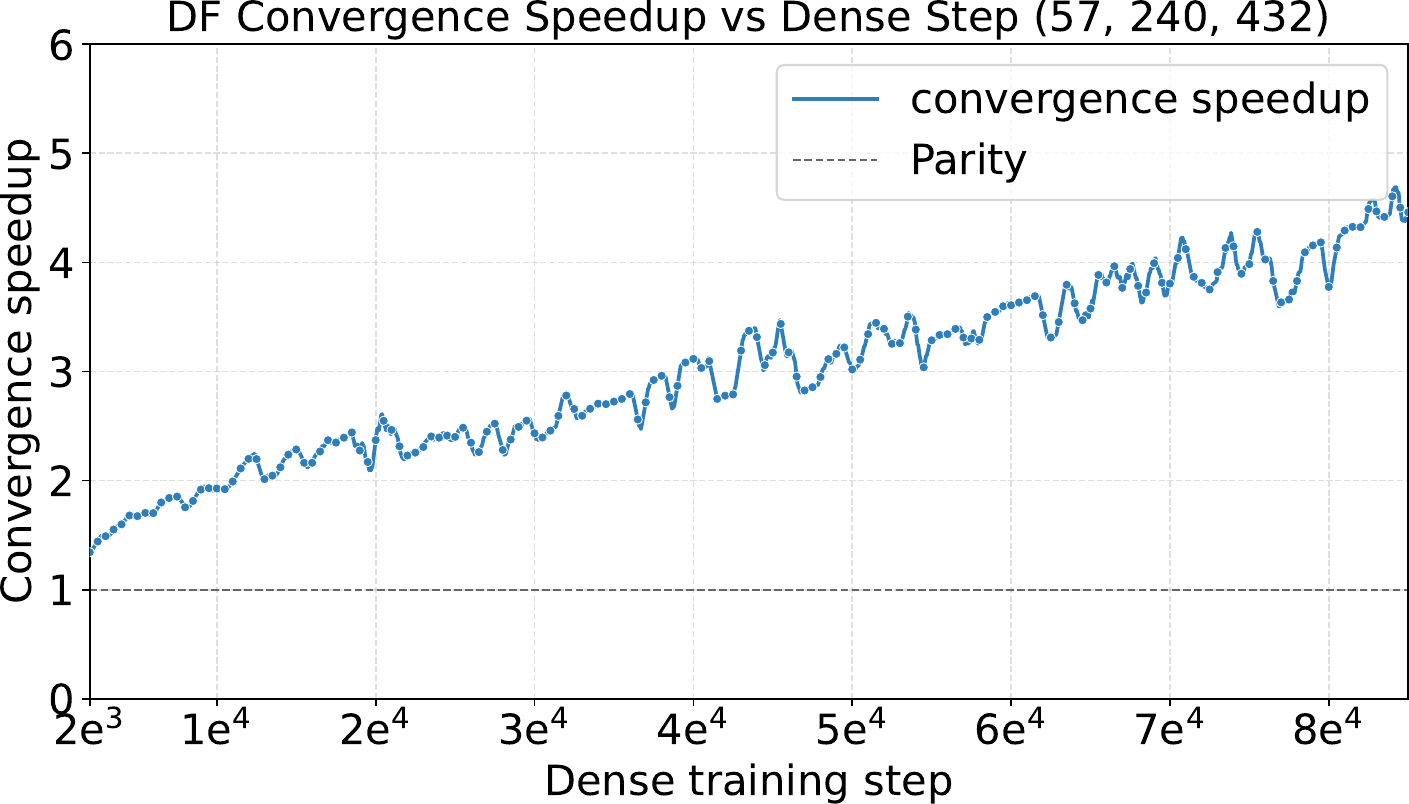}
\caption{240P 5s video convergence speedup}
\label{fig:large-run-240p-video-speedup}
\end{subfigure}
\par\medskip
\begin{subfigure}[t]{0.322\textwidth}
\centering
\includegraphics[width=\linewidth]{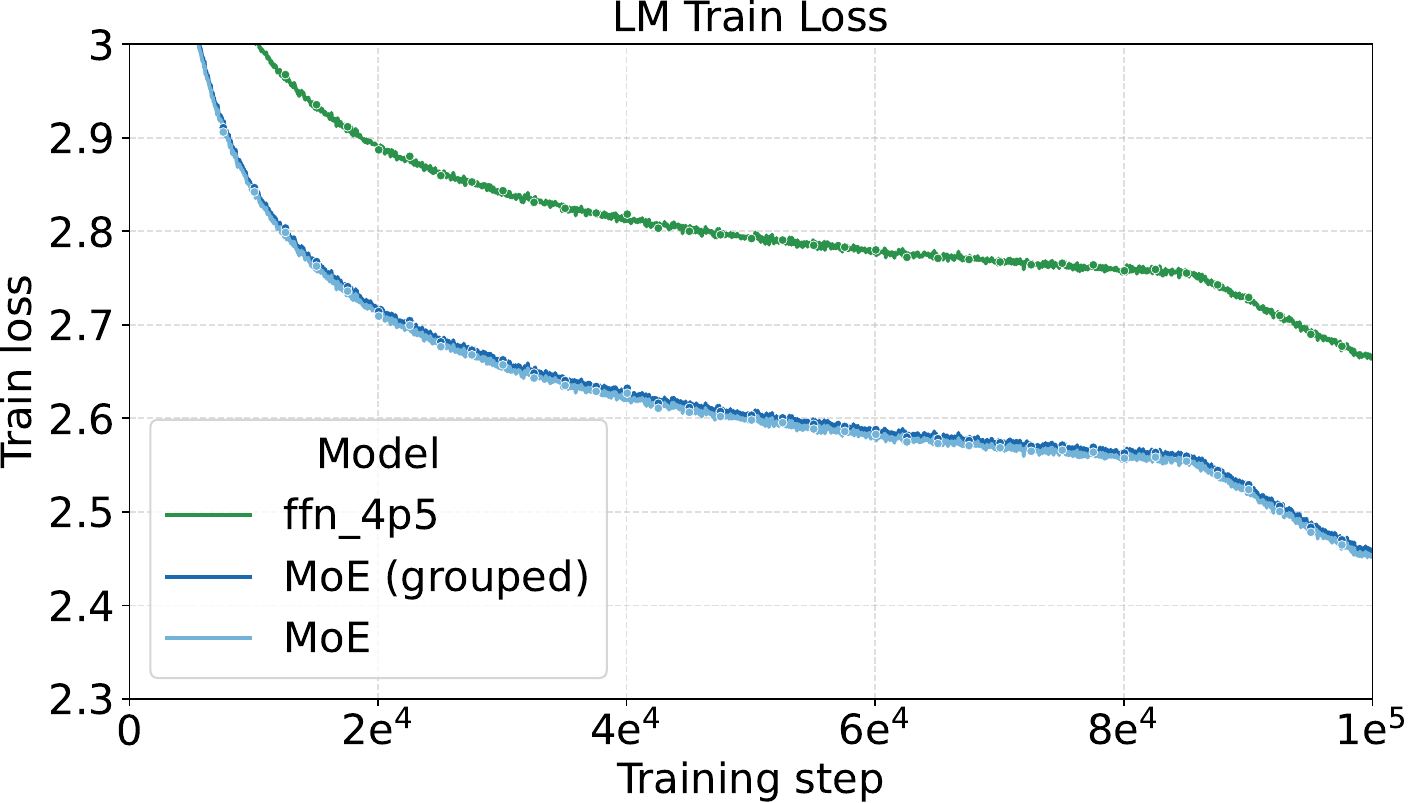}
\caption{LLM training loss}
\label{fig:large-run-text-train-loss}
\end{subfigure}%
\hspace{0.012\textwidth}%
\begin{subfigure}[t]{0.322\textwidth}
\centering
\includegraphics[width=\linewidth]{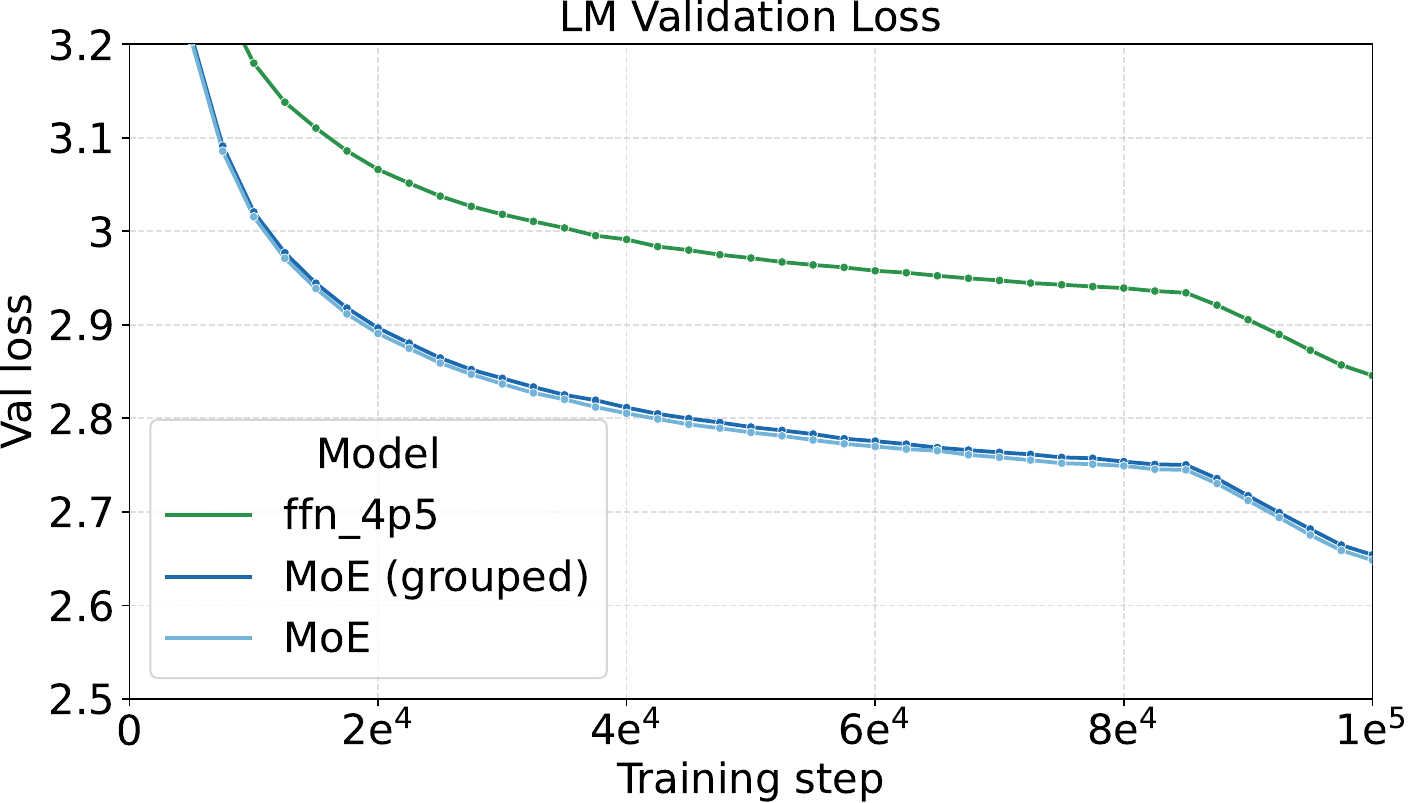}
\caption{LLM validation loss}
\label{fig:large-run-text-val-loss}
\end{subfigure}%
\hspace{0.012\textwidth}%
\begin{subfigure}[t]{0.322\textwidth}
\centering
\includegraphics[width=\linewidth]{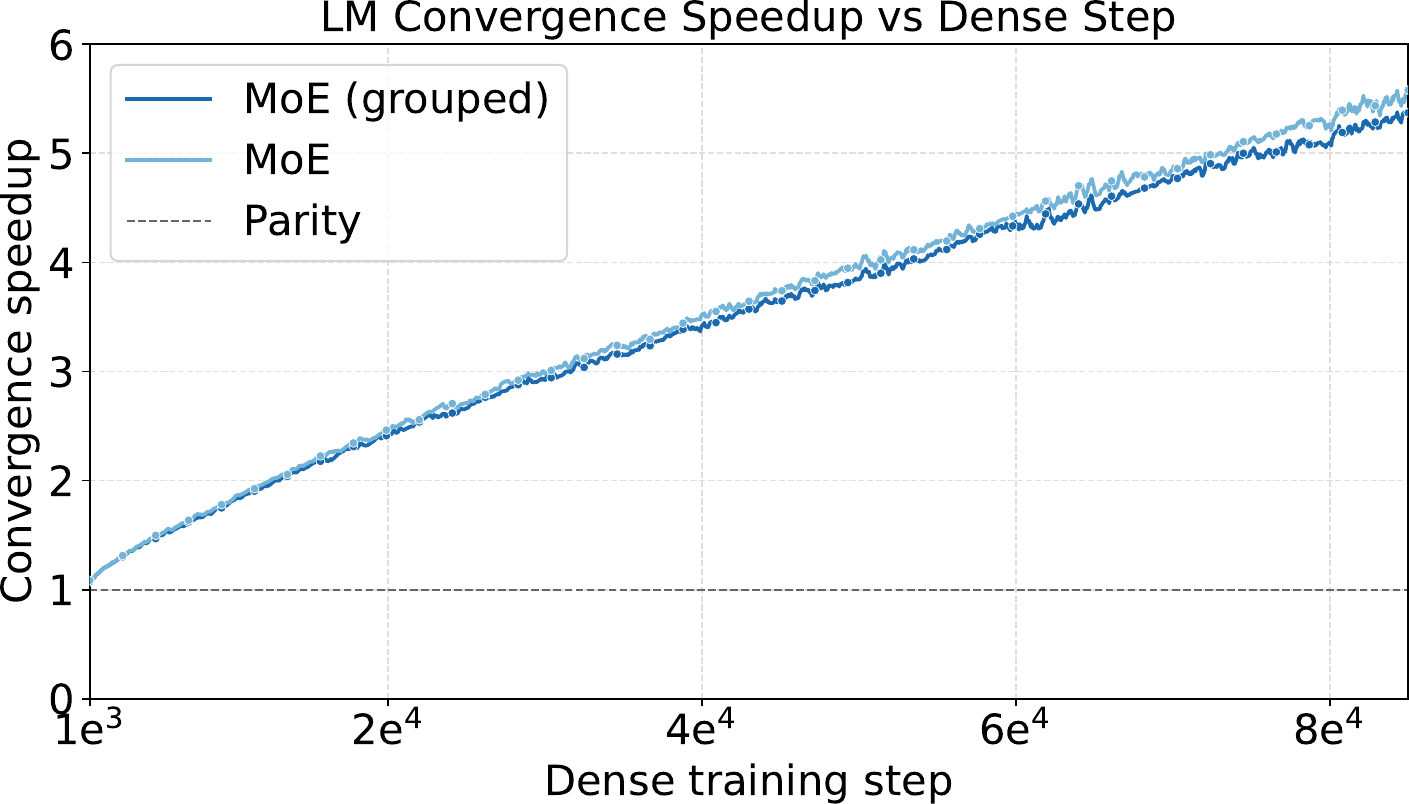}
\caption{LLM convergence speedup}
\label{fig:large-run-text-speedup}
\end{subfigure}
\caption{Complete-muE for larger-scale MoE training for multi-modal generation and LLMs. }
\label{fig:large-scale-complete-mue}
\end{figure}
Figure~\ref{fig:large-scale-complete-mue} shows that the recipe transfers beyond proxy sweeps. MoE models keep lower training loss across image, key-frame, video, and LM settings. We measure convergence speedup as dense steps divided by MoE steps needed to reach the same loss during the stable-LR phase of WSD. The MoE reaches roughly $2.5\times$ speedup on $256$P images and roughly $4.5\times$ on $240$P $5$s videos; the LLM MoE variants reach roughly $5.3\times$--$5.5\times$. The group-balanced variant is slightly behind the non-grouped routed variant, matching the smaller scale LM findings. Crucially, all four diffusion regimes ($256$P images, $512$P images, $240$P key frames, $240$P $5$s videos) share a single hyperparameter setting (LR=$2.26\times10^{-3}$, WD=$0.01$) and the LM run uses one other (LR=$5\times10^{-4}$, WD=$0.05$); every setting still delivers consistent MoE-over-dense convergence speedup without any per-setting retuning. This is direct large-scale evidence for the tune-dense-once-and-transfer recipe: one dense calibration suffices to deliver consistent MoE gains across the entire multimodal/LM training landscape.

\newcommand{\videosixframes}[1]{%
\includegraphics[width=0.158\linewidth]{#1_0001.png}\hfill
\includegraphics[width=0.158\linewidth]{#1_0002.png}\hfill
\includegraphics[width=0.158\linewidth]{#1_0003.png}\hfill
\includegraphics[width=0.158\linewidth]{#1_0004.png}\hfill
\includegraphics[width=0.158\linewidth]{#1_0005.png}\hfill
\includegraphics[width=0.158\linewidth]{#1_0006.png}%
}

\begin{figure}[!htb]
\centering
\begin{subfigure}[t]{\linewidth}
\centering
\videosixframes{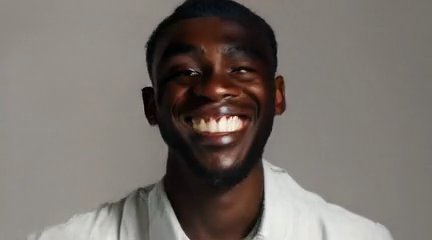}
\caption{Dense}
\label{fig:video-generation-caption-1-dense}
\end{subfigure}
\vspace{1.5mm}
\begin{subfigure}[t]{\linewidth}
\centering
\videosixframes{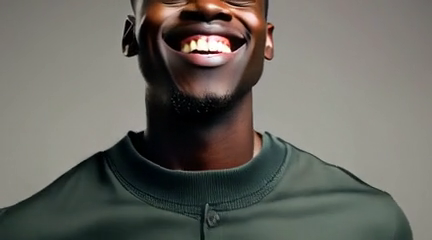}
\caption{MoE}
\label{fig:video-generation-caption-1-moe}
\end{subfigure}
\caption{\textbf{Caption 1:} A cinematic head-and-shoulders video of a Black man in casual attire against a neutral seamless backdrop, softly lit with shallow focus and centered rule-of-thirds composition. The camera performs a handheld dolly-in and zoom-in toward his face as he bursts into a broad, joyful smile with visible teeth, creating a warm and expressive portrait.}
\label{fig:video-generation-caption-1}
\end{figure}

\begin{table}[!htb]
\centering
\scriptsize
\setlength{\tabcolsep}{3pt}
\renewcommand{\arraystretch}{1.05}
\caption{\textbf{Larger-scale dense versus MoE benchmark evaluation.} Scores are percentages.}
\label{tab:large-scale-benchmark-comparison}
\resizebox{\linewidth}{!}{%
\begin{tabular}{l*{7}{c}}
\toprule
\textbf{Model} & \textbf{SVAMP(5)} & \textbf{MMLU(5)} & \textbf{ARC-Easy(5)} & \textbf{ARC-Challenge(5)} & \textbf{COPA(5)} & \textbf{PIQA(5)} & \textbf{HellaSwag(5)} \\
\cmidrule(lr){2-2}\cmidrule(lr){3-5}\cmidrule(lr){6-7}\cmidrule(lr){8-8}
\textbf{Dense} & 7.0 & 25.4 & 64.6 & 33.6 & 70.0 & 72.6 & 56.5 \\
\textbf{MoE \texttt{128e8a4g1s}} & 9.7 & 25.9 & 71.2 & \textbf{43.6} & 80.0 & 77.7 & 69.3 \\
\textbf{MoE \texttt{128e8a1s}} & \textbf{10.0} & \textbf{26.2} & \textbf{72.9} & 43.4 & \textbf{85.0} & \textbf{77.8} & \textbf{69.4} \\
\midrule
\textbf{Model} & \textbf{WinoGrande(5)} & \textbf{LAMBADA(5)} & \textbf{BoolQ(5)} & \textbf{AGIEval-LSAT-RC(5)} & \textbf{AGIEval-LSAT-LR(5)} & \textbf{AGIEval-SAT-EN(5)} & \textbf{Average} \\
\cmidrule(lr){2-3}\cmidrule(lr){4-7}\cmidrule(lr){8-8}
\textbf{Dense} & 59.0 & 50.5 & 62.6 & 20.9 & \textbf{27.3} & 26.2 & 44.3 \\
\textbf{MoE \texttt{128e8a4g1s}} & 65.2 & 61.9 & 63.4 & 22.8 & 27.1 & 20.9 & 49.1 \\
\textbf{MoE \texttt{128e8a1s}} & \textbf{66.5} & \textbf{63.4} & \textbf{63.9} & \textbf{27.2} & 24.3 & \textbf{27.2} & \textbf{50.6} \\
\bottomrule
\end{tabular}
}
\renewcommand{\arraystretch}{1.0}
\end{table}

The qualitative video example in Figure~\ref{fig:video-generation-caption-1} is consistent with the loss trends: the MoE sample shows cleaner structure and less flicker for this prompt, while the quantitative evidence remains the loss, speedup, and benchmark evaluation.
Table~\ref{tab:large-scale-benchmark-comparison} reports the final $100$k-step LM evaluation. Both MoE variants improve the dense average score: $44.3\to49.1$ for \texttt{128e8a4g1s} and $44.3\to50.6$ for \texttt{128e8a1s}. The non-grouped variant is best on average, while the group-balanced variant is strongest on ARC-Challenge.
\section{Conclusion}
We presented \emph{Complete-muE}, a compositional AdamW transfer rule for FFN/MoE families. We identify the key bottleneck and uses dense FFN to dense MoE, and dense MoE to sparse MoE to bridget the gap of transfer problem. The resulting recipe covers activated experts, total capacity, fixed-density granularity, shared/group-balanced hybrids, and width/depth/batch/duration changes.

Controlled LM and diffusion sweeps show relatively stable hyperparameter optima for LR, WD, and initialization across all MoE combinations, with mild drift consistent with Bridge~II's non-strict SDE behavior. In practice this drift is small enough that hyperparameters tuned on a single dense reference transfer near-optimally to all MoE configurations---\emph{tune dense once, transfer to all MoE settings} is the practical recipe at the core of Complete-muE. Both controlled small-scale axis sweeps and large-scale multimodal/LM runs directly verify this recipe: a single dense calibration delivers consistent MoE gains across MoE axes and across modalities ($256$P/$512$P images, $240$P key frames, $240$P $5$s videos, LM). The H100 benchmark shows that total-capacity scaling is much cheaper than fine granularity at fixed active width. Larger runs reach roughly convergence speedup up to $4.5\times$ on $240$P $5$s videos, and $5.3\times$--$5.5\times$ on LLM training for only 100k training steps.


\bibliographystyle{unsrt}
\bibliography{neurips_2026}


\clearpage
\appendix

\section{Additional derivations for Complete-muE}
\label{app:complete-mue}

This appendix collects the derivations that were moved out of the main paper. The presentation mirrors the main method section: notation, dense FFN active-width transfer, Dense MoE factorization, activated-expert SDE analysis, compositional derivations for capacity and granularity, the hybrid/shared-expert/group-balanced-routing extension, and a brief note on how the global AdamW transfer composes with Complete-muE.

\begin{table*}[t]
\centering
\scriptsize
\caption{Notation used in the method and appendix.}
\label{tab:mue-notation}
\begin{tabularx}{\textwidth}{@{}p{0.19\textwidth}X@{}}
\toprule
Symbol & Meaning \\
\midrule
$\rho_d,\rho_H$ & residual/model-width ratio $d/d_\star$ and active-width ratio $H/d$. \\
$\rho_L,\rho_B,\rho_D$ & depth, global batch, and global token-budget ratios relative to the reference model. \\
$\rho_{H_a},\rho_N$ & active-width ratio induced by activated experts, $\rho_{H_a}=ah/d$, and total-expert ratio $N/N_\star$. \\
$\rho_B^{\mathrm{exp}},\rho_D^{\mathrm{exp}}$ & expert-side effective batch and expert-token-budget ratios used in the SDE analysis. \\
$s$ & routing density $a/N$. \\
$d$ & residual/model width. \\
$H$ & effective active FFN width; for Dense MoE, $H=Nh$; for sparse MoE, $H=ah$. \\
$H_{\mathrm{tot}}$ & total active width of a shared/group-balanced-routing/hybrid block, $H_{\mathrm{tot}}=\sum_{m\in\mathcal{D}}H_m+a\sum_{g\in\mathcal{G}} h_g$. \\
$h$ & per-expert hidden width. \\
$\mathcal{D},\mathcal{G}$ & index sets of always-active dense/shared submodules and routed groups in a hybrid block. \\
$u(x)$ & FFN hidden activation vector after the up/gate projections. \\
$W_{\mathrm{up}},W_v,W_g$ & FFN/MoE up and gate projections; in MoE the same notation refers to the corresponding per-expert copies. \\
$v_{\mathrm{up}},\sigma_{\mathrm{up}},\eta_{\mathrm{up}}$ & initialization variance, standard deviation, and AdamW learning rate for an FFN/MoE up or gate projection; these follow the ordinary backbone-width $\mu$P rule. \\
$q_u,\mu_u$ & second moment and mean absolute value of one hidden coordinate of $u(x)$. \\
$W_{\mathrm{down}}$ & FFN output / down projection. \\
$v_{\mathrm{down}},\sigma_{\mathrm{down}}$ & initialization variance and standard deviation of one entry of $W_{\mathrm{down}}$. \\
$\eta_{\mathrm{down}}$ & AdamW learning rate applied to $W_{\mathrm{down}}$. \\
$N$ & total number of routed experts. \\
$a$ & number of activated experts per token; Dense MoE corresponds to $a=N$. \\
$\mathcal{A}(x)$ & selected expert set for token $x$, with $|\mathcal{A}(x)|=a$. \\
$\pi_e(x)$ & normalized routing weight on expert $e$, satisfying $\sum_e \pi_e(x)=1$ over the active experts. \\
$A(H),A_a$ & Complete-muE output-branch multiplier for active width $H$, and its sparse-MoE specialization at $H_a=ah$. \\
$r_a$ & route scale multiplying the normalized routed sum. \\
$F_{a,N}$ & finite-width routing factor $a\,\mathbb{E}[\sum_{e=1}^N \pi_e(x)^2]$. \\
$W_{\mathrm{gate}}$ & router projection producing logits $\ell(x)=W_{\mathrm{gate}}x+b$. \\
$v_{\mathrm{gate}},\sigma_{\mathrm{gate}},\eta_{\mathrm{gate}}$ & initialization variance, standard deviation, and learning rate of the router projection. \\
$\eta,\lambda,\epsilon,\beta_1,\beta_2$ & global AdamW learning rate, weight decay, numerical stabilizer, and momentum coefficients. \\
$B$ & global token batch per optimizer step. \\
$D$ & global training token budget / total number of tokens seen during training. \\
$T$ & number of optimizer steps. \\
$B_{\mathrm{exp}}(a)$ & mean routed-token batch seen by one expert when $a$ experts are activated. \\
$D_{\mathrm{exp}}(a)$ & mean expert-token budget across training, $D_{\mathrm{exp}}(a)=T\,B_{\mathrm{exp}}(a)$. \\
$\sigma_{\mathrm{exp}}(a)$ & per-expert gradient-noise scale in the simplified SDE model; under balanced routing it scales as $B_{\mathrm{exp}}(a)^{-1/2}$. \\
$\sigma_0(a)$ & effective SDE signal-to-noise parameter, $\sigma_0(a)=\eta\,\sigma_{\mathrm{exp}}(a)$. \\
$H_{\mathrm{SDE}}$ & SDE integration horizon, $H_{\mathrm{SDE}}=T\eta^2$. \\
$L$ & number of residual layers. \\
\bottomrule
\end{tabularx}
\end{table*}

\subsection{Dense FFN active-width transfer derivation}
\label{app:dense-derivation}

Consider a dense FFN output branch
\begin{equation}
\label{eq:base-ffn}
y(x)=c_H W_{\mathrm{down}}u(x),
\end{equation}
where $u(x)\in\mathbb{R}^{H}$ denotes the hidden activation after the up/gate projections. Let
\[
q_u:=\mathbb{E}[u_j(x)^2],
\qquad
\mu_u:=\mathbb{E}[|u_j(x)|],
\]
and initialize $W_{\mathrm{down},ij}\sim\mathcal{N}(0,v_{\mathrm{down}})$. Under the AdamW gradient-magnitude-normalized regime,
\begin{equation}
\label{eq:adamw-normalized}
\Delta W_{\mathrm{down}}=-\eta_{\mathrm{down}}\,\mathcal{P}(G_{\mathrm{down}}),
\qquad
\mathcal{P}(cG)\approx \mathcal{P}(G)\quad \text{for } c>0,
\end{equation}
so matching one-step functional updates only depends on explicit forward multipliers. The forward variance and one-step functional update scale as
\begin{align}
\Var[y_i(x)] &\asymp c_H^2 H v_{\mathrm{down}} q_u, \label{eq:base-forward-scale}\\
\mathbb{E}|\Delta y_i(x)| &\asymp c_H\eta_{\mathrm{down}}H\mu_u. \label{eq:base-update-scale}
\end{align}
To match these quantities to the unit-expansion dense companion at the same backbone width $d$, we set
\begin{equation}
\label{eq:ffn-abc}
A_H=c_H=\frac{d}{H},
\qquad
B_H^2=v_{\mathrm{down}}(d,H)=\frac{H}{d}v_{\mathrm{down}}^{(1)}(d),
\qquad
C_H=\eta_{\mathrm{down}}(d,H)=\eta_{\mathrm{down}}^{(1)}(d).
\end{equation}
Equivalently,
\[
\sigma_{\mathrm{down}}(d,H)=\sqrt{\frac{H}{d}}\,\sigma_{\mathrm{down}}^{(1)}(d).
\]
Eq.~\eqref{eq:ffn-abc} is the active-width ABC form used throughout the paper. The up/gate projections still follow ordinary backbone-width $\mu$P,
\begin{equation}
\label{eq:ffn-up-mup}
\sigma_{\mathrm{up}}(d)=\rho_d^{-1/2}\sigma_{\mathrm{up},\star},
\qquad
\eta_{\mathrm{up}}(d)=\rho_d^{-1}\eta_{\mathrm{up},\star},
\end{equation}
and the same rule applies componentwise to gated MLP projections. The router readout also remains ordinary $\mu$P,
\begin{equation}
\label{eq:router-standard-mup}
\sigma_{\mathrm{gate}}(d)=\rho_d^{-1/2}\sigma_{\mathrm{gate},\star},
\qquad
\eta_{\mathrm{gate}}(d)=\rho_d^{-1}\eta_{\mathrm{gate},\star}.
\end{equation}

\subsection{Dense MoE factorization and normalized route scale}
\label{app:router-derivation}

For Dense MoE or sparse MoE, let expert $e$ produce
\[
o_i^{(e)}(x)=\sum_{\ell=1}^{h}W_{\mathrm{down},i\ell}^{(e)}u_{\ell}^{(e)}(x),
\]
and let the routed branch be
\begin{equation}
\label{eq:moe-output}
y_i^{\mathrm{MoE}}(x)=r_a\sum_{e=1}^{N}\pi_e(x)\,o_i^{(e)}(x),
\qquad
\sum_{e=1}^{N}\pi_e(x)=1.
\end{equation}
Since one expert has width $h$, its output variance satisfies
\begin{equation}
\label{eq:single-expert-forward}
\Var[o_i^{(e)}(x)]\asymp h v_{\mathrm{down}} q_u.
\end{equation}
Using independence and zero-mean initialization across experts,
\begin{equation}
\label{eq:moe-forward}
\Var[y_i^{\mathrm{MoE}}(x)]
\asymp
H v_{\mathrm{down}} q_u\cdot \frac{r_a^2}{a^2}F_{a,N},
\qquad
F_{a,N}:=a\,\mathbb{E}\Big[\sum_{e=1}^{N}\pi_e(x)^2\Big],
\end{equation}
where $H=ah$ is the active width. For the one-step update, the expert gradient is
\[
G_{\mathrm{down},i\ell}^{(e)}=r_a\pi_e(x)g_i u_{\ell}^{(e)}(x),
\]
so Eq.~\eqref{eq:adamw-normalized} removes the leading dependence on the positive scalar $r_a\pi_e(x)$. Hence
\begin{equation}
\label{eq:moe-update}
\mathbb{E}|\Delta y_i^{\mathrm{MoE}}(x)|\asymp r_a\eta_{\mathrm{down}}h\mu_u.
\end{equation}
Matching this to the dense FFN update at the same active width $H=ah$ yields the route scale
\begin{equation}
\label{eq:route-scale-exact-main}
r_a=a,
\end{equation}
which is Eq.~\eqref{eq:route-scale-main} in the main paper. Exact forward matching would instead use
\begin{equation}
\label{eq:route-scale-exact}
r_a^{\mathrm{exact-fwd}}=\frac{a}{\sqrt{F_{a,N}}}.
\end{equation}
In practice we keep the simpler main rule $r_a=a$ because $F_{a,N}$ is usually close to one at initialization.

This bounded correction can be understood by expanding the normalized router around equal active logits. Write $\ell_j=\bar\ell+\delta_j$ for $j\in\mathcal{A}(x)$ with $\sum_j \delta_j=0$. For normalized softmax,
\begin{equation}
\label{eq:f-softmax-expand}
\pi_j=\frac{1}{a}+\frac{\delta_j}{a}+O(\|\delta\|^2),
\qquad
F_{a,N}=1+\frac{1}{a}\,\mathbb{E}\Big[\sum_{j\in\mathcal{A}(x)}\delta_j^2\Big]+O\!\bigl(\mathbb{E}\|\delta\|^3\bigr).
\end{equation}
For normalized sigmoid, letting $\kappa(\bar\ell):=\sigma'(\bar\ell)/\sigma(\bar\ell)$ with $\sigma(\cdot)$ the logistic sigmoid,
\begin{equation}
\label{eq:f-sigmoid-expand}
\pi_j=\frac{1}{a}+\frac{\kappa(\bar\ell)}{a}\delta_j+O(\|\delta\|^2),
\qquad
F_{a,N}=1+\kappa(\bar\ell)^2\frac{1}{a}\,\mathbb{E}\Big[\sum_{j\in\mathcal{A}(x)}\delta_j^2\Big]+O\!\bigl(\mathbb{E}\|\delta\|^3\bigr).
\end{equation}
Thus $F_{a,N}$ remains an order-one constant near one unless the selected logits are already highly separated.

\subsection{Expert-side SDE analysis for activated experts}
\label{app:sde-derivation}

The layer rule in Eq.~\eqref{eq:sparsity-transfer} already matches the routed-FFN layer scales. What remains is the expert-side stochastic effect induced by changing how many tokens each expert receives. Under approximate load balancing,
\begin{equation}
\label{eq:per-expert-batch}
B_{\mathrm{exp}}(a)\approx B\frac{a}{N},
\qquad
D_{\mathrm{exp}}(a)=TB_{\mathrm{exp}}(a)\approx TB\frac{a}{N}.
\end{equation}
Therefore
\[
\sigma_{\mathrm{exp}}(a)\propto B_{\mathrm{exp}}(a)^{-1/2}\propto \sqrt{\frac{N}{Ba}}.
\]
Changing the activated-expert count from $a$ to $a'$ therefore changes the expert noise scale by
\begin{equation}
\label{eq:expert-noise-ratio}
\frac{\sigma_{\mathrm{exp}}(a')}{\sigma_{\mathrm{exp}}(a)}
=
\sqrt{\frac{B_{\mathrm{exp}}(a)}{B_{\mathrm{exp}}(a')}}
=
\frac{1}{\sqrt{\rho_B^{\mathrm{exp}}}},
\qquad
\rho_B^{\mathrm{exp}}
:=
\frac{B_{\mathrm{exp}}(a')}{B_{\mathrm{exp}}(a)}
=
\frac{a'}{a}.
\end{equation}
Model one expert by the stochastic gradient decomposition
\begin{equation}
\label{eq:expert-noise-model}
g_t^{(e)}(a)=\bar g^{(e)}+\sigma_{\mathrm{exp}}(a)\,\xi_t^{(e)},
\qquad
\xi_t^{(e)}\sim \mathcal{N}(0,I).
\end{equation}
Under a simplified normalized RMSPropW/AdamW proxy,
\begin{equation}
\label{eq:expert-rmsprop-iter}
\vartheta_{t+1}^{(e)}
=
\vartheta_t^{(e)}
-
\eta\left(
\frac{g_t^{(e)}(a)}{\sigma_{\mathrm{exp}}(a)}+\lambda\vartheta_t^{(e)}
\right).
\end{equation}
Introducing the continuous-time step $\Delta\tau=\eta^2$, define
\begin{equation}
\label{eq:expert-sde-objects}
\sigma_0(a):=\eta\,\sigma_{\mathrm{exp}}(a),
\qquad
\widetilde{\lambda}:=\frac{\lambda}{\eta},
\qquad
H_{\mathrm{SDE}}:=T\eta^2.
\end{equation}
Then Eq.~\eqref{eq:expert-rmsprop-iter} is the Euler--Maruyama discretization of
\begin{equation}
\label{eq:expert-rmsprop-sde}
d\Theta_{\tau}^{(e)}
=
-\frac{1}{\sigma_0(a)}\bar g^{(e)}\,d\tau
-\widetilde{\lambda}\Theta_{\tau}^{(e)}\,d\tau
-dW_{\tau}^{(e)},
\qquad
0\le \tau \le H_{\mathrm{SDE}}.
\end{equation}
The Brownian coefficient is normalized to one, so the per-expert SDE is governed by three quantities: the signal-to-noise parameter $\sigma_0(a)$ (smaller means less noise relative to gradient drift), the normalized weight decay $\widetilde{\lambda}$, and the optimization horizon $H_{\mathrm{SDE}}$.

\paragraph{Reference: exact batch scaling at fixed expert-token budget.}
Before analyzing activated-expert transfer, consider a clean reference case. Keep the architecture and sparsity $a$ fixed, increase the global batch from $B$ to $\kappa_B B$, and reduce optimizer steps from $T$ to $T/\kappa_B$ so that total trained tokens are unchanged. Under balanced routing, each expert then sees $B_{\mathrm{exp}}'(a)=\kappa_B B_{\mathrm{exp}}(a)$ and $D_{\mathrm{exp}}'(a)=D_{\mathrm{exp}}(a)$, giving $\rho_B^{\mathrm{exp}}=\kappa_B$ and $\rho_D^{\mathrm{exp}}=1$. Preserving all three SDE quantities ($\sigma_0'=\sigma_0$, $\widetilde{\lambda}'=\widetilde{\lambda}$, $H_{\mathrm{SDE}}'=H_{\mathrm{SDE}}$) requires
\begin{equation}
\label{eq:expert-exact-rule}
\eta'=\sqrt{\rho_B^{\mathrm{exp}}}\,\eta,
\qquad
\lambda'=\sqrt{\rho_B^{\mathrm{exp}}}\,\lambda.
\end{equation}
This is the dense fixed-token batch-transfer rule applied to the expert process~\cite{malladi2022sdes,apple2025completedmup}.

\paragraph{Why activated-expert transfer is not an exact SDE invariance problem.}
In activated-expert transfer we instead change $a$ to $a'$ at fixed global batch $B$ and fixed steps $T$. Unlike the reference case ($\rho_D^{\mathrm{exp}}=1$), this changes \emph{both} expert batch and expert duration by the same factor:
\begin{equation}
\label{eq:expert-activated-factors-appendix}
\rho_B^{\mathrm{exp}}=\frac{B_{\mathrm{exp}}(a')}{B_{\mathrm{exp}}(a)}=\frac{a'}{a},
\qquad
\rho_D^{\mathrm{exp}}=\frac{D_{\mathrm{exp}}(a')}{D_{\mathrm{exp}}(a)}=\frac{a'}{a}.
\end{equation}
Applying rule~\eqref{eq:expert-exact-rule} here would scale $\eta$ by $\sqrt{a'/a}$, simultaneously shifting the SDE horizon to $H_{\mathrm{SDE}}'=T(\eta')^2=(a'/a)H_{\mathrm{SDE}}$ and perturbing non-sparse parameters, which do not obey the same sparsity-dependent transformation. We therefore do not claim an exact full-model SDE invariance for activated-expert transfer.

\paragraph{Hypothesis: after layer-level matching, the SDE correction cancels.}
Following the approximation philosophy of Complete$(d)$P~\cite{apple2025completedmup}, we hypothesize that once Eq.~\eqref{eq:sparsity-transfer} is in place, changing $a$ primarily shifts the expert-side SNR $\sigma_0(a)$ while the global AdamW terms and SDE horizon remain unchanged. Under this hypothesis, the combined correction applies the expert batch factor together with an iso-horizon duration factor:
\begin{equation}
\label{eq:expert-combined-rule}
\eta'
\approx
\underbrace{\sqrt{\rho_B^{\mathrm{exp}}}}_{\text{expert batch}}
\cdot
\underbrace{(\rho_D^{\mathrm{exp}})^{-1/2}}_{\text{iso-horizon duration}}
\cdot\eta
=
\eta\sqrt{\frac{\rho_B^{\mathrm{exp}}}{\rho_D^{\mathrm{exp}}}},
\end{equation}
and similarly for $\lambda'$. Since $\rho_B^{\mathrm{exp}}=\rho_D^{\mathrm{exp}}=a'/a$, the correction is unity:
Therefore the dense-style batch/duration correction cancels,
\begin{equation}
\label{eq:expert-approx-rule}
\eta'\approx \eta\sqrt{\frac{\rho_B^{\mathrm{exp}}}{\rho_D^{\mathrm{exp}}}}=\eta,
\qquad
\lambda'\approx \lambda\sqrt{\frac{\rho_B^{\mathrm{exp}}}{\rho_D^{\mathrm{exp}}}}=\lambda,
\end{equation}
which is Eq.~\eqref{eq:sde-cancel} in the main paper. What does change is the expert-side signal-to-noise parameter:
\begin{equation}
\label{eq:expert-sigma0-drift}
\sigma_0(a')=\eta\,\sigma_{\mathrm{exp}}(a')=\frac{\sigma_0(a)}{\sqrt{\rho_B^{\mathrm{exp}}}}.
\end{equation}
Making the MoE less sparse (larger $a'$) reduces $\sigma_0(a')$ and thereby improves expert-side SNR at fixed optimization horizon $H_{\mathrm{SDE}}$; this is the direct analogue of the iso-horizon picture in Complete$(d)$P~\cite{apple2025completedmup}. The interpretation is not that changing $a$ leaves training unchanged. Rather, after the Complete-muE layer reparameterization, the remaining first-order effect of changing $a$ is a shift in expert-side SNR at roughly fixed optimization horizon, not a new global LR/WD transfer rule. Because the $\sigma_0(a)$ shift is not absorbed by the first-order $\eta,\lambda$ correction, mild hyperparameter drift across $a$ is expected---bounded in magnitude by the $\sigma_0$ shift---and Figure~\ref{fig:activated-experts-scaling} confirms this empirically: optima remain relatively stable across activated-expert counts with only minor drift.

\paragraph{Generalization to imbalanced routing.}
The balanced analysis above is the clean closed-form proxy. In real MoE training the router is not perfectly load balanced, so different experts receive different numbers of tokens at different steps. Let $\ell_t^{(e)}(a)$ denote the realized normalized load of expert $e$ at step $t$, with $\sum_e \ell_t^{(e)}(a)=a$, so that
\begin{equation}
\label{eq:expert-imbalance-load}
B_{\mathrm{exp},t}^{(e)}(a)=B\,\ell_t^{(e)}(a),
\qquad
D_{\mathrm{exp}}^{(e)}(a)=\sum_{t=1}^T B_{\mathrm{exp},t}^{(e)}(a).
\end{equation}
The expert-noise scale and SNR parameter become step-dependent,
\begin{equation}
\label{eq:expert-imbalance-sigma0}
\sigma_{\mathrm{exp},t}^{(e)}(a)\propto \bigl(B_{\mathrm{exp},t}^{(e)}(a)\bigr)^{-1/2},
\qquad
\sigma_{0,t}^{(e)}(a):=\eta\,\sigma_{\mathrm{exp},t}^{(e)}(a),
\end{equation}
so there is generally no single expert-independent constant $\sigma_0(a')$ after changing $a$. The first-order cancellation nevertheless survives at the expert-averaged level. Define the time-averaged load
\[
\bar \ell^{(e)}(a):=\frac{1}{T}\sum_{t=1}^{T}\ell_t^{(e)}(a),
\qquad
\bar B_{\mathrm{exp}}^{(e)}(a)=B\bar \ell^{(e)}(a),
\qquad
D_{\mathrm{exp}}^{(e)}(a)=T\bar B_{\mathrm{exp}}^{(e)}(a).
\]
For every expert with nonzero average load under both settings, the effective batch and duration factors are equal:
\begin{equation}
\label{eq:expert-imbalance-factors}
\bar m_{B,e}^{\mathrm{exp}}
:=
\frac{\bar B_{\mathrm{exp}}^{(e)}(a')}{\bar B_{\mathrm{exp}}^{(e)}(a)}
=
\frac{\bar\ell^{(e)}(a')}{\bar\ell^{(e)}(a)},
\qquad
\bar m_{D,e}^{\mathrm{exp}}
:=
\frac{D_{\mathrm{exp}}^{(e)}(a')}{D_{\mathrm{exp}}^{(e)}(a)}
=
\frac{\bar\ell^{(e)}(a')}{\bar\ell^{(e)}(a)}
=
\bar m_{B,e}^{\mathrm{exp}}.
\end{equation}
Therefore the same expert-side approximation gives
\begin{equation}
\label{eq:expert-imbalance-cancel}
\eta'_e\approx\eta\sqrt{\frac{\bar m_{B,e}^{\mathrm{exp}}}{\bar m_{D,e}^{\mathrm{exp}}}}=\eta,
\qquad
\lambda'_e\approx\lambda\sqrt{\frac{\bar m_{B,e}^{\mathrm{exp}}}{\bar m_{D,e}^{\mathrm{exp}}}}=\lambda,
\end{equation}
so the same cancellation holds expertwise to first order. What changes in the imbalanced case is not the first-order cancellation, but the fact that $\sigma_{0,t}^{(e)}(a)$ fluctuates across experts and training steps; these mild deviations contribute to---but do not dominate---the bounded hyperparameter drift visible in Figure~\ref{fig:activated-experts-scaling}, where learning-rate optima remain relatively stable across activated-expert counts even without strict load balancing.

\paragraph{Summary.}
After the Complete-muE layer-level reparameterization, changing the activated-expert count is best viewed as keeping the deterministic routed-FFN update matched while shifting the expert-side SNR at roughly fixed optimization horizon. The first-order SDE correction cancels, so AdamW hyperparameter transfer remains effective to first order even though different sparsity levels can still reach different attainable losses. Equivalently, activated-expert transfer is \emph{relatively stable} rather than a strict invariance: the residual $\sigma_0(a)$ shift produces mild, bounded hyperparameter drift across $a$, consistent with Figure~\ref{fig:activated-experts-scaling}.

\subsection{Compositional derivations: total experts and granularity}
\label{app:composition-derivation}

The two bridges established above yield all remaining MoE scaling cases by composition. The key organizing principle is that total-expert and granularity changes are \emph{not} new primitive rules---they are derivable by composing the dense-width and activated-expert transfer rules. In each case the final sparse layer recovers the same active-width rule $A_a=d/(ah)$, route scale $r_a=a$, and output initialization, confirming that the active width $H_a=ah$ is the single governing quantity.

\paragraph{Total expert count.}
Consider transfer from $(N,a,h)$ to $(N',a,h)$ with fixed activated experts $a$ and fixed per-expert width $h$. Introduce the Dense-MoE companions $(N,N,h)$ and $(N',N',h)$, obtained by activating all experts. The transfer decomposes into a dense-width step from total width $Nh$ to $N'h$ followed by the reverse sparsity transfer from $N'$ activated experts back to $a$. These two steps are explicitly:
\begin{enumerate}
    \item \textbf{Dense-width step.} Transfer the auxiliary Dense MoE from total width $Nh$ to $N'h$ via Bridge~I. The output multiplier and initialization become
    \[
    c^{(1)}=\frac{d}{N'h},
    \qquad
    \sigma_{\mathrm{down}}^{(1)}=\sqrt{\frac{N'h}{d}}\,\sigma_{\mathrm{down},\star}^{(1)},
    \]
    and the auxiliary route scale becomes $r_{N'}=N'$.
    \item \textbf{Reverse-sparsity step.} At fixed total experts $N'$, reduce the activated count from $N'$ back to $a$ via Bridge~II. The active-width rule at $H_a=ah$ gives
    \[
    c^{(2)}=\frac{d}{ah},
    \qquad
    \sigma_{\mathrm{down}}^{(2)}=\sqrt{\frac{ah}{d}}\,\sigma_{\mathrm{down},\star}^{(1)},
    \]
    and restores the route scale $r_a=a$.
\end{enumerate}
The extra dense-width factor from Step~1 is exactly reversed by Step~2. Because the sparsity rule exactly reverses the extra active-width factor introduced by the dense step, the final sparse layer again uses
\[
A_a=\frac{d}{ah},
\qquad
\sigma_{\mathrm{down}}(d,a)=\sqrt{\frac{ah}{d}}\,\sigma_{\mathrm{down}}^{(1)}(d),
\qquad
r_a=a,
\]
up to the bounded routing factor $F_{a,N}$. This is identical to the rule obtained by sparsifying the original companion $(N,N,h)$ directly to $a$ activated experts. Thus total-expert transfer is not a new primitive hyperparameter-transfer rule; it is a composition of dense-width transfer and activated-expert transfer.

The SDE bookkeeping is consistent with the same conclusion. Under balanced routing,
\begin{equation}
\label{eq:expert-count-batch-duration}
B_{\mathrm{exp}}(N)\approx B\frac{a}{N},
\qquad
D_{\mathrm{exp}}(N)\approx TB\frac{a}{N},
\end{equation}
so for $N\to N'$ we again have $\rho_B^{\mathrm{exp}}=\rho_D^{\mathrm{exp}}=N/N'$, and the first-order SDE cancellation from Appendix~\ref{app:sde-derivation} gives $\eta'\approx\eta$, $\lambda'\approx\lambda$, consistent with the deterministic derivation above. The SDE perspective is best read as a consistency check after the layer-level cancellation has been established. What remains is the capacity--noise trade-off discussed in the main text: increasing $N$ raises representational capacity but reduces the expert-wise token budget, worsening expert-side SNR:
\begin{equation}
\label{eq:expert-count-noise}
\sigma_{\mathrm{exp}}(N)\propto\sqrt{\frac{N}{Ba}},
\qquad
\sigma_0(N)=\eta\,\sigma_{\mathrm{exp}}(N)\propto\eta\sqrt{\frac{N}{Ba}}.
\end{equation}
These two effects compete: over some range of $N$ the extra capacity may dominate and loss decreases while the optimal LR stays nearly unchanged; at larger $N$, saturation or reversal is possible once the SNR penalty becomes dominant. The transferable-hyperparameter claim is therefore only that the same raw AdamW hyperparameters remain appropriate after the two-step cancellation, not that increasing $N$ must monotonically improve loss. As with activated-expert transfer, the reverse-sparsity step within this composition is not a strict SDE invariance---$\sigma_0(N)$ shifts with $N$ and is not absorbed by the first-order $\eta,\lambda$ correction---so mild hyperparameter drift across $N$ is expected. Figure~\ref{fig:total-experts-transfer} confirms this empirically: the optimal loss region remains relatively stable across capacity settings, with only minor drift consistent with the non-strict SDE transfer.

\paragraph{Granularity at fixed routing density.}
By granularity scaling we mean changing the expert partition $(N,h)\mapsto(N',h')$ while keeping the routing density fixed,
\begin{equation}
\label{eq:fixed-sparsity-ratio}
s:=\frac{a}{N}=\frac{a'}{N'}.
\end{equation}
Each sparse layer can be viewed as its Dense-MoE companion followed by the same sparsification ratio $s$: $(N,N,h)\mapsto(N,a,h)$ and $(N',N',h')\mapsto(N',a',h')$. The Dense-MoE companions have total widths $H_{\mathrm{dense}}=Nh$ and $H_{\mathrm{dense}}'=N'h'$, while the sparse active widths are
\begin{equation}
\label{eq:granularity-active-width}
H_a=ah=sNh,
\qquad
H_{a'}=a'h'=sN'h'.
\end{equation}
Therefore
\begin{equation}
\label{eq:granularity-width-ratio}
\frac{H_{a'}}{H_a}=\frac{N'h'}{Nh}=\frac{H_{\mathrm{dense}}'}{H_{\mathrm{dense}}},
\end{equation}
which is exactly the Dense-MoE companion width ratio. Granularity transfer at fixed density therefore reduces to Dense-MoE width transfer, while the routed output continues to use $r_a=a$ and $r_{a'}=a'$ on the two sides. On the stochastic side, both expert batch size and expert trained duration remain unchanged because
\begin{equation}
\label{eq:granularity-sde}
B_{\mathrm{exp}}=Bs,
\qquad
D_{\mathrm{exp}}=TBs,
\end{equation}
which are unchanged when $s$ is fixed, so no additional SDE correction is introduced.

\subsection{Hybrid, shared, and group-balanced-routing MoE blocks}
\label{app:hybrid-derivation}

The previous sections already provide all ingredients needed for shared experts, grouped MoE, and their combination. The rule is simplest when stated in three steps. First, every always-active dense submodule contributes active width but \emph{no} route scale. Second, every routed MoE group is selected using group-balanced routing, while global normalization ($\sum_{g,e}\pi_{g,e}(x)=1$) and global route scale $r=a$ are applied jointly across all selected experts. Third, after those dense and routed pieces are identified, one adds their active widths and treats the whole block as one equivalent expanded dense FFN, so one common FFN-output ABC parametrization is applied \emph{jointly} to all active submodules.

Let $\mathcal{D}$ index always-active dense/shared branches and $\mathcal{G}$ index routed groups. Dense branch $m\in\mathcal{D}$ has hidden width $H_m$. Routed group $g\in\mathcal{G}$ has $N_g$ total experts and per-expert width $h_g$; groups are used for balanced routing selection only. All selected experts share global routing weights $\pi_{g,e}(x)$ normalized jointly over all groups ($\sum_{g,e}\pi_{g,e}(x)=1$) and global route scale $r=a$. The total active width of the whole block is therefore
\begin{equation}
\label{eq:hybrid-total-width}
H_{\mathrm{tot}}=\sum_{m\in\mathcal{D}} H_m + a\sum_{g\in\mathcal{G}} h_g.
\end{equation}
All dense/shared up projections and all expert-local up/gate projections still follow the ordinary backbone-width $\mu$P rule; the hybrid-specific coupling appears only in the common output-layer ABC parametrization through the total expansion ratio $\rho_{H_{\mathrm{tot}}}=H_{\mathrm{tot}}/d$ and in the global route scale $r=a$.

Using this total active width, Complete-muE parameterizes the block as
\begin{equation}
\label{eq:hybrid-general-appendix}
y(x)=c_{\mathrm{tot}}\left[
\sum_{m\in\mathcal{D}} W_{\mathrm{down}}^{(m)}u^{(m)}(x)
+
a\sum_{g\in\mathcal{G}}\sum_{e=1}^{N_g}\pi_{g,e}(x)\,o_g^{(e)}(x)
\right],
\qquad
c_{\mathrm{tot}}=\frac{d}{H_{\mathrm{tot}}}.
\end{equation}
Equivalently, the whole block shares one common active-width ABC triple
\[
A_{\mathrm{tot}}=c_{\mathrm{tot}},
\qquad
B_{\mathrm{tot}}=\sigma_{\mathrm{down}}(d,H_{\mathrm{tot}}),
\qquad
C_{\mathrm{tot}}=\eta_{\mathrm{down}}(d,H_{\mathrm{tot}}),
\]
while the routed groups collectively carry one global route scale $r=a$; in a full model transfer, the actual output-layer init std and LR are obtained by further multiplying the standard backbone-width factors $\rho_d^{-1/2}$ and $\rho_d^{-1}$.

A shared-expert block is the special case with one always-active shared branch of width $H_{\mathrm{sh}}$ and one routed branch $(N,a,h)$, so
\[
y(x)=c_{\mathrm{tot}}\left[o_{\mathrm{sh}}(x)+a\sum_{e=1}^{N}\pi_e(x)\,o^{(e)}(x)\right],
\qquad
H_{\mathrm{tot}}=H_{\mathrm{sh}}+ah.
\]
This can be derived in three compositional steps. First split a dense FFN of width $H_{\mathrm{tot}}$ into two dense coordinate blocks of widths $H_{\mathrm{sh}}$ and $ah$; as a pure dense decomposition, both blocks inherit the same output multiplier $c_{\mathrm{tot}}$ and the same output initialization from the parent dense FFN, while the shared block needs no route scale. Next convert the routed-equivalent dense block of width $ah$ into a Dense-MoE companion with $a$ experts of width $h$, which introduces only the normalized-routing factor $a$. Finally apply the total-expert transfer from $(a,a,h)$ to $(N,a,h)$. By the capacity-scaling result above, this changes total routed capacity without introducing any additional first-order hyperparameter change. Hence the shared branch keeps the dense FFN setting and the routed branch keeps the same dense split plus the route scale, yielding the shared-expert formulation used in the main text.

The $\mu$P justification is straightforward. If the active hidden coordinates were concatenated into one vector of width $H_{\mathrm{tot}}$, Eq.~\eqref{eq:hybrid-general-appendix} would be exactly the dense FFN rule from Eq.~\eqref{eq:ffn-abc}; keeping separate output projections is simply its block decomposition. Consequently all active output projections share the same initialization and optimizer scale,
\begin{equation}
\label{eq:hybrid-init}
\sigma_{\mathrm{down}}(d,H_{\mathrm{tot}})=\sqrt{\frac{H_{\mathrm{tot}}}{d}}\,\sigma_{\mathrm{down}}^{(1)}(d),
\qquad
\eta_{\mathrm{down}}(d,H_{\mathrm{tot}})=\eta_{\mathrm{down}}^{(1)}(d),
\end{equation}
with the ordinary $\rho_d^{-1/2}$ and $\rho_d^{-1}$ factors composed on top if the backbone width changes. The routed groups additionally require the global route scale $r=a$, while dense/shared branches do not.

Assuming independent zero-mean initialization across active branches, the forward scale obeys
\begin{equation}
\label{eq:hybrid-forward}
\Var[y_i(x)]\asymp c_{\mathrm{tot}}^2 v_{\mathrm{down}}(d,H_{\mathrm{tot}}) q_u\left(\sum_{m\in\mathcal{D}}H_m+a\sum_{g\in\mathcal{G}}h_g F_g\right),
\end{equation}
where $F:=a\,\mathbb{E}\bigl[\sum_{g\in\mathcal{G}}\sum_{e=1}^{N_g}\pi_{g,e}(x)^2\bigr]$ is a single global order-one factor ($F\approx 1$ at initialization under uniform routing). The dense/shared branches contribute exactly their active widths. The routed groups contribute $a\sum_g h_g$ up to the global factor $F$. In practice $F\approx 1$, so Complete-muE sets $F=1$ and uses the common initialization rule in Eq.~\eqref{eq:hybrid-init} for all active submodules. Likewise the one-step functional update scales as
\begin{equation}
\label{eq:hybrid-update}
\mathbb{E}|\Delta y_i(x)|\asymp c_{\mathrm{tot}}\eta_{\mathrm{down}}(d,H_{\mathrm{tot}})\mu_u\left(\sum_{m\in\mathcal{D}}H_m+a\sum_{g\in\mathcal{G}}h_g\right)
=c_{\mathrm{tot}}\eta_{\mathrm{down}}(d,H_{\mathrm{tot}})H_{\mathrm{tot}}\mu_u,
\end{equation}
where we used $r=a$ and $\sum_{g,e}\pi_{g,e}(x)=1$. Hence shared experts, MoE with group-balanced routing, and arbitrary hybrids all reduce to the same total-active-width rule used in the main text.

\paragraph{Shared experts.}
A shared-expert MoE is the special case $|\mathcal{D}|=1$, $|\mathcal{G}|=1$:
\begin{equation}
\label{eq:shared-moe}
y(x)=c_{\mathrm{tot}}\left[
W_{\mathrm{down},\mathrm{sh}}\,u_{\mathrm{sh}}(x)
+
a\sum_{e=1}^{N}\pi_e(x)\,o^{(e)}(x)
\right],
\qquad
H_{\mathrm{tot}}=H_{\mathrm{sh}}+ah.
\end{equation}
The shared branch carries no route scale; the routed branch carries $r_a=a$; and both sets of output weights use the same initialization rule in Eq.~\eqref{eq:hybrid-init}. For example, if $H_{\mathrm{sh}}=0.5d$ and three routed experts are active each of width $h=0.5d$, then $H_{\mathrm{tot}}=0.5d+3\times0.5d=2d$, hence $c_{\mathrm{tot}}=1/2$ and
\[
\sigma_{\mathrm{down}}(d,H_{\mathrm{tot}})=\rho_d^{-1/2}\sqrt{2}\,\sigma_{\mathrm{down},\star}
\]
for both the shared-expert output weights and the routed-expert output weights.

\paragraph{Grouped MoE.}
For grouped MoE without shared experts, $\mathcal{D}=\varnothing$ and $G$ routed groups are simultaneously active:
\begin{equation}
\label{eq:grouped-moe}
y(x)=c_{\mathrm{tot}}\,a\sum_{g=1}^{G}\sum_{e=1}^{N_g}\pi_{g,e}(x)\,o_g^{(e)}(x),
\qquad
H_{\mathrm{tot}}=a\sum_{g=1}^{G} h_g.
\end{equation}
Groups are used for balanced routing selection only; all selected experts share one global route scale $r=a$ and global normalization ($\sum_{g,e}\pi_{g,e}(x)=1$), and all output projections share the same initialization rule determined by $H_{\mathrm{tot}}$. This remains true whether the implementation concatenates the active group features before a single down-projection or uses separate group projections and sums the outputs.

\paragraph{Shared experts plus grouped MoE.}
The combination is immediate: place always-active branches in $\mathcal{D}$, routed groups in $\mathcal{G}$, assign one global route scale $r=a$ across all routed experts (not per-group), and compute one common FFN multiplier from $H_{\mathrm{tot}}$ in Eq.~\eqref{eq:hybrid-total-width}. This is the most general Complete-muE rule for hybrid MoE blocks.

\subsection{How global AdamW transfer composes with Complete-muE}
\label{app:global-optimizer}

Table~\ref{tab:optimizer-rules} in the main text contains the remaining global AdamW transfer factors inherited from CompleteP / Complete$(d)$P~\cite{cerebras2025completep,apple2025completedmup}. These factors are applied \emph{after} the layer-level Complete-muE rule in Table~\ref{tab:mue-rules}. The connecting link between the SDE analysis in Appendix~\ref{app:sde-derivation} and the global optimizer table is the general expert-side correction formula: after the layer-level reparameterization, any change to the optimizer schedule or MoE architecture produces an expert-side LR and weight-decay multiplier of the form
\begin{equation}
\label{eq:expert-general-correction}
\eta'\approx\eta\sqrt{\frac{\rho_B^{\mathrm{exp}}}{\rho_D^{\mathrm{exp}}}},
\qquad
\lambda'\approx\lambda\sqrt{\frac{\rho_B^{\mathrm{exp}}}{\rho_D^{\mathrm{exp}}}},
\end{equation}
where $\rho_B^{\mathrm{exp}}$ and $\rho_D^{\mathrm{exp}}$ denote the expert-side batch and duration ratios. Whether this correction introduces a new multiplier on top of Table~\ref{tab:mue-rules} depends on which quantities change.

\paragraph{Global SDE framework for all-layer transfer.}
Let $\vartheta_t$ denote the rescaled parameter of any layer (dense or routed) after the Complete-muE layer-level reparameterization in Table~\ref{tab:mue-rules}. For a global batch of size $B$, the per-step stochastic gradient decomposes as
\begin{equation}
\label{eq:global-noise-model}
g_t = \bar{g} + \sigma_{\mathrm{noise}}\,\xi_t,
\qquad
\xi_t\sim\mathcal{N}(0,I),
\qquad
\sigma_{\mathrm{noise}}\propto B^{-1/2},
\end{equation}
where $\bar g$ is the mean gradient and $\sigma_{\mathrm{noise}}$ captures finite-batch stochasticity. Under the simplified RMSPropW proxy, the preconditioned parameter update reads
\begin{equation}
\label{eq:global-rmsprop-iter}
\vartheta_{t+1}
=
\vartheta_t
-
\eta\!\left(
\frac{g_t}{\sigma_{\mathrm{noise}}}
+
\lambda\vartheta_t
\right)
=
\vartheta_t
-
\frac{\eta}{\sigma_{\mathrm{noise}}}
\bigl(\bar g + \lambda\sigma_{\mathrm{noise}}\vartheta_t\bigr)
-
\eta\xi_t.
\end{equation}
Introducing the three SDE governing objects
\begin{equation}
\label{eq:global-sde-objects}
\sigma_0:=\eta\,\sigma_{\mathrm{noise}},
\qquad
\widetilde{\lambda}:=\frac{\lambda}{\eta},
\qquad
H_{\mathrm{SDE}}:=T\eta^2,
\end{equation}
and the rescaled time $\tau=t\eta^2$, Eq.~\eqref{eq:global-rmsprop-iter} becomes the Euler--Maruyama discretization of
\begin{equation}
\label{eq:global-rmsprop-sde}
d\Theta_\tau
=
-\frac{1}{\sigma_0}\bar g\,d\tau
-
\widetilde\lambda\Theta_\tau\,d\tau
-
dW_\tau,
\qquad
0\le\tau\le H_{\mathrm{SDE}}.
\end{equation}
In this normalized form the Brownian coefficient is fixed at one: the stochastic noise level is encoded entirely in $\sigma_0$ (smaller $\sigma_0$ means lower diffusion relative to drift), the weight-decay strength in $\widetilde\lambda$, and the training budget in $H_{\mathrm{SDE}}$. Two configurations yield statistically identical optimization trajectories if and only if all three objects match: $\sigma_0'=\sigma_0$, $\widetilde\lambda'=\widetilde\lambda$, and $H_{\mathrm{SDE}}'=H_{\mathrm{SDE}}$. The two cases below differ precisely in which of these conditions can be satisfied simultaneously.

\paragraph{Case 1: Fixed total tokens, changing batch size (exact transfer).}
Scale the global batch from $B$ to $B'=\kappa_B B$ and reduce the optimizer-step budget proportionally,
\begin{equation}
\label{eq:global-exact-tokens}
T'=\frac{T}{\kappa_B},
\end{equation}
so that the total trained tokens $BT=B'T'$ are unchanged. Every layer---dense and routed alike---then has
\begin{equation}
\label{eq:global-batch-factors}
\rho_B:=\frac{B'}{B}=\kappa_B,
\qquad
\rho_D:=\frac{B'T'}{BT}=1.
\end{equation}
From Eq.~\eqref{eq:global-noise-model}, the gradient-noise scale transforms as $\sigma_{\mathrm{noise}}'=\sigma_{\mathrm{noise}}/\sqrt{\kappa_B}$. We match all three SDE objects of Eq.~\eqref{eq:global-sde-objects} step by step. (i)~Matching $\sigma_0'=\sigma_0$, i.e.\ $\eta'\sigma_{\mathrm{noise}}'=\eta\sigma_{\mathrm{noise}}$, gives $\eta'=\sqrt{\kappa_B}\,\eta$. (ii)~Matching $\widetilde\lambda'=\widetilde\lambda$, i.e.\ $\lambda'/\eta'=\lambda/\eta$, then gives $\lambda'=\sqrt{\kappa_B}\,\lambda$. (iii)~The horizon is automatically preserved: $H_{\mathrm{SDE}}'=T'(\eta')^2=(T/\kappa_B)(\kappa_B\eta^2)=T\eta^2=H_{\mathrm{SDE}}$. All three SDE objects of Eq.~\eqref{eq:global-rmsprop-sde} are simultaneously preserved, so the transfer is exact:
\begin{equation}
\label{eq:global-exact-rule}
\eta'=\sqrt{\kappa_B}\,\eta=\sqrt{\rho_B}\,\eta,
\qquad
\lambda'=\sqrt{\kappa_B}\,\lambda=\sqrt{\rho_B}\,\lambda,
\qquad
(\rho_D=1).
\end{equation}
This is the CompleteP fixed-token batch rule, applied uniformly to all layers---dense and routed---with no additional expert-specific multiplier.

\paragraph{Case 2: Fixed training iterations, changing batch size (approximate transfer).}
Now keep the optimizer-step budget $T$ fixed while scaling the batch from $B$ to $B'=\kappa_B B$. Total trained tokens grow by $\kappa_B$, giving
\begin{equation}
\label{eq:global-approx-factors}
\rho_B=\kappa_B,
\qquad
\rho_D=\kappa_B
\qquad(\rho_B=\rho_D).
\end{equation}
In this case the two SDE invariance conditions $\sigma_0'=\sigma_0$ and $H_{\mathrm{SDE}}'=H_{\mathrm{SDE}}$ become mutually incompatible at fixed $T$: preserving $\sigma_0'=\sigma_0$ would require $\eta'=\sqrt{\kappa_B}\,\eta$, but then $H_{\mathrm{SDE}}'=T(\eta')^2=\kappa_B H_{\mathrm{SDE}}\ne H_{\mathrm{SDE}}$; conversely, preserving $H_{\mathrm{SDE}}'=H_{\mathrm{SDE}}$ forces $\eta'=\eta$, leaving $\sigma_0'=\eta\sigma_{\mathrm{noise}}'=\sigma_0/\sqrt{\kappa_B}\ne\sigma_0$. The practical choice is to fix the horizon and accept the $\sigma_0$ shift:
\begin{equation}
\label{eq:global-approx-rule}
\eta'\approx\eta,
\qquad
\sigma_0'=\frac{\sigma_0}{\sqrt{\kappa_B}},
\qquad
H_{\mathrm{SDE}}'=H_{\mathrm{SDE}}.
\end{equation}
Matching $\widetilde\lambda'=\widetilde\lambda$ at $\eta'=\eta$ gives $\lambda'=\lambda$ automatically. In Eq.~\eqref{eq:global-rmsprop-sde} this means the normalized SDE retains the same horizon $H_{\mathrm{SDE}}$ and decay rate $\widetilde\lambda$, while the diffusion amplitude $\sigma_0$ decreases: the target run sees less stochastic noise per step. This transfer is approximate rather than exact; the error for $\sigma_0'$ term is of order $|\kappa_B-1|$ in $\sigma_0$. Structurally this mirrors Bridge~II's activated-expert transfer: the residual $\sigma_0$ shift is not absorbed by the first-order $\eta,\lambda$ correction, so mild hyperparameter drift across batch sizes is expected. Figure~\ref{fig:batch-size-transfer-fixed-iterations} confirms this empirically: the optimal LR region remains relatively stable across batch sizes $128$--$1024$ at fixed training iterations, with only minor drift consistent with the bounded $\sigma_0$ shift.

\paragraph{Global batch and duration change at fixed MoE architecture.}
When the global batch size $B$ and training duration $T$ change while the MoE architecture remains fixed, every routed expert sees the same proportional change as the global model: $\rho_B^{\mathrm{exp}}=\rho_B$ and $\rho_D^{\mathrm{exp}}=\rho_D$. Eq.~\eqref{eq:expert-general-correction} then reduces to $\eta'\approx\eta\sqrt{\rho_B/\rho_D}$ and $\lambda'\approx\lambda\sqrt{\rho_B/\rho_D}$, recovering the familiar CompleteP rule. Cases~1--2 above make precise when this reduction is exact (fixed tokens, $\rho_D=1$) versus approximate (fixed iterations, $\rho_D=\kappa_B$).

\paragraph{MoE-specific changes at fixed global batch and duration.}
When instead the MoE architecture changes---activated-expert count $a$, total expert count $N$, or granularity at fixed routing density $s=a/N$---while global batch $B$ and total steps $T$ are held fixed, both the expert-side batch and the expert-side duration change by the same factor. As derived in Appendix~\ref{app:sde-derivation},
\[
\rho_B^{\mathrm{exp}}=\rho_D^{\mathrm{exp}},
\]
so the correction in Eq.~\eqref{eq:expert-general-correction} cancels to unity: $\eta'\approx\eta$ and $\lambda'\approx\lambda$. The layer-level reparameterization in Table~\ref{tab:mue-rules} already absorbs all first-order architectural changes---no additional global LR or WD multiplier is required for MoE-specific scaling axes.

For an FFN/MoE output branch with active-width ratio $\rho_H=H/d$ and backbone-width ratio $\rho_d=d/d_\star$, the full transfer therefore takes the form
\begin{equation}
\label{eq:full-composition}
\sigma_{\mathrm{down}}=\rho_d^{-1/2}\rho_H^{1/2}\sigma_{\mathrm{down},\star},
\qquad
\eta_{\mathrm{down}}=\rho_d^{-1}\sqrt{\frac{\rho_B}{\rho_D}}\,\eta_{\mathrm{down},\star},
\end{equation}
with the analogous batch/duration factors applied to weight decay, AdamW $\epsilon$, and $1-\beta_{1,2}$ according to Table~\ref{tab:optimizer-rules}. For tensors whose scaling is controlled only by the residual width $d$---the router readout and the FFN/MoE up/gate projections---the $\rho_H$ factor is absent and only the standard backbone-width and global optimizer multipliers remain.

Operationally, Complete-muE can be read as the following sequence. First choose the local FFN/MoE rule from Table~\ref{tab:mue-rules}: determine the active width, assign the route scale, and set the output-layer initialization. Then compose the global optimizer factors from Table~\ref{tab:optimizer-rules}: depth-sensitive residual groups use the usual CompleteP parameter grouping, while batch and duration changes contribute the familiar $\sqrt{\rho_B/\rho_D}$ correction on $\eta,\lambda$, together with the corresponding AdamW $\epsilon$ and momentum adjustments. The two-case structure above makes the scope of these global factors precise: they cover global schedule changes (Case A) but not MoE architectural changes (Case B), which are handled entirely within Table~\ref{tab:mue-rules}.


\section{LLM benchmark evaluation details}
\label{app:nlp-evaluation-details}

This appendix summarizes the post-training language benchmark protocol used for the results in Table~\ref{tab:large-scale-benchmark-comparison}. We do not include in-training benchmark probes in this paper. The final evaluation covers $13$ datasets spanning symbolic problem solving, world knowledge, commonsense reasoning, language understanding, and reading comprehension. Multiple-choice benchmarks are reported with accuracy, while free-response arithmetic is reported with exact match. The benchmark scores in Table~\ref{tab:large-scale-benchmark-comparison} use the 5-shot setting shown in that table.

\begin{table}[ht]
\centering
\caption{Post-training NLP evaluation datasets and reporting settings.}
\label{tab:appendix-nlp-evaluation-datasets}
\setlength{\tabcolsep}{2pt}
\small
\resizebox{\linewidth}{!}{%
\begin{tabular}{l l l c c l c}
\toprule
Area & Dataset & Split & \#Samples & Shot & Metric & Random Acc. \\
\midrule
Symbolic problem solving & SVAMP~\cite{patel2021nlp} & test & 300 & 5 & Exact Match & 0\% \\
World knowledge & MMLU~\cite{hendrycksmeasuring} & test & 14{,}042 & 5 & Multiple-choice Acc. & 25\% \\
World knowledge & ARC-Easy~\cite{clark2018think} & test & 2{,}376 & 5 & Multiple-choice Acc. & 25\% \\
World knowledge & ARC-Challenge~\cite{clark2018think} & test & 1{,}172 & 5 & Multiple-choice Acc. & 25\% \\
Commonsense reasoning & COPA~\cite{gordon2012semeval} & val & 100 & 5 & Multiple-choice Acc. & 50\% \\
Commonsense reasoning & PIQA~\cite{bisk2020piqa} & val & 1{,}838 & 5 & Multiple-choice Acc. & 50\% \\
Language understanding & HellaSwag~\cite{zellers2019hellaswag} & val & 10{,}042 & 5 & Multiple-choice Acc. & 25\% \\
Language understanding & WinoGrande~\cite{sakaguchi2021winogrande} & val & 1{,}267 & 5 & Cloze Acc. & 50\% \\
Language understanding & LAMBADA~\cite{paperno2016lambada} & test & 5{,}153 & 5 & Language-modeling Acc. & 0\% \\
Reading comprehension & BoolQ~\cite{clark2019boolq} & val & 3{,}270 & 5 & Multiple-choice Acc. & 50\% \\
Reading comprehension & AGIEval-LSAT-RC~\cite{zhong2024agieval} & test & 268 & 5 & Multiple-choice Acc. & 25\% \\
Reading comprehension & AGIEval-LSAT-LR~\cite{zhong2024agieval} & test & 510 & 5 & Multiple-choice Acc. & 25\% \\
Reading comprehension & AGIEval-SAT-EN~\cite{zhong2024agieval} & test & 206 & 5 & Multiple-choice Acc. & 25\% \\
\bottomrule
\end{tabular}%
}
\end{table}

\paragraph{Symbolic problem solving.}
SVAMP~\cite{patel2021nlp} contains $300$ short arithmetic word problems with free-form numerical answers. We use a chain-of-thought prompt before the final answer and score the final prediction with exact match; the chance-level baseline is effectively $0\%$.

\paragraph{World knowledge.}
MMLU~\cite{hendrycksmeasuring} evaluates broad academic knowledge over $57$ subjects and $14{,}042$ four-choice questions. ARC-Easy and ARC-Challenge~\cite{clark2018think} contain grade-school science questions, with $2{,}376$ and $1{,}172$ examples respectively. ARC-Easy focuses on more direct science knowledge, while ARC-Challenge emphasizes harder questions that require additional reasoning. These three datasets are reported with multiple-choice accuracy, with a $25\%$ random baseline.

\paragraph{Commonsense reasoning.}
COPA~\cite{gordon2012semeval} is a $100$-example two-choice task in which the model selects the more plausible cause or effect for a premise. PIQA~\cite{bisk2020piqa} contains $1{,}838$ two-choice physical commonsense questions about everyday actions and affordances. Both are evaluated with accuracy, and the random baseline is $50\%$.

\paragraph{Language understanding.}
HellaSwag~\cite{zellers2019hellaswag} asks the model to choose the most plausible ending for a partial narrative from four candidates, over $10{,}042$ validation examples. WinoGrande~\cite{sakaguchi2021winogrande} contains $1{,}267$ two-choice coreference examples formulated as cloze questions. LAMBADA~\cite{paperno2016lambada} evaluates long-context word prediction on $5{,}153$ passages by asking the model to predict the final token. HellaSwag uses a $25\%$ random baseline, WinoGrande uses a $50\%$ random baseline, and LAMBADA is treated as an open-vocabulary language-modeling task.

\paragraph{Reading comprehension.}
BoolQ~\cite{clark2019boolq} evaluates yes/no reading comprehension over $3{,}270$ passage-question pairs. AGIEval-LSAT-RC, AGIEval-LSAT-LR, and AGIEval-SAT-EN~\cite{zhong2024agieval} evaluate reading comprehension, logical reasoning, and SAT-style English respectively, with $268$, $510$, and $206$ four-choice examples. BoolQ has a $50\%$ random baseline, while the AGIEval subsets use a $25\%$ random baseline.


\end{document}